%% file: main.tex
\documentclass[dsc, EN, table]{ufabcFHZh}

\usepackage[T1]{fontenc}
\usepackage[utf8]{inputenc}
\usepackage{pdfpages}
\usepackage{pgfplots}
\usepackage[linesnumbered, ruled, vlined]{algorithm2e}
\usepackage{algorithmicx}
\usepackage{multirow}

\usepackage{upquote}
\usepackage{xspace}
\usepackage{array}
\usepackage[section]{placeins}
\usepackage{pdflscape}
\usepackage{caption}
\usepackage{subcaption}
\usepackage{tikz-cd}

\DeclareFontFamilySubstitution{TS1}{aer}{cmr}
\DeclareFontFamilySubstitution{TS1}{aett}{lmtt}
\DeclareFontFamilySubstitution{T1}{aett}{lmtt}

\input{NewCommands/Input_NewCommand}
\input{NewCommands/Input_NewCommand_Figures}
\SetKwProg{Fn}{\textbf{function}}{\string:}{}%

\togglefalse{LinksComCores} 		

\input{Pacotes/Input_Pacotes}

\makelosymbols

\makeloabbreviations

\addto\extrasenglish{

}
\pgfplotsset{compat=1.18} 

\begin{document}
\renewcommand{\emph}[1]{{\textit{#1}}}


\title{Functional Program Synthesis with Higher-Order Functions and Recursion Schemes}


\author{}{Matheus Campos Fernandes}

\advisor{Prof. Dr.}{Fabrício}{Olivetti de França}{}
\advisor{Prof. Dr.}{Emílio}{de Camargo Francesquini}{}

\examiner{Prof. Dr.}{Alcides Fonseca}{Examinador Externo I}
\examiner{Prof. Dr.}{Márcio Basgalupp}{Examinador Externo II}
\examiner{Prof. Dr.}{Mario Leston}{Examinador Interno I}
\examiner{Prof. Dr.}{Jerônimo Pellegrini}{Examinador Interno II}

\examiner{Prof. Dr.}{Fabrício Olivetti de França}{Presidente}
\examiner{Prof. Dr.}{Emílio de Camargo Francesquini}{Co-Orientador}
\coordinator{Prof. Dr.}{Fabrício}{Olivetti de França}{}
\date{31}{10}{2025} 

\department{CCM}
\maketitle

\frontmatter

{
\begin{center}
\vspace*{\fill}
This study was financed in part by the Coordenação de Aperfeiçoamento de Pessoal de Nível Superior – Brasil (CAPES) – Finance Code 001
\vspace*{\fill}
\end{center}
}

\dedication{
\textit{Aos Doutores:\\
Elizabeth Magalhães de Oliveira\\
Henrique Prado Fernandes
}}

\begin{agradecimentos}
	\input{pre_textual/agradecimentos}
\end{agradecimentos}

\begin{epigrafe}
	\input{pre_textual/epigraph}
\end{epigrafe}

\begin{foreignabstract}
	\input{pre_textual/abstract}
\end{foreignabstract}
\keywordeng
{Program Synthesis};
{Genetic Programming};
{Functional Languages};
{Recursion Schemes}

\begin{abstract}
	\input{pre_textual/resumo}
\end{abstract}
\palavrachave
{Síntese de Programas};
{Programação Genética};
{Linguagens Funcionais};
{Esquemas de recursão}

\listoffigures

\listoftables
\chapter*{List of Acronyms}
\input{pre_textual/acronym}
\printloabbreviations
\printlosymbols

\include{pre_textual/Include_Sim_1_0}

\makeatletter \let\ps@plain\ps@empty \makeatother 
\tableofcontents

\setcounter{pagenumber_frontmatter}{\number\value{page}}


\mainmatter

\iftoggle{toggleVersaoFinal}
{\setcounter{page}{\number\value{pagenumber_frontmatter} + 2}}
{\setcounter{page}{\number\value{pagenumber_frontmatter} + 1}}

\include{text/intro}
\include{text/ps}
\include{text/hotgp}
\include{text/origami}

\include{text/bananas}

\include{text/acdc}

\include{text/outro}


\backmatter

\bibliographystyle{Bibliografia/abntFHZ5}


\bibliography{Bibliografia/bib}
\input{Apendices/origami_solutions}


\end{document}

%% file: NewCommands/Input_NewCommand.tex
\newcommand{\code}[1]{\texttt{#1}}
\newcommand{\evolv}[1]{\textcolor{olive}{\underline{\texttt{#1}}}}
\newcommand{\chLit}[1]{\textquotesingle{#1}\textquotesingle}
\newcommand{\stLit}[1]{"#1"}
\newcommand{\negative}[1]{-#1}
\newcommand{\infix}[1]{\textasciigrave#1\textasciigrave }
\newcommand{\ie}{\emph{i.e.},\ }
\newcommand{\bestResult}[1]{\underline{\textbf{#1}}}
\newcommand{\canonical}{\color{blue}}

\newcommand{\checksum}{checksum\xspace}
\newcommand{\collatzNumbers}{collatz-numbers\xspace}
\newcommand{\compareStringLengths}{compare-string-lengths\xspace}
\newcommand{\countOdds}{count-odds\xspace}
\newcommand{\digits}{digits\xspace}
\newcommand{\doubleLetters}{double-letters\xspace}
\newcommand{\evenSquares}{even-squares\xspace}
\newcommand{\forLoopIndex}{for-loop-index\xspace}
\newcommand{\grade}{grade\xspace}
\newcommand{\lastIndexOfZero}{last-index-of-zero\xspace}
\newcommand{\median}{median\xspace}
\newcommand{\mirrorImage}{mirror-image\xspace}
\newcommand{\negativeToZero}{negative-to-zero\xspace}
\newcommand{\numberIo}{number-io\xspace}
\newcommand{\pigLatin}{pig-latin\xspace}
\newcommand{\replaceSpaceWithNewline}{replace-space-w-nl.\xspace}
\newcommand{\scrabbleScore}{scrabble-score\xspace}
\newcommand{\smallOrLarge}{small-or-large\xspace}
\newcommand{\smallest}{smallest\xspace}
\newcommand{\stringDifferences}{string-differences\xspace}
\newcommand{\stringLengthsBackwards}{string-lengths-back.\xspace}
\newcommand{\sumOfSquares}{sum-of-squares\xspace}
\newcommand{\superAnagrams}{super-anagrams\xspace}
\newcommand{\syllables}{syllables\xspace}
\newcommand{\vectorAverage}{vector-average\xspace}
\newcommand{\vectorsSummed}{vectors-summed\xspace}
\newcommand{\wallisPi}{wallis-pi\xspace}
\newcommand{\wordStats}{word-stats\xspace}
\newcommand{\xWordLines}{x-word-lines\xspace}

\newcommand{\lbparen}{%
  \mathopen{%
    \sbox0{$()$}%
    \setlength{\unitlength}{\dimexpr\ht0+\dp0}%
    \raisebox{-\dp0}{%
      \begin{picture}(.32,1)
      \linethickness{\fontdimen8\textfont3}
      \roundcap
      \put(0,0){\raisebox{\depth}{$($}}
      \polyline(0.32,0)(0,0)(0,1)(0.32,1)
      \end{picture}%
    }%
  }%
}

\newcommand{\rbparen}{%
  \mathclose{%
    \sbox0{$()$}%
    \setlength{\unitlength}{\dimexpr\ht0+\dp0}%
    \raisebox{-\dp0}{%
      \begin{picture}(.32,1)
      \linethickness{\fontdimen8\textfont3}
      \roundcap
      \put(-0.08,0){\raisebox{\depth}{$)$}}
      \polyline(0,0)(0.32,0)(0.32,1)(0,1)
      \end{picture}%
    }%
  }%
}

\usepackage{etoolbox} 		
\usepackage{colortbl}

\newcommand{\stripedRows}{\rowcolors{2}{gray!25}{white}}

\patchcmd{\quote}{\rightmargin}{\itshape\leftmargin4cm \rightmargin 0 }{}{}
\newtoggle{LinksComCores} 	

%% file: Pacotes/Input_Pacotes.tex
\usepackage[section]{placeins}
\usepackage{lastpage}		
\usepackage{indentfirst}	
\usepackage{color}			
\usepackage{microtype} 		
\usepackage{amsmath} 		
\usepackage{amssymb} 		

\usepackage{ae}
\usepackage{aecompl}
\usepackage{graphicx}	
\usepackage{float} 		
\usepackage[font= footnotesize]{caption} 
\usepackage{subcaption}
\usepackage{epstopdf}
\usepackage{ulem}
\usepackage{nomencl}
\makenomenclature
\usepackage{lipsum} 
\usepackage{morefloats}

\usepackage[Bjornstrup]{fncychap} 
\makeatletter
\renewcommand*{\@makechapterhead}[1]{%
	\vspace*{-1cm}
	{\parindent \z@ \raggedright \normalfont
		\ifnum \c@secnumdepth >\m@ne
		\if@mainmatter
		\DOCH
		\fi
		\fi
		\interlinepenalty\@M
		\if@mainmatter
		\DOTI{#1}%
		\else%
		\DOTIS{#1}%
		\fi
	}
}
\renewcommand*{\@makeschapterhead}[1]{%
	\vspace*{-1cm}
	{\parindent \z@ \raggedright
		\normalfont
		\interlinepenalty\@M
		\DOTIS{#1}
		\vskip 40\p@
	}
}
\makeatother

\usepackage{comment}

\usepackage{booktabs} 
\usepackage{multirow} 

\usepackage{paralist} 		
\usepackage[inline]{enumitem}

\usepackage{breqn} 		 
\usepackage{array} 		 
\usepackage{subeqnarray} 
\usepackage{cancel} 	 

\usepackage{acronym}
\usepackage{pgfgantt}

\usepackage[colorlinks=true,
linkcolor 		= red,
anchorcolor 	= black,
citecolor 		= green,
filecolor 		= cyan,
	\iftoggle{LinksComCores}{%
	}{%
		hidelinks
	}
]{hyperref} 

\usepackage {listings,soul}
\usepackage[frozencache,cachedir=.]{minted}

\usepackage{todonotes}
\usepackage{pict2e}
\usepackage{xspace}
\usepackage{xfrac}
\usepackage{stmaryrd}

%% file: pre_textual/agradecimentos.tex


\vspace{5 mm}

Sou grato primeiramente aos meus orientadores, Fabrício e Emílio, pela oportunidade
de trabalharmos juntos. 
Muito obrigado por todas as conversas, ensinamentos, dedicação e apoio e por todo o conhecimento que a orientação de vocês me permitiu adquirir. 
Nunca, sob nenhuma hipótese, me deixaram faltar nada, e a isso sou muito grato.

Agradeço à minha esposa, Júlia, por todo o suporte acadêmico e não acadêmico, paciência, parceria, incentivo, e por sempre acreditar em mim, sobretudo quando eu mesmo não o fazia.

À minha família, em especial meus pais José e Silvia e minha irmã Beatriz, pelo amor, educação e exemplo, e por terem me provido de condições que possibilitaram trilhar este caminho da forma como fiz.

Agradeço a todos os colegas que tive o prazer de conhecer durante essa jornada na UFABC, e a todos os seus professores e funcionários, por toda a competência e dedicação.

Agradeço também a todos os professores que tive na vida (acadêmicos ou não) por tanto me ensinarem e fomentarem minha curiosidade.


%% file: pre_textual/epigraph.tex



\vspace*{\fill}
\begin{flushright}
	\textit{``Our curiosity goes with you on your journey. You walk in the footsteps of those who came before you, and your path guides those who will follow later.'' \\
		(The Nomai)}
\end{flushright}


%% file: pre_textual/abstract.tex
Program synthesis is the process of generating a computer program following a set of specifications, which can be a high-level description of the problem and/or a set of input-output examples. 
The synthesis can be modeled as a search problem in which the search space is the set of all the programs valid under a grammar. 
As the search space is vast, brute force is usually not feasible, and search heuristics, such as genetic programming, also have difficulty navigating it without guidance.
This text presents two novel genetic programming algorithms that synthesize pure, typed, and functional programs: HOTGP and Origami.
HOTGP leverages the knowledge provided by the rich data types associated with the specification and the built-in grammar to constrain the search space and improve the performance of the synthesis. 
Its grammar is based on Haskell's standard base library (the synthesized code can be directly compiled using any standard Haskell compiler) and includes support for higher-order functions, $\lambda$-functions, and parametric polymorphism. 
Experimental results show that HOTGP is competitive with the state of the art.
Additionally, Origami is an algorithm that tackles the challenge of effectively handling loops and recursion by exploring Recursion Schemes.
The main advantage of writing a program using Recursion Schemes is that the programs are composed of well-defined templates with only a few parts that need to be synthesized.
Preliminary results on a prototype implementation show that, once the choice of which recursion scheme is made, the synthesis process can be simplified.
The first implementation of Origami is able to synthesize solutions in several Recursion Schemes and data structures, achieving a considerable improvement over its prototype.
Results also show that it is competitive with other GP methods in the literature, as well as LLMs.
%
The latest version of Origami employs AC/DC, a novel procedure designed to improve the search-space exploration.
%
Experimental results in $3$ different benchmarks show that it achieves considerable improvement over its previous version by raising success rates on every problem. 
Compared to similar methods in the literature, it has the highest count of problems solved with success rates of $100\%$, $\geq75\%$, and $\geq25\%$ across all benchmarks. 
In $18\%$ of all benchmark problems it stands as the only method to reach $100\%$ success rate, being the first known approach to achieve it on any problem in PSB2.
It also demonstrates competitive performance to large language models, achieving the highest overall win-rate against Copilot among all GP methods.

%% file: pre_textual/resumo.tex
Síntese de programas é o processo de gerar um programa de computador a partir de especificações, que pode ser uma descrição de alto nível do problema e/ou um conjunto de exemplos de entrada-saída.
A síntese pode ser modelada como um problema de busca, em que o espaço de busca é o conjunto de todos os programas válidos sob uma gramática.
Como o espaço de busca é amplo, força bruta geralmente não é viável, e heurísticas de busca, como a programação genética, também têm dificuldade de navegar sem orientação.
Este texto apresenta dois novos algoritmos de programação genética que sintetizam programas puros, tipados e funcionais: HOTGP e Origami.
HOTGP usa o conhecimento fornecido pelos tipos de dados associados à especificação e à gramática interna para restringir o espaço de busca e melhorar o desempenho da síntese.
Sua gramática é baseada na biblioteca padrão da linguagem Haskell (o código sintetizado pode ser compilado diretamente usando qualquer compilador Haskell) e inclui suporte para funções de alta ordem, funções $\lambda$ e polimorfismo paramétrico.
Os resultados experimentais mostram que o HOTGP é competitivo com o estado da arte.
Adicionalmente, Origami é um algoritmo que aborda o desafio de lidar com laços e recursão de maneira eficaz, explorando esquemas de recursão.
A principal vantagem de se escrever um programa utilizando Esquemas de Recursão é que os programas são compostos por modelos bem definidos com apenas algumas partes que precisam ser sintetizadas.
Resultados preliminares de uma implementação protótipo mostram que, uma vez feita a escolha de qual esquema de recursão usar, o processo de síntese pode ser simplificado.
A primeira implementação do Origami é capaz de sintetizar soluções em diversos esquemas de recursão e estruturas de dados, alcançando melhora considerável em relação ao seu protótipo.
Os resultados também mostram que ele é competitivo tanto com outros métodos de programação genética presentes na literatura quanto com LLMs.
A versão mais recente do Origami emprega o AC/DC, um procedimento novo projetado para melhorar a exploração do espaço de busca.
Resultados experimentais em $3$ benchmarks diferentes demonstram que ele obtém avanços consideráveis em relação à versão anterior, elevando as taxas de sucesso em todos os problemas.
Comparado a métodos similares na literatura, apresenta a maior quantidade de problemas resolvidos com taxas de sucesso de $100\%$, $\geq75\%$ e $\geq25\%$ em todos os \emph{benchmarks}.
Em $18\%$ dos problemas de todos os \emph{benchmarks}, é o único método a atingir $100\%$ de sucesso, sendo a primeira abordagem conhecida a atingí-lo em qualquer problema do PSB2.
Também demonstra desempenho competitivo em relação a LLMs, alcançando a maior taxa geral de vitórias contra o Copilot entre todos os métodos de programação genética.
\vspace{-.8em}

%% file: pre_textual/acronym.tex
\begin{acronym}
\acro{ADATE}{Automatic Design of Algorithms Through Evolution}
\acro{AC}{Abridge-by-Clipping}
\acro{AC/DC}{Abridge-by-Clipping, Diversify-by-Culling}
\acro{CBGP}{Code Building Genetic Programming}
\acro{DC}{Diversify-by-Culling}
\acro{DSLS}{Down-Sampled Lexicase Selection}
\acro{EA}{Evolutionary Algorithm}
\acro{FP}{Functional Programming}
\acro{G3P}{Grammar Guided Genetic Programming}
\acro{G3P+}{Grammar Guided Genetic Programming Extended}
\acro{GE}{Grammatical evolution}
\acro{GP}{Genetic Programming} 
\acro{HOTGP}{Higher-Order Typed Genetic Programming} 
\acro{HOF}{Higher-Order Function} 
\acro{PBE}{Programming By Examples} 
\acro{PS}{Program Synthesis} 
\acro{PSB1}{General Program Synthesis Benchmark 1} 
\acro{PSB2}{General Program Synthesis Benchmark 2} 
\acro{PolyPSB}{Polymorphic Program Synthesis Benchmark} 
\acro{SFGP}{Strongly Formed Genetic Programming} 
\acro{STGP}{Strongly Typed Genetic Programming} 
\acro{UMAD}{Uniform Mutation by Addition and Deletion}
\acro{RS}{Recursion Scheme}
\acro{LLM}{Large Language Model}
\acro{RHH}{Ramped Half-and-Half}

\end{acronym}

%% file: pre_textual/Include_Sim_1_0.tex
\chapter*{List of Symbols}
\label{cha:symbols}

\newcommand{\gac}{\ensuremath{g_{\mathit{ac}}}\xspace}
\newcommand{\gdc}{\ensuremath{g_{\mathit{dc}}}\xspace}

\begin{itemize}
\item[\gac] 		Interval, in generations, between successive applications of the Abridgment-by-Clipping (AC) procedure.
\item[\gdc] 		Interval, in generations, between successive applications of the Diversify-by-Culling (DC) procedure.
\end{itemize}


%% file: text/intro.tex
\chapter{Introduction}
\label{cha:intro}

Computer programming can be seen as the task of creating a set of instructions that, when executed, can provide as output a solution to a specific problem. 
This task involves several steps, starting from an abstract description of the solution (\emph{i.e.}, an algorithm) to a concrete implementation written in a programming language.

Given the importance of creating computer programs and the repetitive tasks usually involved, an often sought \emph{holy grail} is the ability to automatically generate source code, either in part or in its entirety. 
This automatic generation would follow a certain high-level specification, which reduces the burden of manually creating the program. This problem is known as \ac{PS}~\citep{gulwani2010dimensions}. 

A program specification is a high-level description of the objective of the program~\citep{koza2005genetic}. This specification can have different formats, from natural language to a more formal notation.
A common approach to specifying a program to solve a problem is providing a set of example inputs and their expected outputs.

In this case, the task of the synthesizer is to find a program that correctly maps each pair of input-output provided by the examples. 
The main advantage of this approach is that it can be easier to create input-output examples than to write the actual program.
On the other hand, providing a representative set containing corner cases might be difficult, possibly making the specification ambiguous and allowing for many alternative programs that do not behave as intended, even if they correctly map those examples.

\ac{PS} can be modeled as a search where the search space contains the set of all possible programs valid under a pre-specified grammar.
The objective is to find a program that meets the given specification.
The size of the search space makes it impractical to employ a na\"ive approach for selecting the best candidate from the enumeration of all possible programs.
For this reason, \ac{PS} is often done using a (meta-)heuristic approach, usually \ac{GP}~\citep{koza1992genetic}.

Traditionally, \ac{PS} research focuses on imperative programming, for example, by employing stack-based languages such as Push~\citep{spector2001autoconstructive}.
However, imperative languages expose programmers and synthesizers to errors which arise from implicit state changes~\citep{yu1999analysis, ray2014large}.
On the other hand, pure functional languages remove side effects by construction and prevent this specific category of errors.
When combined with strong typing, which some functional languages employ, ill-typed programs are forbidden from even existing, thus providing one more level of defense against bugs.

These are benefits that functional programmers have reaped for a long time, but \ac{PS} has just started to explore.
Recent research in the field has shown that this is a powerful combination~\citep{garrow2022functional}, with gaps that warrant exploration, as some functional constructs have still not been considered in the context of \ac{PS}.

One such example is by using \acp{RS}~\citep{meijer1991functional}. 
These are a reduced set of patterns that can implement a wide variety of recursive algorithms in many different data structures, by just changing the implementation of consumer and producer functions.
In a \ac{PS} setting, this would allow a synthesizer to work on simpler programs, as it would just have to synthesize the consumer/producer functions.

\section{Objectives}

Taking into account the motivation presented thus far, we formulate our research problem statement as follows:

\textbf{Problem Statement:}\emph{
    \ac{PS} poses challenges in terms of complexity and cost, primarily attributed to the vast search space involved.
    This study assesses whether the integration of functional programming principles, strong typing, and recursion schemes can mitigate these challenges, potentially leading to a more efficient synthesis, and identifies  potential drawbacks associated with this approach.
}

To this end, we propose and evaluate two novel PS algorithms that employ these concepts: \acf{HOTGP} and Origami.
This work also aims to contribute to the literature of \ac{PS}, \ac{GP}, and \ac{FP} in general, by providing a theoretical background, literature review, and presenting valuable experiments to those fields.

We highlight that a considerable portion of the contents of the current text has been published in journals, conferences, and as book chapters~\citep{fernandes2023hotgp, origami, bananas}, in which the candidate was the main author.
This text will also assume basic familiarity with the syntax of the ML-language family, specifically the Haskell programming language.

\section{Organization}

The remainder of this work is organized as follows:

\begin{itemize}
    \item \textbf{\autoref{cha:ps} - \nameref{cha:ps}} provides the theoretical background needed as a foundation for this text, as well as a literature review of recent methods;
    \item \textbf{\autoref{cha:hotgp} - \nameref{cha:hotgp}} introduces and evaluates our first novel algorithm, which expands traditional Genetic Programming by using functional programming constructs;
    \item \textbf{\autoref{cha:origami} - \nameref{cha:origami}} proposes Origami, a novel algorithm that employs the concept of recursion schemes to provide a framework for evolving recursive programs;
    \item \textbf{\autoref{cha:bananas} - \nameref{cha:bananas}} provides and evaluates the first complete implementation of Origami, following the description given in the previous chapter;
    \item \textbf{\autoref{cha:acdc} - \nameref{cha:acdc}} introduces a novel refinement procedure for a more efficient exploration of the search space, while extending Origami and performing a more comprehensive experimental evaluation;
    \item \textbf{\autoref{cha:outro} - \nameref{cha:outro}} summarizes the conclusions obtained by the previous chapters, providing some closing thoughts.
\end{itemize}

%% file: text/ps.tex
\chapter{Background and Motivation}
\label{cha:ps}

\ac{PS} is defined by \citet{manna1980deductive} as ``the systematic derivation of a program from a given specification''.
Formally, it can be described by two predicates $\varphi$ and $\Psi$ by the following:
\begin{quote}
    Given 
    an input predicate $\varphi(\textbf{x})$ 
    and an output predicate $\Psi(\textbf{x}, z)$, 
    construct a program computing a partial function $z = f(\textbf{x})$ 
    such that 
        if $\textbf{x}$ is an input vector satisfying $\varphi(\textbf{x})$, 
        then $f(\textbf{x})$ is defined and $\Psi(\textbf{x},f(\textbf{x}))$ is true. \citep{manna1971toward}
\end{quote}

In this definition, $\varphi$ dictates which inputs are valid for the given program, and $\Psi$ compares a given input to a candidate output and checks whether that is a valid program or not.

In more practical terms, \citet{gulwani2010dimensions} defines \ac{PS} as ``the task of discovering an executable program from user intent expressed in the form of some constraints'', and describes it as consisting of three dimensions:

\begin{enumerate}
    \item the kind of constraints that it accepts as an expression of user intent;
    \item the space of programs over which it searches;
    \item the search technique it employs.
\end{enumerate}

A user can express their intent by providing Logical Specifications, defining certain properties that must be satisfied by the generated program; by using a natural language description; by giving a detailed trace of a program; by providing a partial or incomplete program; or by providing a set of input and output examples.

In the present work, our interest is focused on the case for input-output examples, in which  \ac{PS} is called Inductive Synthesis or \ac{PBE}.
\ac{PBE} has the advantage of input-output examples being the simplest form of specification~\citep{gulwani2010dimensions}, and usually readily available~\citep{frankle2016example}.

In the remainder of this chapter, we explore different approaches to the \ac{PS} problem in terms of the two remaining dimensions.
\autoref{sec:gp} introduces \ac{GP} as a search technique for \ac{PS} and explores some of the challenges faced by its implementations.
In \autoref{sec:fp}, we introduce Functional Programming and Type Theory concepts, discussing how they can benefit \ac{PS} by constraining the search space.
\autoref{sec:fp_in_gp} provides a literature review of \ac{PS} methods which synthesize programs that employ Functional Programming concepts.
\autoref{sec:stgp} explores in detail one of those methods: \acf{STGP}.
Finally, \autoref{sec:ps_outro} presents concluding thoughts and connects the discussed concepts to the two methods introduced by this text.

\section{Genetic Programming}\label{sec:gp}

The \ac{GP} algorithm as described by Koza~\citep{koza1992genetic,koza2005genetic} is part of the of \ac{EA}~\citep{holland1975adaptation} family.
\acp{EA} simulate the process of natural selection and evolution to find a locally optimal solution. 
In general, they work by \emph{evolving} a set of solutions by repeatedly changing and combining them. 
The search is guided by a selective pressure imposed during the selection of candidates for reproduction.

In the context of \ac{PS}, \ac{GP} is employed to evolve candidate programs that aim to satisfy a given specification.
However, other uses of \ac{GP} have been documented, including symbolic regression~\citep{lenat11984automated, koza1992genetic, kronberger2024},
boolean function synthesis~\citep{pawlak2018competent, kalkreuth2023general},
control policies~\citep{nordin1995genetic, nadizar2024naturally}, 
and data mining applications~\citep{freitas2002data}, such as the discovery of classification rules, association rules, and feature selection or construction.

\begin{algorithm}[b!]
\caption{A general representation of the Genetic Programming algorithm}\label{algo:gp}

\SetKwFunction{EvolveProgram}{evolveProgram}%
\SetKwFunction{tossCoinWithProb}{tossCoinWithProb}%
\SetKwFunction{selectTwoIndividuals}{selectTwoIndividuals}%
\SetKwFunction{selectOneIndividual}{selectOneIndividual}%
\SetKwFunction{crossover}{crossover}%
\SetKwFunction{mutate}{mutate}%
\SetKwFunction{combine}{combine}%
\DontPrintSemicolon
\Fn{\EvolveProgram{popSize, crossoverRate}}{
    \DataSty{pop} $\gets$ \DataSty{initializePop(popSize)}\;\label{line:init}

    \While{stop criteria not met}{
        \DataSty{nextPop} $\gets \{\}$\;
        \While{$|\DataSty{nextPop}| < $ \DataSty{popSize}}{
            \If{\tossCoinWithProb{crossoverRate}}{
                \DataSty{ind1}, \DataSty{ind2} $\gets$ \selectTwoIndividuals{pop}\;
                \DataSty{ind1}, \DataSty{ind2} $\gets$ \crossover{ind1, ind2}\;
                \DataSty{nextPop} $\gets$ \DataSty{nextPop} $\cup\;\{\DataSty{ind1}, \DataSty{ind2}\}$
            }
            \Else{
                \DataSty{ind} $\gets$ \selectOneIndividual{pop}\;
                \DataSty{ind} $\gets$ \mutate{ind}\;
                \DataSty{nextPop} $\gets$ \DataSty{nextPop} $\cup\;\{\DataSty{ind}\}$
            }
        }
        \DataSty{pop} $\gets$ \combine{pop, nextPop}\;
    }
}
\end{algorithm}

This process is described by \autoref{algo:gp}.
It starts by randomly generating candidate solutions (called \textit{individuals}) to the problem (\autoref{line:init}). 
These individuals compose an initial \emph{population}, which can be initialized using standard GP tree-generation methods.

The \emph{grow} method expands each node by selecting either a terminal or a nonterminal until the depth limit is reached, producing trees of irregular shape.
In contrast, the \emph{full} method always selects nonterminals until the last level, where only terminals are allowed, producing perfectly balanced trees.
\ac{RHH} combines both: for each depth between a minimum and maximum, half of the individuals are generated with \emph{full} and the other half with \emph{grow}, increasing structural diversity in the initial population.

This population then undergoes a selection process in which some individuals will be chosen to generate offspring.
Generally speaking, ``good'' individuals should have a higher probability of being selected.
To quantify an individual's quality, we specify a \textit{fitness} measure.
Then, to advance the search, the selected individuals can go through two different operations: \textit{mutation} and \textit{crossover}.

Mutation takes an individual and produces a modified copy by applying a perturbation to promote diversity in the population.
In contrast, crossover combines two or more individuals, generating a new solution. 
The new solution could contain the good parts of both individuals, thus generating a better solution.

The last step involves combining the old population with the new one according to some strategy. 
Popular strategies for this include keeping only the new population, discarding the previous; keeping only the best individuals of the combined populations; or making the probability of an individual surviving in the new population a function of its fitness.

This entire process repeats until some stop criteria are met, such as a maximum number of generations, or when a certain fitness threshold is achieved, or when a certain computational budget is spent, such as time or memory.

\begin{figure}[t]
    \begin{subfigure}[b]{0.25\linewidth}
        \centering
        \includegraphics[width=\linewidth]{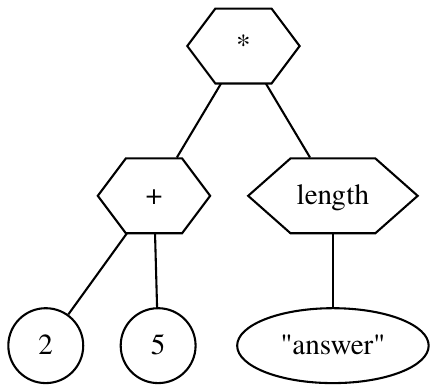}
        \caption{A program expressed as a syntax tree.}
        \label{fig:tree_example}
    \end{subfigure}
    \hfill
    \begin{subfigure}[b]{0.55\linewidth}
        \centering
        \includegraphics[width=\linewidth]{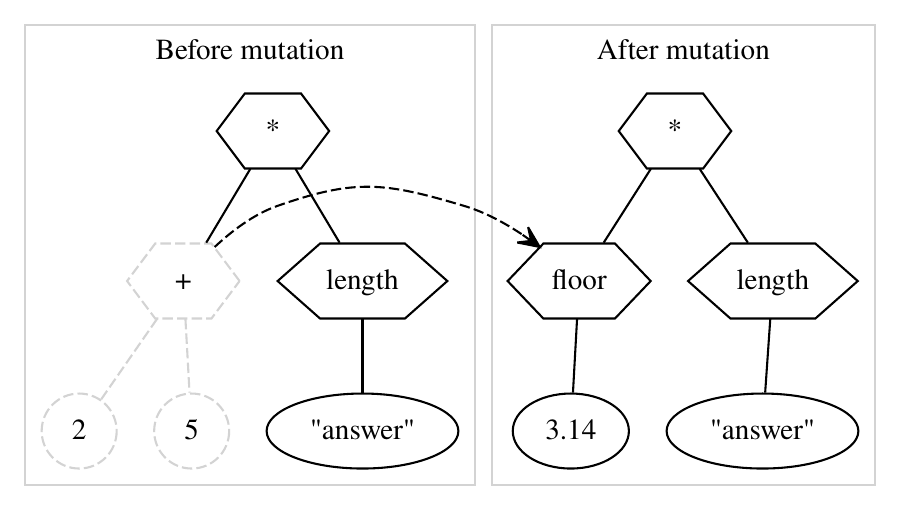}
        \caption{An example of mutation on a tree. In this, the entire \code{2+5} tree is replaced by a \code{floor(3.15)} tree.}
        \label{fig:tree_mutation}
    \end{subfigure}
    \caption{Examples of syntax trees. Leaves (terminals) are represented as ellipses and nodes (functions or non-terminals) as hexagons.}
    \label{fig:tree_graphs}
\end{figure}

In \ac{GP}, programs are traditionally represented as syntax trees~\citep{koza2005genetic}.
In a syntax tree, nodes are functions or operations, leaves are the values that these functions operate on, and the entire program is just an evaluable expression.
\autoref{fig:tree_example} illustrates the representation of the program \code{(2 + 5) * length("answer")}.

\begin{figure}[t]
    \centering
        \includegraphics[width=\linewidth]{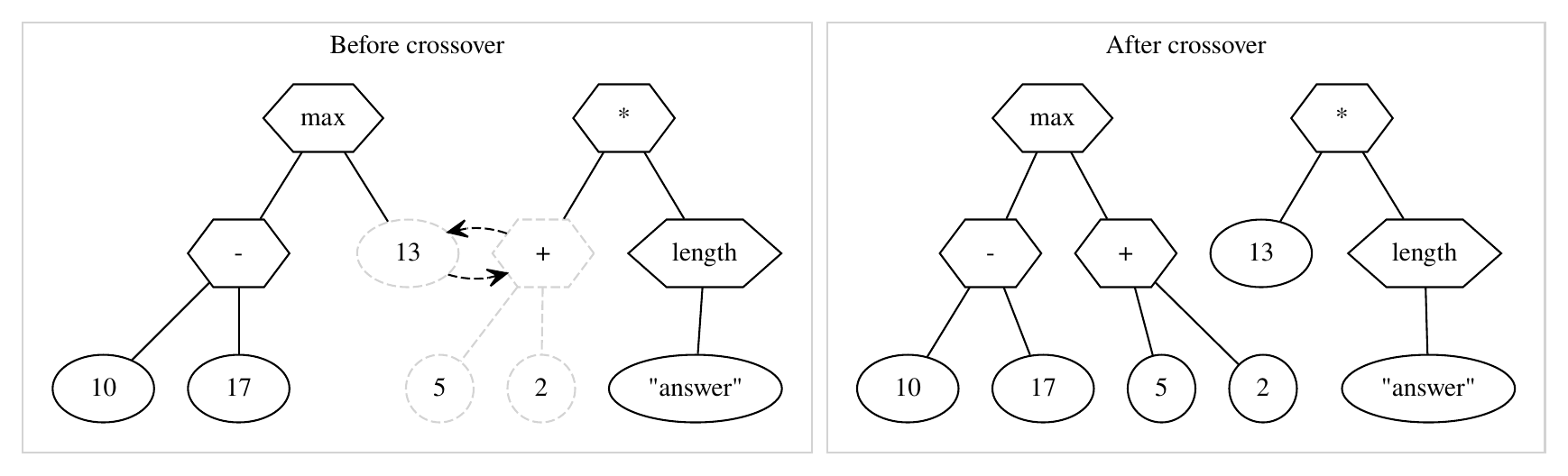}
    \caption{An example of crossover on two trees.}
    \label{fig:tree_cx}
\end{figure}

With this representation, the mutation operators simply replace a random subtree with a randomly generated tree, as illustrated in \autoref{fig:tree_mutation}. 
As alternative approaches, we can also replace a single node for another one of the same arity, change the value contained by a leaf node, etc.
The crossover operator swaps subtrees of the parents involved in the operation, as illustrated by \autoref{fig:tree_cx}.
The fitness measure used in this algorithm is any measure of error when evaluating the generated program with the input-output examples. 


Even though this is a straightforward representation, others have been proposed. For example, \citet{nordin1994compiling, banzhaf1998genetic} propose the use of a list of instructions, while \citet{pushgp} defends the use of a stack of instructions.

Some notable approaches of \ac{GP} for general program synthesis include:
\begin{itemize}
\item PushGP~\citep{helmuth2018program};
\item \ac{CBGP}~\citep{pantridge2020code};
\item \ac{GE}~\citep{o2001grammatical};
\item \ac{G3P}~\citep{manrique2009grammar}.
\end{itemize}
\ac{GP} also has recently been applied in many practical scenarios, including medical ~\citep{zhang2019genetic} and engineering applications~\citep{shariati2019application},
project scheduling~\citep{lin2020genetic},
and
COVID-19 pandemic forecasting~\citep{salgotra2020time}.

However, even though \ac{GP} has presented some success in standard \ac{PS} benchmarks~\citep{psb1,umad,kelly2019improving,hemberg2019domain,dsls}, it still struggles to consistently find solutions to some tasks that are trivial for humans.
Some of the problems commonly faced are:

\begin{enumerate}
    \item Traversing the search space is challenging, as sometimes a small change in the program code significantly impacts its output;
    \item Without additional information about the program, the search relies on the completeness of the examples;
    \item Some synthesizers can create stateful programs that can have unpredictable behavior depending on how the states are changed.
\end{enumerate}

In the next section, we explore one possible solution to alleviate these problems, which is to employ a statically typed and purely functional paradigm.

\section{On Functional Programming and Types}\label{sec:fp}

In \ac{FP}, a program is a \emph{pure} function and is defined as the composition of pure functions.
A pure function has, by definition, the fundamental property of 
\emph{referential transparency}~\citep{russell1910principia, strachey1967fundamental, sondergaard1990referential}. 

This means that any expression (including the whole program which is, by itself, an expression) can safely be substituted by the result of its evaluation.
In order to preserve this property, a pure function must have no other effect other than to compute its result, disallowing any side effects such as variable assignments, mutations, and stateful operations (including input and output).

However, iteration is a crucial part of programming, and it is typically expressed in imperative languages by using stateful loops.
In order to achieve the same results, functional programs employ \emph{recursion}.
Just as the \code{while} loop is an abstraction of \code{goto}, functional languages provide abstractions to common recursion patterns, such as \code{map}, \code{filter}, and \code{fold} functions.
Those abstractions are implemented as \acp{HOF}~\citep{jones1995system}, which in this context means a function that receives a function as one of its arguments.

Usually, functional languages also provide the tools to create $\lambda$-functions.
A $\lambda$-function, or anonymous function, or simply lambda, is a function definition not bound to a name.
These constructs allow the programmer to define new functions without the formality of naming them.
This goes hand-in-hand with \acp{HOF}, making it practical to define functions that are just meant to be passed as arguments.

Another more general approach is \acp{RS}~\citep{meijer1991functional}, which represent abstractions of \code{fold} and other recursive functions.
\acp{RS} are a set of common strategies for performing recursion while traversing data structures.
For example, consider the \code{fold} function present in many programming languages, which encapsulates traversing a list and producing an accumulated result, such as counting elements or performing a sum.
This function is abstracted by \emph{catamorphism}, one of the many \acp{RS}.
It generalizes \code{fold} to any traversable data structures, and can naturally also produce the same behavior as \code{map} and \code{filter}.


It is not uncommon (although not technically required) for functional languages to be statically typed, which means that each constant has a type, and functions have argument types and one output type.
As previously stated, the type notation for this text is inspired by the ML family languages, in which we denote concrete types by an upper case name.
For example, ``\code{42} is an integer'' is denoted as \code{42~::~Int}, while ``\code{"Hello world"} is a string'' is \code{"Hello world"~::~String}.
Furthermore, functions are always in the curried format \code{$a_1 \to a_2 \to \dots \to a_n \to b$}, where $a_i$ is the type of the $i^{th}$ argument of the function and $b$ is its return type: \code{add~::~Int~$\to$~Int~$\to$~Int}.
We write \code{[Int]} to represent the type of a list containing elements of type \code{Int} and \code{(Int,~Bool)} to represent the type of a pair whose first element is of type \code{Int} and the second of type \code{Bool}.
For other less-common data structures, we can write their names followed by their \emph{type parameters}, such as \code{Set Int} to represent a set whose elements are of type \code{Int}, and \code{Map String Float} for a map whose keys are of type \code{String} and values are of type \code{Float}.

Many type systems have the feature of \emph{parametric polymorphism} --- also known as generics in some programming languages --- which allows the type system to handle not only concrete types but also \emph{type variables}.
For example, consider we want to define the type of the \code{if} function, defined as \code{if~condition~branch1~branch2}. 
The first argument is the condition we want to evaluate, which must always be of type \code{Bool}. 
The two remaining arguments are the two possible branches that we pick and can be of any type, as long as they are the same and correspond to the expected return type. 
So, if the expected return type is \code{Int}, then both \code{branch1} and \code{branch2} must be of type \code{Int}.

Formally, we denote the type of the \code{if} function as $\forall \code{a}~.~\code{Bool}~\to~\code{a}~\to~\code{a}~\to~\code{a}$, which can be read as: \code{if} is a function that, given a value of type \code{Bool} and two values of the same type \code{a}, returns a value of type \code{a}. The letter \code{a} in this context is a \emph{type variable}, which can be assumed to be any type as per its universal quantifier ($\forall$). 
In this text, a type variable is always denoted in lowercase-single-letter (\emph{e.g.}, \code{a}), and $\forall$ will be henceforth implied whenever the type deals with a type variable.
The main benefit of having parametric polymorphism is that there is no need for multiple similar functions and data structures whose difference is the types on which they operate.

\subsection{Benefits to Program Synthesis}

The application of \ac{FP} and strong type systems helps to mitigate the three problems we described.
The property of referential transparency constrains the search space to only functions with no side effects.
Apart from making the search more effective, the use of pure functions is often associated to a reduction of the number of possible bugs~\citep{ray2014large}.
Furthermore, it also makes input transformations explicit and predictable, making mutation and crossover much more stable in comparison to imperative code.

Specifically regarding the reliance on example completeness, we argue that the user usually has more information about the program than simply some input-output examples.
For example, the user may allow only operations of certain types to be used during the \ac{PS}, either by prior knowledge or by preference.
The type information can play an essential role in the synthesis process as it can guide the search and constrain the search space making it easier to find programs closer to the correct solution.

In some instances, a well-defined type can constrain the program space to such an extent that only a single solution exists.
Take, for example, the type signature \code{a~$\to$~a}.
As explained by~\citet{wadler1989theorems}, a function that is fully generic in \code{a} cannot manufacture a new value of an unknown type (\emph{parametricity}).
Under this constraint, the identity function is the only implementation allowed by that type, as any other would either violate referential transparency or the type information.

Less extreme cases exhibit similar phenomena.
A function of type \code{(a, b)~$\to$~a} is uniquely determined: it must project the first component of the pair (\code{fst}), since no other value of type \code{a} can be constructed.
Likewise, the type \code{a~$\to$~Bool} admits only two implementations: a function that always returns \code{True} and one that always returns \code{False}.
Even for concrete types, similar restrictions occur.
For instance, \code{Bool~$\to$~Bool} allows only four total functions: the identity, negation, the constant-\code{True} function, and the constant-\code{False} function.
All of these examples illustrate how type information can constrain the space, removing unnecessary candidates~\citep{montana1995strongly}.

Within this same line of reasoning, we can illustrate the differences between imperative and functional paradigms for \ac{PS} using an example specification: given a list \code{xs}, return the sum of all elements smaller than 100.
A valid imperative style program is presented in \autoref{alg:imperative}.
Changes to any of the initial values of \code{s} (\autoref{line:sum}), any of the three sections of the \code{for} loop (\autoref{line:for}) or the constant $100$ (\autoref{line:compare}) can drop the accuracy from $100\%$ to $0\%$.
The choice to access the $i^{th}$ index might come naturally to a programmer, and is a \emph{very} common one, but for the synthesizer it should be no different than accessing the $(i + 1)^{th}$ or $42^{nd}$ index.
Finally, any additional statement inside the loop body that affects the value of either \code{i} or \code{s} might also decrease the achieved accuracy.

\begin{algorithm}[b]
    \caption{An imperative solution to the example specification.}\label{alg:imperative}
    \SetKwFunction{solution}{solution}%
    \SetKwProg{Fn}{\textbf{function}}{\string:}{}%
    \SetKwFunction{length}{length}%
    \DontPrintSemicolon

    \Fn{\solution{xs}}{
        \DataSty{s} $\gets 0$\; \label{line:sum}
        \For{$\DataSty{i} \gets 0;\ \DataSty{i} < \length{xs};\ \DataSty{i} \gets \DataSty{i} + 1$  \label{line:for} }{
            \If{
                $\DataSty{xs[i]} < 100$  \label{line:compare}
            }{
                \DataSty{s} $\gets$ \DataSty{s} + \DataSty{xs[i]}\;
            }
        }
        \Return \DataSty{s}
    }
\end{algorithm}

Let us compare this to a functional solution, such as the one in \autoref{alg:functional}.
At first glance, we can see that the code is shorter, as the \code{sum} and \code{filter} functions abstract away many of the complexities of the imperative solution.
Instead of manually iterating a list, the traversal is made implicitly by the \code{filter} \ac{HOF}, with the current element being provided as an argument to the predicate function.
Thus, the program does not have to deal with the complexities of keeping track of the counter \code{i}: initializing it to $0$; deciding to stop the iteration when it reaches the length of the list; incrementing it by $1$ every time; and using it to index the list.
Similarly, the \code{sum} function abstracts away the complexity of dealing with the \code{s} accumulator variable: initializing it to $0$; updating its value when needed; and returning it at the end.
From the synthesizer's perspective, the only common challenge in both implementations is the discovery of the \code{x < 100}, which is integral to this problem.

\begin{algorithm}[t]
    \caption{A functional solution to the example specification.}\label{alg:functional}
    \vspace{1em}
    \begin{minted}[linenos]{haskell}
solution :: [Int] -> Int
solution xs = sum (filter (\x -> x < 100) xs)
    \end{minted}
\end{algorithm}

Additionally, in this situation, we could also use the \code{fold} function to achieve the same results, as shown in \autoref{alg:functional_rs}.
As previously mentioned, this function is equivalent to the catamorphism \ac{RS}, 
but specialized to the \code{List} data structure.
The main advantage from the synthesizer's perspective is that the path of the recursion is defined by the base structure, and it only has to search for the smaller parts of the program which define the behavior of that specific problem (the function \code{f}, highlighted in cyan).

\begin{algorithm}[t!]
    \caption{A \ac{RS}-based solution to the example specification.}\label{alg:functional_rs}
    \vspace{1em}
    \begin{minted}[linenos,escapeinside=@@]{haskell}
solution xs = fold f xs
    where f x acc = @\color{cyan}\underline{\code{if x < 100 then x + acc else acc}}@
    \end{minted}
\end{algorithm}

We should highlight the importance of the representativeness of the examples, regardless of the paradigm, as it makes the specification less ambiguous.
For instance, if the input-example lists do not contain values between $95$ and $100$, choosing a wrong constant for the filter/if condition, such as $98$ or $96$, would still achieve $100\%$ accuracy on the training set, even though the code will produce wrong results when one considers all possible inputs.

In the following section, we present a brief review of the literature concerned with \ac{PS} implementations that generate typed and/or functional programs.

\section{Functional PS Algorithms}\label{sec:fp_in_gp}

To the best of our knowledge, \ac{ADATE}~\citep{olsson1995inductive} is the earliest example of \ac{PS} targeting functional code.
This work aimed at synthesizing recursive ML language programs using incremental transformations. 
The algorithm starts with an initial program described by the token "\code{?}" that always returns a \emph{don't know} value, which is equivalent to a function that always returns null.
After that, \ac{ADATE} systematically expands the expression into a pattern matching of the input type, synthesizing a program for each branch of the pattern match, and replacing the general case with a recursive call. 

\citet{montana1995strongly} proposes the \ac{STGP} algorithm, an adaption of \ac{GP} that considers the types of each function and terminal during the \ac{PS}.
Its goal is to further constrain the search space by allowing only correctly-typed programs to exist (\emph{i.e.}, programs in which all functions operate on values with the appropriate data types).

\ac{STGP} and a standard untyped \ac{GP} were compared by \citet{haynes1995strongly} using the ``Pursuit Problem''.
This problem models a game where four predators pursue a prey.
The goal is to create an algorithm for the predators to capture the prey as fast as possible. The prey always runs away from the nearest predator, and the predators only have information about themselves and the prey, but not about the other predators.
Results show that a good \ac{STGP} program can be generated faster than a good GP program. Moreover, the best \ac{STGP} program has a higher capture rate than the best \ac{GP} program.

PolyGP~\citep{yu1997polygp,yu2001polymorphism} extends \ac{STGP} with support to \acp{HOF} and $\lambda$-functions. 
It also differs from \ac{STGP} by using a type unification algorithm instead of a lookup table to determine the concrete types when using polymorphic functions. 
The $\lambda$-functions use the same initialization procedure of the main \ac{PS}, but the available terminals are limited to the input parameters.
As these $\lambda$-functions do not have any type restriction, they can be invalid, in which case they must be discarded and regenerated. 
The algorithm searches for a composition of $\lambda$-functions with a user-defined set of terminals and nonterminals as in \ac{STGP}.

\citet{katayama2005systematic} proposes MagicHaskell, a breadth-search approach that searches for a correctly-typed functional program using SKIBC~\citep{turner1979new} combinators. 
This simplifies the \ac{PS} by reducing the search space. 
MagicHaskell also introduces the use of the de Bruijn lambda to find equivalent expressions and memoization to improve performance~\citep{de1972lambda}. 
Additionally, it implements fusion rules to simplify the synthesized program further. 
This particular approach was reported not to work well with larger problems~\citep{10.1145/3067695.3082533}.

Schema-Guided Inductive Functional Programming~\citep{hofmann2010schema} follows a data-driven analytical approach to \ac{PS}.
In contrast to most of the methods so far, its process is driven by the shape of the examples rather than by external exploration strategies.
Instead of enumerating or evolving candidate programs, it derives the recursive structure of the target function by detecting which \ac{RS} satisfies the properties presented by the input–output examples.

\ac{SFGP}~\citep{sfgp2012} is an extension to \ac{STGP}. 
\ac{SFGP} not only assigns known data-types to terminals but also node-types to functions. 
A node-type identifies if a given node is a variable, an expression, or an assignment.
Each subtree of a function can also be required to be of a certain node-type.
The authors argue that this extra information is helpful to build correctly typed \emph{imperative} programs (\emph{e.g.}, the first child of an assignment must have the ``Variable'' node-type and match the data-type of the second child). 
They conducted experiments on $3$ datasets, with a reduced grammar that deals mainly with integers, and reported high success rates with a lower computational effort than competing methods.

\citet{g3p} criticize a common technique in \ac{GP}, which is to provide a different grammar for each problem. 
They argue that this leads to difficulties in grammar reuse, as they are specifically tailored to each problem.
They propose a general grammar to the \ac{G3P} algorithm, a grammar-guided system~\citep{manrique2009grammar}.
Similar to our work, they employ a different grammar for each set of types instead of a different grammar per benchmark problem, encoding its type system into its grammar, effectively enforcing that all syntactically-correct programs are also type-correct.
When performing experiments on \ac{PSB1}, the proposed grammar had difficulty with problems involving characters and strings.
The authors then proposed an improved and expanded grammar leading to \ac{G3P+}~\citep{g3pe}.

In this same line, \citet{garrow2022functional} compared the generation of Python and Haskell programs using \ac{G3P}.
Their approach supports \acp{HOF}, but limits the function arguments to pre-defined commonly used functions. 
Experimental results showed that the Haskell version consistently outperforms Python in most selected benchmarks. 
Implementing general $\lambda$-functions was left as future work by the authors since that would add complexity to the search space and must be carefully handled as a different construct from the main program.

Notably, \citet{swan2019stochastic} presented the first algorithm for synthesizing programs that exploit \acp{RS}. 
Their work focuses only on catamorphisms over natural numbers using Peano representation (\ie the inductive type of natural numbers). The authors evaluate their approach with variations of the Fibonacci sequence, successfully obtaining the correct programs.

\citet{santos2020refined} discuss desiderata for \ac{PS} approaches by further constraining the search space, similar to what is done by \ac{STGP}.
They propose the use of Refinement Types to this aim.
As this is an ongoing project, to the best of our knowledge, there are still no experimental evaluations or comparative results.

\citet{pantridge2022functional} propose an adaptation of the \ac{CBGP} algorithm~\citep{pantridge2020code} as a means to incorporate elements of functional programming such as \acp{HOF} and $\lambda$-functions.
\ac{CBGP} uses the same representation of PushGP with three primary constructs: \code{APP}, to apply a function; \code{ABS}, to define a function of $0$ or more arguments and; \code{LET}, to introduce local variables in the current scope. 
It also uses concepts from type theory to ensure the correctness of the polymorphic types. 
\ac{CBGP} achieved higher generalization rates for a subset of benchmark problems. 
However, for other problems, the generalization rate was close to $0$.
The authors noted that the evolutionary search avoided using $\lambda$-functions and preferred to employ pre-defined functions in \acp{HOF} such as \code{map}. 
These results show some indirect evidence of the benefits provided by type-safety to \ac{PS}, in particular, with regard to the stability of the solutions over different executions of the search algorithm.

\citet{he2022incorporating} investigate the reuse of already synthesized programs as subprograms to be incorporated in the nonterminal set.
The main idea is that, if the algorithm has already synthesized solutions to simpler tasks, these solutions can be used to build more complex solutions, in an incremental process. 
Their results show a significant benefit could be obtained by adding handcrafted modules in $4$ selected benchmarks.

A \ac{PS} algorithm that also employs \acp{RS} was recently proposed by \citet{hong2024recursive}.
The algorithm uses stochastic search to synthesize programs that use paramorphism, a variation of catamorphism that also allows access to the original structure that was the source for the current accumulated value.
It was evaluated on a selection of problems, both from previous benchmarks and derived from the Haskell standard library, achieving better results than its competitors.

Another approach recently explored for PS is the use of \acp{LLM}.
Unlike the other approaches described previously, LLMs are not grounded in \ac{GP} and operate on a textual description of the problem instead of a set of examples.
One example is GitHub Copilot, which was evaluated in the context of \ac{PS}~\cite{copilot}.
It was able to synthesize a correct program more often than a selection of GP methods in more than half of the problems in \ac{PSB1} and \ac{PSB2}.
However, \acp{LLM} are vulnerable to data contamination, as the benchmarks and their solutions may appear in the training corpus, inflating performance through memorization rather than genuine generalization.
Studies show that LLMs perform better on problems released before the training data cutoff~\citep{liu2024no}.
Specifically, ChatGPT achieves a higher success rate by $48$ percentage points on problems published before 2021 than on those published after.

As \ac{STGP} was among the first to propose and employ types for \ac{GP}, it has naturally influenced following works~\citep{sfgp2012, santos2020refined}.
In the following section, we explore \ac{STGP}'s technical details, in order to give a concrete example of how types can be integrated into \ac{GP}, as well as to build the background to understand the algorithms we propose in the chapters to come.

\section{Strongly Typed Genetic Programming}\label{sec:stgp}

\ac{STGP} is described by \citet{montana1995strongly} as:
\begin{quote}
an enhanced version of genetic programming which enforces data type constraints and whose use of generic functions and generic data types makes it more powerful than other approaches to type constraint enforcement.
\end{quote}

In contrast to standard \ac{GP}, where a given nonterminal must be capable of handling any data type, \ac{STGP} imposes extra constraints to enforce type-correctness.
These constraints state that every possible terminal (\emph{i.e.}, constants and arguments) must have a clearly defined type, and every possible non-terminal (function) must define the type of the arguments it takes and the type of its return value.
In other words, it should be well defined that the type of the constant value \code{15} is \code{Int} and that the function \code{AddInt} takes two arguments of type \code{Int} and returns a value of type \code{Int}.
Another important contribution of the \ac{STGP} is that the types of the nonterminals can employ parametric polymorphism.

Using type information during evolution constrains the search space by adding two new restrictions:
\begin{enumerate}
    \item The root node of the tree must return the expected return type of the program;
    \item Every non-root node (which is, by definition, an input argument to its parent node) must return the type required by its parent node.
\end{enumerate}

In order to enforce these restrictions, \ac{STGP} introduces some changes to the standard untyped \ac{GP} initialization process.
When choosing which element to use for a node, only elements that possess the correct type, satisfying both restrictions above, are considered. 
\ac{STGP} also keeps track of the depth of the shallowest tree that returns a given type and uses this information to, at each depth level, consider only functions whose arguments would not exceed the maximum defined depth.

Mutation and crossover were also changed, as they must respect the aforementioned restrictions.
When a node is selected for mutation, it is simply discarded and replaced by another node of \emph{the same type}.
The same algorithm used by the grow initialization is used to generate valid trees of a given type.
Crossover starts by randomly selecting the node in the first parent.
However, for the second parent, \ac{STGP} only considers nodes that return \emph{the same type} as the node selected from the first parent.
If there is no such node, crossover is not applied, and \ac{STGP} returns the parents.
These two modifications make it so the typing stays consistent, and an ill-typed tree is never generated from any of the two procedures.

The authors also argue in favor of handling runtime errors as part of the evolutionary process, penalizing individuals that present them.
This is the opposite of the original \ac{GP} approach~\citep{koza1992genetic}, which enforces that a value must always be returned.
For example: a division by zero in \ac{STGP} would result in an error signal, while in \ac{GP} it just returns 1.
\ac{STGP} also introduces the \code{Void} type for functions that do not return anything (\emph{i.e.}, procedures) and local variables which can be statefully changed during computation.

As for the \ac{EA} part, some changes were also made, namely:
\begin{itemize}
    \item Steady-state replacement: for each
generation, only one individual (or a small number of individuals) is generated and placed in the population rather than generating a whole new population;
    \item Exponential fitness normalization: this introduces a Parent-Scalar constant hyperparameter $0 < P_{scalar} < 1$. When selecting parents for reproduction, the individuals are ranked by their fitness. The probability of the $n^{th}$ best individual is given by $p(n) = P_{scalar} \times p(n-1)$.
\end{itemize}

\section{Final Remarks}\label{sec:ps_outro}

In this chapter, we presented the necessary theoretical background used as the basis for the methods proposed by this text.
We defined the concept of \ac{PS} and described the dimensions that form it.
Then, we introduced \ac{GP}, one of the most common search techniques for \ac{PS}, and explained in detail how it works.
From there, we combined concepts from Functional Programming and Type Theory, and described their possible benefits to \ac{GP} and \ac{PS} in general.
The detailed description of \ac{STGP} illustrated those benefits, by extending standard \ac{GP} in order to enforce strong types and discard ill-typed programs, effectively improving the search process.

In the following chapters, we introduce two new techniques.
The first one, \ac{HOTGP}, is related to \ac{STGP} in the sense that it also employs strong typing techniques to improve \ac{GP}, but by additionally allowing more advanced \ac{FP} constructs such as \acp{HOF} and $\lambda$-functions.
Then we introduce Origami, an idea that bases itself heavily in \ac{FP} concepts in order to generate recursive programs that employ \acp{RS}.

%% file: text/hotgp.tex
\chapter{Higher-Order Typed Genetic Programming}
\label{cha:hotgp}

This chapter proposes a new \ac{GP} algorithm, named \acf{HOTGP}, that searches for pure, typed, and functional programs. 
While still influenced by \ac{STGP}, \ac{HOTGP} introduces \acp{HOF} and $\lambda$-functions, drops the support for impure functions, and uses a general set of functions extracted from Haskell's base library. 
The main differences, detailed in the next sections, are:

\begin{itemize}
    \item \ac{HOTGP} builds programs using a pure functional program paradigm (a subset of the Haskell programming language) while \ac{STGP} is modeled after a combination of typed-LISP and ADA, allowing impure functions;
    \item Since \ac{HOTGP} is designed to only support pure functions, all side effects, including local variables (mutable state) and IO, are disallowed by design;
    \item As we shifted to a different language, appropriate changes to the set of functions were performed;
    \item Instead of specifying a strict set of terminals and non-terminals which are specific to each problem, we specify generic sets based on the input and output types\footnote{This is a design choice that was not explored by \ac{STGP} nor PolyGP.};
    \item Moreover, we use a more generic set of non-terminals (all available in the standard Haskell base library) instead of very specific functions that often need to be implemented by the user. 
    This characteristic, combined with the use of a subset of the Haskell language, allows for all the synthesized code to be immediately consumed by a Haskell compiler without modification;
    \item Finally, \ac{HOTGP} has support for \acp{HOF} (functions that accept $\lambda$-functions as input) to handle advanced constructs in the synthesized programs;
\end{itemize}

The remainder of this chapter is organized as follows.
\autoref{sec:hotgp} presents the \ac{HOTGP} algorithm, its main differences, advantages and challenges.
In \autoref{sec:hotgp_results}, we present the results obtained by running \ac{HOTGP} in a standard PS benchmark, comparing it to the state of the art.
\autoref{sec:hotgp_outro} provides a summary of the chapter, highlighting the key points of \ac{HOTGP}, and discusses future work.

We highlight that the contents of this chapter have been previously presented at the Genetic and Evolutionary Computation Conference (GECCO)~\citep{fernandes2023hotgp}.
A considerable portion of the text presented here is an adaptation or extension of the text presented in that paper.

\section{Higher-Order Typed Genetic Programming}\label{sec:hotgp}

Even though both \ac{HOTGP} and \ac{STGP} share the use of strong types, in both experimental evaluations of \ac{STGP}~\citep{montana1995strongly, haynes1995strongly}, the authors employed a limited function-set specifically crafted for each one of the benchmark problems. 
For example, to solve the Multidimensional Least Squares Regression problem, they used a minimal set of functions with matrix and vector operators such as \code{matrix\_transpose, matrix\_inverse, mat\_vec\_mult, mat\_mat\_mult}. 
Instead, this text uses a more general set of functions, common to all problems, all of which were extracted from the standard Haskell base library.

We argue that, in a practical scenario, providing only the functions needed for each problem is undesirable since it involves giving too much information to the algorithm. 
This is, in our opinion, not ideal since this piece of information might not be readily available beforehand. 
A much more reasonable demand on the user is to ask them for the acceptable types that each problem should handle internally.
This kind of information usually only requires as much intuition on the problem as providing examples.
This position is also defended by \citet{g3p}, proposing that providing a specific grammar for each problem results in ``bespoke grammars, making them difficult to reuse.''

\ac{HOTGP} supports primitive types such as integers, floating-point numbers, booleans, and characters, as well as parametric types in the form of pairs, linked lists, and $\lambda$-functions. 
These can be freely combined to form more complex types, \emph{e.g.}, a list of pairs of $\lambda$-functions or a string (represented as a list of characters).

As a consequence of using a subset of the Haskell language, \ac{HOTGP} precludes the use of impure functions. 
An essential property of pure functions is that, being without side effects, they are easier to compose. 
Thus, whenever the return type of one function is the same as the input type of another function, they can be composed to form a new, more complex pure function. 

The full set of functions allowed by \ac{HOTGP} is shown in \autoref{tab:functions}.
Most functions are common operations for their specific types. 
Since we employ a strongly-typed language, we also require conversion functions, such as \code{intToFloat} and \code{showInt}. 
Additional functions of common use include
sum and product for lists of numbers (integers and floating points),
\code{range}, which generates a list of numbers (equivalent to Haskell's \code{[x,y..z]});
\code{zip}, that pairs the elements of two lists given as input;
\code{take}, that returns the first $n$ elements of a list;
and \code{unlines}, that transforms a list of strings into a single string, joining them with a newline character.
In particular, \code{unlines} is needed for the benchmarks requiring the program to print text to the standard output (in our case, since we are working on a pure language, we simply return the output string).

\input{tables/functions_smaller}

It is worth noting that we included three \emph{constructor} functions: \code{toPair}, \code{cons}, and \code{singleton}. 
This is a deliberate choice to simplify the function-set. 
Let us take 2-tuples (pairs) as an example. 
We must be able to cope with constructions such as \code{(1, 2)} or \code{(1 + 2, 3 * 4)} (pairs of literals and pairs of expressions). 
However, both of these can be represented as applications of \code{toPair}.
The first example can be represented as \code{toPair 1 2}, which means applying the \code{toPair} function to the arguments \code{1} and \code{2}.
Similarly, the second example becomes \code{toPair (addInt 1 2) (multInt 3 4)}.

In other words, the construction of a pair is a simple function application with no special treatment. 
This has the added benefit of being directly compatible with the mutation and crossover operators already defined for regular nodes. 
Under the same reasoning, the evolution process can generate linked lists using a combination of the \code{cons} and \code{singleton} functions. 
For example, the list of the literals \code{1}, \code{2}, \code{3} can be represented as \code{cons~1~(cons 2 (singleton 3))}; and the list of the expressions \code{1 + 2}, \code{3 * 4}, \code {5 - 6} can be represented as \code{cons~(addInt~1~2) (cons~(multInt~3~4) (singleton~(subInt~5~6)))}.
As was also the case with pairs, this has the added benefit of enabling crossover and mutation to happen on just the heads or just the tails of such lists.

\ac{HOTGP} also allows the user to select which types it can use, to constrain the search space further.
Whenever the user selects a subset of the available types, the non-terminal set is inferred from \autoref{tab:functions} by selecting only those functions that support the selected types.
For example, if we select only the types \code{Int} and \code{Bool} we would allow functions such as \code{addInt}, \code{and}, \code{gtInt}, but would not allow functions such as \code{head}, \code{floor}, \code{showInt}.

Future implementations of this algorithm could support ad-hoc polymorphism, employing Haskell's type classes, so we could simply have \code{add, mult, sub} that determine their types by the context instead of having specific symbols for each type.

Another important distinction from \ac{STGP} to \ac{HOTGP} is the absence of the \code{Void} type, and constructs for creating local variables.
Therefore, impure functions and mutable state are not representable by \ac{HOTGP}. 
By construction, \ac{HOTGP} does not allow side effects and can only represent pure programs.
On the other hand, similarly to \ac{STGP}, runtime errors (such as divisions by zero) can still happen.
When they do, the fitness function assigns an infinitely bad fitness value to that solution.

\subsection{Higher-Order Functions and \texorpdfstring{$\lambda$}{λ}-functions}\label{sec:lambdas}

The main novelty of \ac{HOTGP} is the use of \acp{HOF}. 
To that end, adding support to $\lambda$-functions is essential, as they are first-class values that can be used as arguments to \acp{HOF}.

The introduction of lambdas requires additional care when creating or modifying a program.  
When evaluated, \ac{HOTGP}'s lambdas only have access to their own inputs, and not to the main program's.
In other words, they do not capture the environment in which they were created or in which they are executed.
This means that lambda terminals can be essentially considered ``sub-programs'' inside our program, and are generated as such.
We use the same initialization process from the main programs, using
the function type required by the current node and employing the \emph{grow} method.
However, two additional constraints must be respected.

Constraint 1 requires all lambdas to use their argument in at least one of its subtrees, which significantly reduces the possibility of the creation of a lambda that just returns a constant value. 
We argue that, for higher-order-function purposes, a lambda is required to use its argument in order to produce interesting results; otherwise the program could be simplified eliminating the use of this \ac{HOF} and returning a constant.
Note that this is only true because \ac{HOTGP} precludes the generation of expressions with side effects.

Constraint 2 takes the form of a configurable maximum depth of the lambda-trees, which is imposed to prevent our programs from growing too large.
However, as these lambdas can be nested, this hyperparameter alone is not enough to properly constrain the size of a program.
For instance, take a lambda as simple as \code{{\textbackslash}x~->~map~otherLambda~x}.
Depending on the allowed types, \code{otherLambda~=~{\textbackslash}x~->~map yetAnotherLambda x} would be a valid function and so on, which could lead to lambdas of arbitrarily large size.
Therefore, to prevent excessively large lambda nesting, we constrain nested functions to always be \code{{\textbackslash}x~->~x} (the identity function).

To enforce these constraints, similarly to \ac{STGP}, \ac{HOTGP} employs type-possibility tables to generate lambdas during both initialization and mutation, at the point where a lambda node must be constructed.
For the main program tree, as the argument and output types are known beforehand, both \ac{STGP} and \ac{HOTGP} only need to create two tables: one for the \emph{grow} and one for the \emph{full} method.
However, \ac{HOTGP} needs to generate lambdas involving every possible type allowed by the current program.  
Due to the recursive nature of the table, different argument types can lead up to vastly different type-possibility tables, so we need to keep one separate table for each possible argument type. 
As a corollary of Constraint 1, those tables are also guaranteed never to grow too large, as they never need to calculate possibilities for depths larger than the maximum lambda depth.
These tables also differ from the main tables in the sense that they only consider a node valid if at least one of its subtrees can have an argument leaf as a descendant, enforcing Constraint 2.

In terms of mutation and crossover, lambdas are treated as regular terminals, that is, as atomic entities.
Mutation never alters their internal structure: it always replaces a lambda node with a newly generated lambda produced by the method described above. 
Crossover treats lambdas in the same fashion: they may be selected and moved as a single unit, but no internal component is inspected, modified, or exchanged.

\subsection{Code Refinements}
\label{sec:code_refinements}
A well-known difficulty faced by \ac{GP} algorithms is the occurrence of \emph{bloat}~\citep{luke2006comparison}, an unnecessary and uncontrollable growth of a program without any benefit to the fitness function. 
This happens naturally as some building blocks that apparently do not affect the program's output survive during successive applications of crossover and mutation. 
Not only do these bloats make the generated program longer and unreadable, but they can also affect the performance on the test set.
For example, consider the task of doubling a number and the candidate solution \code{x0~*~(min~x0~900)}. 
If the training set does not contain input cases such that $\code{x0} > 900$, then this will be a correct solution from the algorithm's point-of-view.

\citet{helmuth2017improving} empirically show that simpler programs often have a higher generalization capability, in addition to being easier to understand and reason about.
\citet{pantridge2022functional}, for example, apply a refinement step at the end of the search, repeatedly trying to remove random sections of the program and checking for improvements.

To alleviate the effect of bloats, we also apply a refinement procedure on the best tree found, considering the training data.
Refinement starts by applying simplification rules, which remove redundancies from the code:
\begin{itemize}
    \item Constant evaluations: if there are no argument terminals involved in a certain tree-branch, it can always safely be evaluated to a constant value, \emph{e.g.} \code{head [4*5, 1+2]} $\equiv$ \code{20};
    \item General law-application: the simplifier has access to a table of hand-written simplification procedures, which are known to be true (laws).
    The complete list of laws is presented in \autoref{tab:simplification}.
\end{itemize}

\begin{table}[p]
    \centering
    
    \caption{Laws applied by the simplification system.}
    \label{tab:simplification}
    \stripedRows
    \begin{tabular}{rll}
    \toprule
    Rules&&Simplifies to\\
    \midrule
        \code{if True then a else b} &  $\equiv$ & \code{a}\\
        \code{if False then b else a} &  $\equiv$ & \code{a}\\
        \code{if b then a else a} &  $\equiv$ & \code{a }\\
        \code {if (not cond) then a else b} &  $\equiv$ & \code{if cond then b else a}\\
        \code{a == a} &  $\equiv$ & \code{True}\\
        \code{a < a} &  $\equiv$ & \code{False}\\
        \code{a > a} &  $\equiv$ & \code{False}\\
        \code{a - a} &  $\equiv$ & \code{0}\\
        \code{a / a} &  $\equiv$ & \code{1}\\
        \code{mod a a} &  $\equiv$ & \code{0}\\
        \code{max a a} &  $\equiv$ & \code{a}\\
        \code{min a a} &  $\equiv$ & \code{a}\\
        \code{a \&\& a} &  $\equiv$ & \code{a}\\
        \code{a || a} &  $\equiv$ & \code{a}\\
        \code{a * b/a} &  $\equiv$ & \code{b}\\
        \code{mod (a * b) a} &  $\equiv$ & \code{0}\\
        \code{fst (a, b)} &  $\equiv$ & \code{a}\\
        \code{snd (b, a)} &  $\equiv$ & \code{a}\\
        \code {b || True} &  $\equiv$ & \code{True}\\
        \code {a || False} &  $\equiv$ & \code{a}\\
        
        \code {b \&\& False} &  $\equiv$ & \code{False}\\
        \code {a \&\& True} &  $\equiv$ & \code{a}\\

        \code {a + 0} &  $\equiv$ & \code{a}\\
        \code {a - 0} &  $\equiv$ & \code{a}\\
        \code {a * 1} &  $\equiv$ & \code{a}\\
        \code {b * 0} &  $\equiv$ & \code{0}\\
        \code {a / 1} &  $\equiv$ & \code{a}\\
        \code {mod b 1} &  $\equiv$ & \code{0}\\

        \code {length (singleton b)} &  $\equiv$ & \code{1}\\
        \code {length (cons b a)} &  $\equiv$ & \code{1 + length a}\\
        \code {length (reverse a)} &  $\equiv$ & \code{length a}\\
        \code {head (singleton a)} &  $\equiv$ & \code{a}\\
        \code {reverse (singleton a)} &  $\equiv$ & \code{singleton a}\\
        \code {sum (singleton a)} &  $\equiv$ & \code{a}\\
        \code {product (singleton a)} &  $\equiv$ & \code{a}\\
        \code {reverse (reverse a)} &  $\equiv$ & \code{a}\\
        \code {take (length a) a} &  $\equiv$ & \code{a}\\
        \code {range a a b} &  $\equiv$ & \code{singleton a}\\
        
        \code{a == if cond then a else b} &  $\equiv$ & \code{cond}\\
        \code{a == if cond then b else a} &  $\equiv$ & \code{not cond}\\

      \bottomrule
    \end{tabular}
\end{table}

\SetKwFunction{LocalSearch}{localSearch}%

\SetKwFunction{subtrees}{getMatchingSubtrees}%
\SetKwFunction{replaceAtPath}{replaceAtPath}%
\SetKwFunction{error}{error}%
\SetKwFunction{numberOfNodes}{nNodes}%
\SetKwFunction{getNext}{nextPreOrderPath}%
\SetKwFunction{getNode}{getNode}%
\SetKw{Continue}{continue}%
\newcommand{\GrayComment}{\color{gray}}%
\SetCommentSty{GrayComment}%

\begin{algorithm}[t]
\caption{The local search procedure}\label{alg:ls}
\DontPrintSemicolon

\Fn{\LocalSearch{tree, path}}{
    $node \gets \getNode(tree, path)$\;
    
    \lIf{$node = \text{null}$\label{lin:stop_crit}}{\Return tree}
    
    $bestTree \gets tree$\;
    
    \ForEach{$child \in node.children$}{
        \lIf{$child.outputType \neq node.outputType$}{\Continue\label{lin:compare_type}}
        
        $newTree \gets \replaceAtPath(tree, path, child)$\label{lin:replace}\;
        
        \If{$\error(newTree) < \error(bestTree)$}{
            \tcp{found a more correct tree}
            $bestTree \gets newTree$\;
        }
        \ElseIf{
            $\error(newTree) = \error(bestTree)$
            \textbf{and}
            $\numberOfNodes(newTree) < \numberOfNodes(bestTree)$
        }{
            \tcp{found an equally-correct tree, but with fewer nodes}
            $bestTree \gets newTree$\;
        }
    }
    
    \If{$bestTree \neq tree$}{
        \tcp{do not advance: current subtree changed}
        \Return \LocalSearch($bestTree$, $path$)\label{lin:not_next}\;
    }
    
    $nextPath \gets \getNext(tree, path)$\;
    
    \Return \LocalSearch($tree$, $nextPath$)\label{lin:next}\;
}
\end{algorithm}

After this step, \ac{HOTGP} applies a Local Search procedure that removes parts of the tree that do not contribute to correctness on the training set or that decrease it.
This procedure replaces a node with each of its children, keeping the version that yields the best result according to correctness and tree size, as described by \autoref{alg:ls}. 
It takes as input a \ArgSty{tree} and a \ArgSty{path} identifying the current node, with \ArgSty{path} initially set to the root of the tree being simplified.
It then inspects the children of the current node that share its output type (\autoref{lin:compare_type}). 
For each such child, it builds a copy of the entire tree in which the subtree at the current path is replaced by that child subtree (\autoref{lin:replace}).
These candidates, along with the original tree, are compared, and the best one in terms of error and number of nodes is kept.
If a change occurs, the algorithm continues from the same position because the subtree has just been replaced (\autoref{lin:not_next}); otherwise, it advances to the next node in pre-order (\autoref{lin:next}).
The procedure terminates when no further positions remain to be examined (\autoref{lin:stop_crit}).

\section{Results}
\label{sec:hotgp_results}


In this section, we compare \ac{HOTGP} to state-of-the-art GP-based program synthesis algorithms found in the literature. 
For this comparison, we employ the ``General Program Synthesis Benchmark Suite''~\citep{psb1}, which contains a total of $29$ benchmark problems for inductive program synthesis\footnote{The full source code for \ac{HOTGP} can be downloaded from: \href{https://github.com/mcf1110/hotgp}{https://github.com/mcf1110/hotgp}.}.


Following the recommended instructions provided by \citet{psb1}, we executed the algorithm using $100$ different seeds for each benchmark problem. 
We used the recommended number of training and test instances and included the fixed edge cases in the training data.
We also used the same fitness functions described in their paper.

For the evolutionary search, we used a steady-state replacement of $2$ individuals per step, with an initial population of $1\,000$, and using a Parent-Scalar of $99.93\%$.
The maximum tree depth was set to $15$ for the main program and $3$ for $\lambda$-functions.
The crossover and mutation rates were both empirically set to $50\%$. 
We allowed a maximum of $300\,000$ evaluations with an early stop whenever the algorithm finds a perfectly accurate solution according to the training data.

We report the percentage of correct solutions found within the $100$ executions taking into consideration the training and test data sets, before and after the refinement process. 
To position such results with the current literature, we compare the obtained results against those obtained by:
\begin{itemize}
    \item \textbf{PushGP}, which is the original baseline presented by the benchmark authors~\citep{psb1};
    \item Recent improvements of PushGP, namely \textbf{\ac{UMAD}}~\citep{umad} and \textbf{\ac{DSLS}}~\citep{dsls};
    \item \textbf{\ac{G3P}}~\citep{g3p} and \textbf{\ac{G3P+}}~\citep{g3pe};
    \item \textbf{\ac{CBGP}}~\citep{pantridge2022functional} and G3P with Haskell and Python grammars (\textbf{G3Phs} and \textbf{G3Ppy})~\citep{garrow2022functional}.
\end{itemize}

We have not compared with \ac{STGP} and PolyGP since their original papers~\citep{montana1995strongly,yu2001polymorphism} predate this benchmark suite. 
All the obtained results are reported in \autoref{tab:comparison}. 
In this table, all the benchmarks that could not be solved with our current function set are marked with ``--'' in \ac{HOTGP} columns. 
For the other approaches, the dash mark means the authors did not test their algorithm for that specific benchmark.

\afterpage{
\begin{landscape}
\begin{table}[p]
\small
\centering
\vspace{-4em}
\caption{Successful solutions found for each problem (\% of executions) considering the training (Tr) and test (Te) data sets. HOTGP* lists the results obtained with HOTGP after the simplification procedure. The best values for the test data sets of each problem are highlighted. The 
\emph{collatz-numbers}, \emph{string-differences}, \emph{wallis-pi} and \emph{word-stats} problems are omitted as no algorithm was able to find results for those problems.
}
\label{tab:comparison}
\stripedRows
\begin{tabular}{l|rr|rr|r|r|r|r|r|rr|rr|rr}
\toprule
 & \multicolumn{2}{c|}{\ac{HOTGP}} 
 & \multicolumn{2}{c|}{\textbf{\ac{HOTGP}*}} 
 & PushGP
 & \ac{UMAD} 
 & \ac{DSLS}
 & \ac{CBGP} 
 & \ac{G3P} 
 & \multicolumn{2}{c|}{\ac{G3P+}} 
 & \multicolumn{2}{c|}{\ac{G3P}hs} 
 & \multicolumn{2}{c}{\ac{G3P}py} \\
Benchmark 
&  Tr &  Te 
&  Tr &  Te 
&  Te       
&  Te       
&  Te       
&  Te       
&  Te       
&  Tr & Te  
&  Tr &  Te 
&  Tr &  Te 
\\
\midrule
checksum & -- & -- & -- & -- & 0 & 5 & \bestResult{18} & -- & 0 & 0 & 0 & -- & -- & -- & -- \\
compare-string-lengths & 100 & \bestResult{100} & 100 & \bestResult{100} & 7 & 42 & 51 & 22 & 2 & 96 & 0 & 94 & 5 & 12 & 0 \\
count-odds & 46 & 46 & 50 & \bestResult{50} & 8 & 12 & 11 & 0 & 12 & 4 & 3 & -- & -- & -- & -- \\
digits & -- & -- & -- & -- & 7 & 11 & \bestResult{28} & 0 & 0 & 0 & 0 & -- & -- & -- & -- \\
double-letters & 0 & 0 & 0 & 0 & 6 & 20 & \bestResult{50} & -- & 0 & 0 & 0 & -- & -- & -- & -- \\
even-squares & 0 & 0 & 0 & 0 & \bestResult{2} & 0 & \bestResult{2} & -- & 1 & 0 & 0 & -- & -- & -- & -- \\
for-loop-index & 73 & 39 & 73 & \bestResult{59} & 1 & 1 & 5 & 0 & 8 & 9 & 6 & -- & -- & -- & -- \\
grade & 37 & 32 & 39 & \bestResult{37} & 4 & 0 & 2 & -- & 31 & 63 & 31 & -- & -- & -- & -- \\
last-index-of-zero & 0 & 0 & 0 & 0 & 21 & 56 & \bestResult{65} & 10 & 22 & 97 & 44 & 0 & 0 & 2 & 2 \\
median & 82 & 73 & 100 & \bestResult{99} & 45 & 48 & 69 & 98 & 79 & 99 & 59 & 100 & 96 & 39 & 21 \\
mirror-image & 1 & 1 & 1 & 1 & 78 & \bestResult{100} & 99 & \bestResult{100} & 0 & 89 & 25 & -- & -- & -- & -- \\
negative-to-zero & 100 & \bestResult{100} & 100 & \bestResult{100} & 45 & 82 & 82 & 99 & 63 & 24 & 13 & 0 & 0 & 68 & 66 \\
number-io & 100 & \bestResult{100} & 100 & \bestResult{100} & 98 & \bestResult{100} & 99 & \bestResult{100} & 94 & 95 & 83 & 100 & 99 & 100 & \bestResult{100} \\
pig-latin & -- & -- & -- & -- & 0 & 0 & 0 & -- & 0 & 4 & \bestResult{3} & -- & -- & -- & -- \\
replace-space-with-newline & 38 & 38 & 38 & 38 & 51 & 87 & \bestResult{100} & 0 & 0 & 29 & 16 & -- & -- & -- & -- \\
scrabble-score & -- & -- & -- & -- & 2 & 20 & \bestResult{31} & -- & 2 & 1 & 1 & -- & -- & -- & -- \\
small-or-large & 28 & \bestResult{59} & 28 & \bestResult{59} & 5 & 4 & 22 & 0 & 7 & 39 & 9 & 30 & 4 & 0 & 0 \\
smallest & 98 & 95 & 100 & \bestResult{100} & 81 & \bestResult{100} & 98 & \bestResult{100} & 94 & 100 & 73 & 100 & \bestResult{100} & 99 & 89 \\
string-lengths-backwards & 87 & 87 & 89 & 89 & 66 & 86 & \bestResult{95} & -- & 68 & 20 & 18 & 0 & 0 & 35 & 34 \\
sum-of-squares & 1 & 1 & 1 & 1 & 6 & \bestResult{26} & 25 & -- & 3 & 5 & 5 & -- & -- & -- & -- \\
super-anagrams & -- & -- & -- & -- & 0 & 0 & 4 & -- & 21 & 43 & 0 & 30 & 5 & 51 & \bestResult{38} \\
syllables & 0 & 0 & 0 & 0 & 18 & 48 & \bestResult{64} & -- & 0 & 53 & 39 & -- & -- & -- & -- \\
vector-average & 78 & 79 & 80 & 80 & 16 & 92 & \bestResult{97} & 88 & 5 & 0 & 0 & 67 & 4 & 0 & 0 \\
vectors-summed & 34 & 34 & 37 & 37 & 1 & 9 & 21 & \bestResult{100} & 91 & 28 & 21 & 100 & 68 & 0 & 0 \\
x-word-lines & -- & -- & -- & -- & 8 & 59 & \bestResult{91} & -- & 0 & 0 & 0 & -- & -- & -- & -- \\

\midrule
\textbf{\# of Best Results} &  & 4 &  & 9 & 1 & 4 & \bestResult{11} & 4 & 0 & & 1 & & 1 & & 2 \\
\midrule

$\mathbf{= 100\%}$ & 3 & 3 & 5 & \bestResult{4} & 0 & 3 & 1 & \bestResult{4} & 0 & 1 & 0 & 4 & 1 & 1 & 1 \\
$\mathbf{\geq 75\%}$ & 7 & 6 & 7 & 7 & 3 & 7 & \bestResult{8} & 7 & 4 & 6 & 1 & 5 & 3 & 2 & 2 \\
$\mathbf{\geq 50\%}$ & 8 & 8 & 9 & 10 & 5 & 9 & \bestResult{13} & 7 & 6 & 8 & 3 & 6 & 4 & 4 & 3 \\
$\mathbf{> 0\%}$ & 15 & 15 & 15 & 15 & 22 & 21 & \bestResult{24} & 9 & 17 & 19 & 17 & 8 & 8 & 8 & 7 \\

\bottomrule
\end{tabular}
\end{table}
\end{landscape}
}

\subsection{Analysis of the results}

Compared to the other algorithms, \ac{HOTGP} has the highest success rate for the test set in $9$ of the benchmark problems, which is a close second to the $11$ problems of \ac{DSLS}.
Both are followed by PushGP and \ac{CBGP}, which got the highest rate for $7$ and $5$ of the benchmarks, respectively.
An important point to highlight is that \ac{HOTGP} obtained a $100\%$ success rate in $4$ problems, a result only matched by \ac{CBGP}.
Specifically, \ac{HOTGP} was the first method that was able to get a $100\%$ success rate in \emph{negative-to-zero} and \emph{compare-string-lengths}, and presented significant improvements over the competition in \emph{count-odds}, \emph{for-loop-index}, and \emph{small-or-large}.
Moreover, \ac{HOTGP} obtained at least $75\%$ in $7$ problems, leaving it tied with \ac{CBGP} and just $1$ problem behind DSLS.
Also, it obtained $50\%$ in $10$ out of the 29 problems, placing it just a little behind \ac{DSLS}, but still slightly in front of \ac{UMAD}.

Overall, these results show that \ac{HOTGP} is competitive with \ac{DSLS} and \ac{UMAD}, while still being as consistent as \ac{CBGP}, obtaining $100\%$ in more problems than the competition.
These 3 methods are very recent and considered the state-of-the-art, positioning \ac{HOTGP} as a strong method for PS.
This brings evidence to the hypothesis that including type information and using a functional paradigm for program synthesis can, indeed, reduce the search space to improve the efficiency of the evolution process.

For example, in the \emph{compare-string-lengths} problem, the input arguments are of the type \code{String}, and the output is a \code{Bool} but allowing intermediate \code{Int} type. 
Looking at \autoref{tab:functions}, we can see that there are a few ways to convert a string to a boolean, as we only support functions in the character level. 
The best we can do is to extract the first character with \code{head} and then convert the character into a boolean with \code{isLetter} or \code{isDigit}. 
We could, for instance, generate a program that does that for both inputs and compares the results with different boolean operators. 
We could also apply a \code{map} function before applying \code{head}. 
Also, to convert a string into an integer given the type constraints for this problem, the only solution is to use the \code{length} function and then apply one of the few operators to compare two integers into a boolean. 
One example of obtained solution is \code{((length x1) > (length x0)) \&\& ((length x1) < (length x2))}.

On the other hand, for the \emph{last-index-of-zero} problem, a possible correct solution using our function-set is \code{fst (head (reverse (filter ($\lambda$y $\to$ 0 == (snd y)) (zip (range 0 (length x0) 1) x0))))}. 
So the synthesizer must first enumerate the input, apply a filter to keep only the elements that contain $0$, reverse the list, take the first element, and return its index. 
One of the best obtained solutions was \code{((length x0) + (if ((head (reverse x0)) == 0) then 1 else 0)) - 2} with $32\%$ of accuracy. 
It simply checks if the last element is $0$, if it is, it returns the length of the list minus one, otherwise it returns the length minus two. 
This is a possible general case for a recursive solution where it checks the last element and, if it is not zero, recurses with the remainder of the list.

\subsubsection{Analysis of the code refinements}

As described in ~\autoref{sec:code_refinements}, the code refinement step always produces an equal or better solution. 
These improvements are more noticeable on the \emph{median} and \emph{for-loop-index} problems.
This is due to the fact that code refinement is sometimes capable of discarding misused numerical constants. 
For example, one solution to the \emph{median} problem with $99\%$ of accuracy on the training set was \code{max -96 (min (max x1 x2) (max (min x1 x2) x0))} 
that only works if the median of the three arguments is greater than $-96$, otherwise it will always return a constant value. 
After the code refinements, \ac{HOTGP} finds the final and correct solution:
\code{min (max (min x2 x1) x0) (max x1 x2)}.

\begin{figure}[t]
    \centering
    \includegraphics[width=\linewidth]{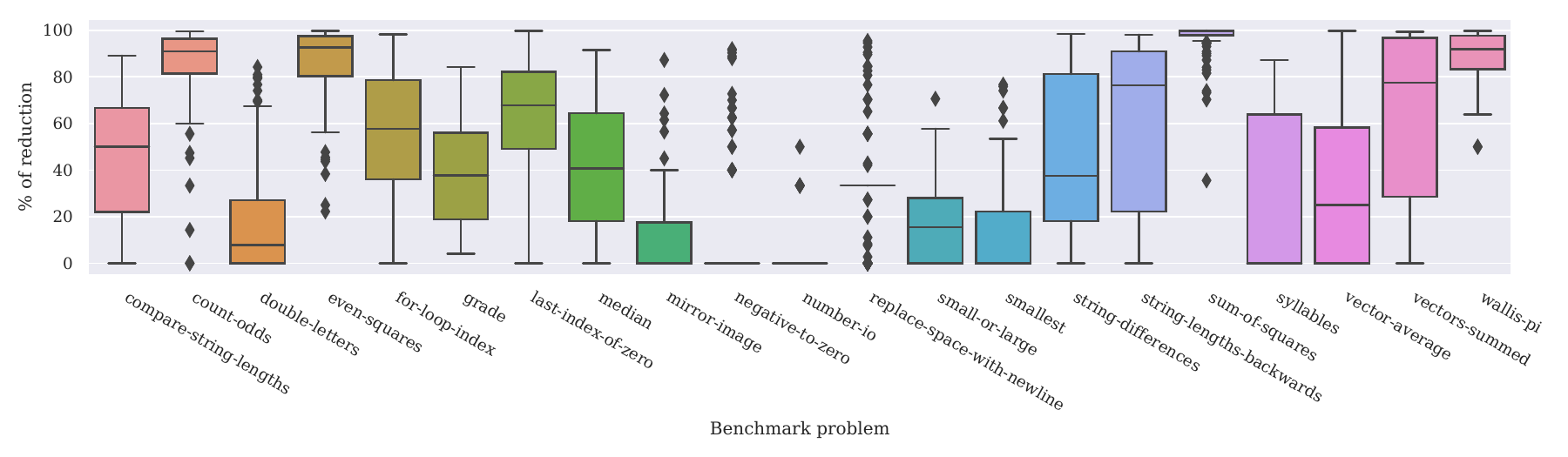}
    \caption{Percentage of reduction in the number of nodes caused by the refinement process.}
    \label{fig:refinement}
\end{figure}

Another benefit of code refinement is reducing the program size, which can improve the readability of the generated program. \autoref{fig:refinement} shows the rate of decrease in the program size after refinements, with a geometric mean of $52\%$.
The refinement process effectiveness varies depending on the nature of the solutions of the problem.
For most problems, the end of the upper quartile is within the $>75\%$ reduction mark, meaning it was not unusual for solutions to get largely simplified.
However, more evident results are yielded in problems such as \emph{counts-odds}, \emph{even-squares}, and \emph{sum-of-squares}, which dealt with fewer types (and thus a reduced function-set) and usually reached the maximum evaluation count, therefore were more susceptible to bloat. 
Notably, \emph{number-io} and \emph{negative-to-zero} had almost no reduction, showing the algorithm could directly find a perfect and near-minimal solution.

To provide further insights into how minimal the correct solutions actually are, and how susceptible to bloat each problem is, \autoref{tab:compare_to_manual} takes the smallest correct solution that \ac{HOTGP} could find for each problem, and compares them to the handwritten solutions crafted by the authors. Even before the refinement procedure, most of the solutions have the same node count as a the handwritten ones, and nearly all of them are reasonably close. The \emph{sum-of-squares} was initially much larger than the manual solution, but after refinement the size reduction is notable.
The only correct solution we found for \emph{mirror-image} also has the biggest reduction of the batch, showing a $87\%$ reduction overall.

\begin{table}[ht]
    \centering
    \caption{Node count of the hand-crafted solutions and the smallest correct solutions found by \ac{HOTGP} and \ac{HOTGP}*. The node count relative to hand-crafted is shown in parenthesis.}
    \label{tab:compare_to_manual}
    \stripedRows
    \begin{tabular}{lrrr}
    \toprule
    Benchmark & Hand-crafted & \ac{HOTGP} & \ac{HOTGP}* \\
    \midrule
    compare-string-lengths     &    $11$ &     11  ($1.0\times$) &      11 ($1.0\times$) \\
    count-odds                 &     $4$ &      4  ($1.0\times$) &       4 ($1.0\times$) \\
    for-loop-index             &     $7$ &     25  ($3.6\times$) &       9 ($1.3\times$) \\
    grade                      &    $27$ &     45  ($1.7\times$) &      29 ($1.1\times$) \\
    median                     &     $9$ &      9  ($1.0\times$) &       9 ($1.0\times$) \\
    mirror-image               &    $10$ &    102 ($10.2\times$) &      13 ($1.3\times$) \\
    negative-to-zero           &     $3$ &      3  ($1.0\times$) &       3 ($1.0\times$) \\
    number-io                  &     $4$ &      4  ($1.0\times$) &       4 ($1.0\times$) \\
    replace-space-with-newline &     $8$ &      8  ($1.0\times$) &       8 ($1.0\times$) \\
    small-or-large             &    $11$ &     12  ($1.1\times$) &      12 ($1.1\times$) \\
    smallest                   &     $7$ &      7  ($1.0\times$) &       7 ($1.0\times$) \\
    string-lengths-backwards   &     $4$ &      5  ($1.2\times$) &       5 ($1.2\times$) \\
    sum-of-squares             &     $7$ &    163 ($23.3\times$) &      30 ($4.3\times$) \\
    vector-average             &     $6$ &      6  ($1.0\times$) &       6 ($1.0\times$) \\
    vectors-summed             &     $5$ &      5  ($1.0\times$) &       5 ($1.0\times$) \\
    \bottomrule
    \end{tabular}
\end{table}




\section{Final Remarks}\label{sec:hotgp_outro}

This chapter presented \ac{HOTGP}, a \ac{GP} algorithm that supports \acp{HOF}, $\lambda$-functions, polymorphic types, and the use of type information to constrain the search space. 
It also sports a function-set based on the Haskell language using only pure functions in the non-terminals set.
Our main arguments in favor of this approach are: i) limiting our programs to pure functions avoids undesirable behaviors; ii) using type-level information and parametric polymorphism reduces the search space directing the \ac{GP} algorithm towards the correct solution; iii) \acp{HOF} eliminate the need of several imperative-style constructs (\emph{e.g.}, for loops).

\ac{HOTGP} differs from most \ac{GP} implementations as it actively uses the information of input and output types to constrain the candidate terminals and non-terminals while creating new solutions or modifying existing ones, and to select feasible points of recombination.

We have evaluated our approach with $29$ benchmark problems and compared the results with $8$ state-of-the-art algorithms from the literature. 
Overall, \ac{HOTGP} got favorable results, achieving high success rates in a wide variety of problems, being very competitive with the state of the art.

We also applied code refinements to the best solution found by the algorithm to reduce the occurrence of \emph{bloat} code. 
This procedure leads to further improvements in the results while at the same time improving the readability of the final program.

Even though we achieved competitive results, we observed that there are still possible improvements. 
First, our non-terminals set is much smaller than some of the state-of-the-art algorithms (\emph{e.g.}, PushGP). 
Future work includes carefully examining the impact of adding new functions to the function-set. 
This inclusion might further simplify the PS or allow us find solutions that are not currently being found. 
On the other hand, including new functions also expands the search space and can hinder some of our current results.

Our approach could also benefit from a more modular perspective for PS. 
In a modular approach, the problem is first divided into simpler tasks which are solved independently and then combined to create the complete synthesized program. 
This approach will require support to different forms of functional composition and the modification of the benchmark to create training data for the different subtasks. 
Such a synthesizer could also be coupled with Wingman\footnote{\href{https://haskellwingman.dev/}{https://haskellwingman.dev/}} (the current implementation of advanced Haskell code generation), which can either synthesize the whole program or guide the process using only the type information, and code holes.

Further research is also warranted concerning more advanced type-level information such as Generalized Algebraic Data Types (GADTs), Type Families, Refinement Types and Dependent Types.
More type information could further constrain the search space and, in some situations, provide additional hints to the synthesis of the correct program. 
Clearly, this must be accompanied by a modification of the current benchmarks and the inclusion of new benchmarks that provides this high-level information about the desired program.

%% file: tables/functions_smaller.tex
\begin{table}[t!]
    \centering
    \caption{Functions supported by HOTGP.}
    \label{tab:functions}
    \stripedRows
    \begin{tabular}{l>{\raggedright\arraybackslash}p{8cm}}
    \toprule
    \textbf{Function Type} & \textbf{Function names}\\
    \midrule

    \code{Int} $\to$ \code{Int} $\to$ \code{Int} &
        \code{addInt},
        \code{subInt},
        \code{multInt},
        \code{divInt},
        \code{modInt},
        \code{maxInt},
        \code{minInt}\\

    \code{Bool} $\to$ \code{Bool} & \code{not}\\

    \code{Bool} $\to$ \code{Bool} $\to$ \code{Bool} &
        \code{and},
        \code{or}\\

    \code{Bool} $\to$ \code{a} $\to$ \code{a} $\to$ \code{a} & \code{if}\\

    \code{Float} $\to$ \code{Float} & \code{sqrt}  \\

    \code{Float} $\to$ \code{Float} $\to$ \code{Float} &
        \code{addFloat},
        \code{subFloat},
        \code{multFloat},
        \code{divFloat}\\

    \code{a} $\to$ \code{[a]} & \code{singleton} \\
    \code{a} $\to$ \code{[a]} $\to$ \code{[a]} & \code{cons} \\
    \code{[a]} $\to$ \code{a} & \code{head} \\
    \code{[a]} $\to$ \code{[a]} & \code{reverse} \\
    \code{[[a]]} $\to$ \code{[a]} & \code{concat} \\

    \code{a} $\to$ \code{b} $\to$ \code{(a,b)} & \code{toPair} \\
    \code{(a,b)} $\to$ \code{a} & \code{fst} \\
    \code{(a,b)} $\to$ \code{b} & \code{snd} \\

    \code{Char} $\to$ \code{Char} $\to$ \code{Bool} & \code{eqChar} \\
    \code{Char} $\to$ \code{Bool} &
        \code{isLetter},
        \code{isDigit} \\

    \code{Int} $\to$ \code{Float} & \code{intToFloat} \\
    \code{Float} $\to$ \code{Int} & \code{floor} \\

    \code{Int} $\to$ \code{Int} $\to$ \code{Bool} &
        \code{gtInt},
        \code{ltInt},
        \code{eqInt} \\

    \code{[a]} $\to$ \code{Int} & \code{length} \\
    \code{Int} $\to$ \code{[a]} $\to$ \code{[a]} & \code{take} \\
    \code{Int} $\to$ \code{Int} $\to$ \code{Int} $\to$ \code{[Int]} & \code{range} \\
    \code{[Int]} $\to$ \code{Int} &
        \code{sumInts},
        \code{productInts} \\

    \code{[Float]} $\to$ \code{Float} &
        \code{sumFloats},
        \code{productFloats} \\

    \code{[[Char]]} $\to$ \code{[Char]} & \code{unlines} \\
    \code{Int} $\to$ \code{[Char]} & \code{showInt} \\
    \code{[a]} $\to$ \code{[b]} $\to$ \code{[(a,b)]} & \code{zip} \\
    \code{(a} $\to$ \code{b)} $\to$ \code{[a]} $\to$ \code{[b]} & \code{map} \\
    \code{(a} $\to$ \code{Bool)} $\to$ \code{[a]} $\to$ \code{[a]} & \code{filter} \\

    \bottomrule
    \end{tabular}
\end{table}

%% file: text/origami.tex
\chapter{Origami}
\label{cha:origami}

The previous chapter introduced \ac{HOTGP} as a \ac{GP} algorithm for synthesizing functional programs.
We demonstrated how \ac{FP} concepts can constrain the search space and guide the search in \ac{PS}, leading to competitive results when applied to a \ac{GP} algorithm.

However, most of the \ac{PS} algorithms so far have not effectively exploited common programming patterns.
Most algorithms, including \ac{HOTGP}, need to scan the search space to find a correct program without any guidance in terms of structure. 
While a less restrictive search space can be desirable to allow the algorithm to navigate towards one of the many solutions, a constrained search space, if correctly done, can speed up the search process allowing the search algorithm to focus only on part of the program.

A general and useful pattern is the use of \acfp{RS}~\citep{meijer1991functional}. 
This pattern captures the common structure of recursive functions and data structures as combinations of \emph{consumer} and \emph{producer} functions, also known as \emph{fold} and \emph{unfold}, respectively. 
They are known to be very general and capable of implementing many commonly used algorithms, ranging from data aggregation to sorting. 
\citet{gibbons2003fun} coined the name \emph{origami programming} and showed many examples of how to write common algorithms using these patterns.
The folding and unfolding process can be generalized through Recursion Schemes, which divide the programming task into three simpler steps: (\emph{i}) choosing the scheme (among a limited number of choices), (\emph{ii}) choosing a fixed point of a data structure that describes the recursion trace and, (\emph{iii}) writing the consumer (fold) and producer (unfold) procedure.

In this chapter we studied the problems in \ac{PSB1}~\citep{psb1} and solved them using \acp{RS}, reporting a selection of distinct solutions. 
This examination revealed that most of the solutions for the proposed problems follow the common pattern of folding, unfolding a data structure, or a composition of both.
With these observations, we explore the crafting of computer programs following these recursive patterns and craft program templates that follow these patterns along with an explanation of how they can constrain the search space of candidate programs. 
Our main goal is to find a set of recursion schemes that simplifies the \ac{PS} process while reducing the search space with type information. 
We also present the general idea on how to evolve such programs, called \emph{Origami}, a \ac{PS} algorithm that first determines the (un)folding pattern it will evolve and then evolves the corresponding template.

The remainder of this chapter is organized as follows. 
In \autoref{sec:recschemes}, we introduce recursion schemes and explain the basic concepts needed to understand our proposal. 
\autoref{sec:origami} presents Origami, a novel algorithm idea to evolve functional programs based on recursion schemes.
\autoref{sec:recursion_in_psb1} outlines some examples of recursive patterns that can be used to solve common programming problems and how Origami could evolve them.
In \autoref{sec:preliminary}, we show some proof-of-concept results adapting \ac{HOTGP} to one of the presented patterns and analyze the results. 
Finally, in \autoref{sec:origami_outro}, we give some final observations about Origami and describe future work.

We highlight that the contents of this chapter have been previously presented at the Genetic Programming Theory \& Practice (GPTP) workshop~\citep{origami}.
A considerable portion of the text presented here is an adaptation or extension of the text presented in that paper.

\section{Recursion Schemes}\label{sec:recschemes}

Recursive functions are sometimes referred to as the \emph{goto} of functional programming languages~\citep{jones2001expressive,harvey1992avoiding,meijer1991functional,gibbons2003fun}. 
They are an essential part of how to build a program in these languages, allowing programmers to avoid mutable state and other imperative programming constructs.
Both \emph{goto} and recursion are sometimes considered harmful when creating a computer program. 
The reason is that the former can make it hard to understand the program execution flow as the program grows larger, and the latter may lead to problems such as \emph{stack overflows} if proper care is not taken. 
This motivated the introduction of higher-order functions as the preferred alternative to direct recursion.

By using higher-order functions which apply the function received as input in a recursive pattern to a given value or data structure, we can represent many common recursion schemes. 
Some of the most well-known examples of these functions are \emph{map}, \emph{filter}, and \emph{fold}. 
The main advantage of using these general patterns is that the programmer does not need to be as careful (guaranteeing termination, or sane memory usage for example) as they would need to be using direct recursion.

Among these higher-order functions, \emph{fold} is the most general, as it can be used to implement the others. 
This pattern is capable of representing many recursive algorithms that start with a list of values and return a transformed value.
Another common general pattern is captured by the \emph{unfold} function that starts from a seed value and unfolds it, generating a list of values in the process.

While \emph{fold} and \emph{unfold} describe most common patterns of recursive functions, they are limited to recursions that follow a linear path (\emph{e.g.}, a list). 
In some scenarios, the recursion follows a nonlinear path such as a binary tree (\emph{e.g.}, most comparison-based sort algorithms). 
To generalize the \emph{fold} and \emph{unfold} operations to different recursive patterns, Recursion Schemes~\citep{garland1973program,meijer1991functional} describe common patterns of generating and consuming inductive data types, not limited to only the consumption or generation of data, but also abstracting the idea of having access to partial results or even backtracking. 
The main idea is that the recursion path is described by a fixed point of an inductive data type and the program-building task becomes limited to some specific definitions induced by the chosen data structure. 
This concept is explained in more detail in the following subsection.

\subsection{Folding lists and trees}

A generic inductive list (\emph{i.e.}, linked list) carrying values of type \code{a} is described in Haskell as:

\begin{minted}[escapeinside=@@]{haskell}
    data List a = Nil | Cons a (List a)
\end{minted}

This is read as ``a list can be either empty (\code{Nil}) or a combination (using \code{Cons}) of a value of type \code{a} followed by another list (\emph{i.e.}, its tail)''. 
In a similar fashion, we could recursively define a binary tree containing values of type \code{a} as:

\begin{minted}[escapeinside=@@]{haskell}
    data Tree a = Empty | Node (Tree a) a (Tree a) 
\end{minted}

From these, we could try to define our first recursion schemes.
With the intent of keeping these first examples simple, we will start with a loose and informal definition of the catamorphism, which will be defined more formally in the following sections.
To avoid confusion with the actual catamorphism definition and implementation explored in the rest of this chapter, in this section it will be simply called \code{fold}.

In the context of a list, we can define that \code{fold} recursively applies an accumulator function \code{f}, from the right.
We start with an initial value $s$ (from \emph{seed}), and use \code{f} to combine it with the last element of the list.
Then we use \code{f} again to combine this result with the penultimate element, and keep going until there are no more elements to combine.
Mathematically, it can be expressed as $f(x_0, f(x_1, f(x_2, \ldots f(x_{n - 1}, s)\ldots )$.
Or, in Haskell:
\begin{minted}{haskell}
    fold :: (a -> b -> b) -> b -> List a -> b
    fold _ s Nil = s
    fold f s (Cons x xs) = f x (fold f s xs)
\end{minted}

This simple pattern can be used to implement a wide variety of functions, such as:
\begin{minted}{haskell}
    sum xs = fold (+) 0 xs
    product xs = fold (*) 1 xs
    length xs = fold (\_ acc -> acc+1) 0 xs
    map f xs = fold (\x acc -> Cons (f x) acc) Nil xs
    concat xs = fold (++) Nil xs
\end{minted}

We can also apply a similar recursion scheme to the \code{Tree} structure.
In this context, \code{f} must accept the accumulated results of the left and right branches, and combine them with the actual value of that node.
\begin{minted}{haskell}
    foldTree :: (b -> a -> b -> b) -> b -> Tree a -> b
    foldTree _ s Empty = s
    foldTree f s (Node l x r) = f (foldTree f s l) x (foldTree f s r)
\end{minted}

Now, we could also define similar functions to the ones we defined for the \code{List} fold:
\begin{minted}{haskell}
    sumTree tree = foldTree (\l x r -> l+x+r) 0 tree
    productTree tree = foldTree (\l x r -> l*x*r) 1 tree
    size tree = foldTree (\l _ r -> l+1+r) 0 tree
    mapTree f tree = foldTree (\l x r -> Node l (f x) r) Empty tree
\end{minted}

While this would theoretically work, we needed to define a separate \code{fold} function for each of the structures we wanted to work with.
If we wanted to work with other recursion schemes, not only \code{fold}, we experience a combinatorial explosion, making it a very labor-intensive process.
Thus, a generic approach would be ideal: a \code{fold} function that automatically works with any new structure that gets implemented --- as well as structures that automatically work with any new scheme that we implement.

\subsection{Functors, Fixed points and Functor Algebras}

In order to make the recursion schemes generic, we start by noticing that both \code{List} and \code{Tree} are ``functors''.
Functors are type constructors \code{f :: Type -> Type} which support a \code{fmap} function, with the purpose of distributively applying a function to each of the nested elements of the Functor\footnote{The functor must also satisfy the two functor laws: 
\code{fmap id = id} and \code{fmap (f ◦ g) = fmap f ◦ fmap g}}.
This is expressed in Haskell by the following type class, and its instances for \code{List} and \code{Tree}:
\begin{minted}[escapeinside=@@]{haskell}
    class Functor f where 
        fmap :: (a -> b) -> f a -> f b
    instance Functor List where
        fmap _ Nil = Nil
        fmap f (Cons x xs) = Cons (f x) (fmap f xs)
    instance Functor Tree where
        fmap _ Empty = Empty
        fmap f (Node l x r) = Node (fmap f l) (f x) (fmap f r)
\end{minted}
\pagebreak
Given a functor \code{f}, we say that a type \code{p} is a fixed point if \code{p} is \emph{isomorphic} to \code{f p}.
Two types are isomorphic when there exist functions converting each to the other with neither loss of information nor change of structure: applying one after the other returns the original value.
We can represent this by the following data type and helper function:
\begin{minted}[escapeinside=@@]{haskell}
    data Fix f = MkFix (f (Fix f))

    unfix :: Fix f -> f (Fix f)
    unfix (MkFix x) = x
\end{minted}

\code{MkFix} is a \emph{value constructor} allowing us to create values of type \code{Fix}.
\code{unfix} is a helper function that extracts one layer of our nested structure. 
Together they witness the isomorphism mentioned above: applying \code{unfix} followed by \code{MkFix} (and vice-versa) returns the original value.
In other words, \code{MkFix~◦~unfix~=~id} and \code{unfix~◦~MkFix~=~id}, which means \code{Fix f} and \code{f (Fix f)} contain exactly the same information.

The \code{Fix} type allows us to write recursive programs for data structures with non-recursive definitions that are very similar to those targeting data structures with recursive definitions.
This becomes very useful to our purposes if we eliminate the recursive definition from the data structures by adding a second parameter:
\begin{minted}[escapeinside=@@]{haskell}
    data ListF a b = NilF | ConsF a b
    data TreeF a b = EmptyF | NodeF b a b

    -- In these functors, fmap  applies the function
    -- to the recursive structure, not to the elements.
    instance Functor (ListF a) where
        fmap :: (b -> c) -> ListF a b -> ListF a c
        fmap f NilF = NilF
        fmap f (ConsF a b) = ConsF a (f b)

    instance Functor (TreeF a) where
        fmap :: (b -> c) -> TreeF a b -> TreeF a c
        fmap f EmptyF = EmptyF
        fmap f (NodeF l a r) = NodeF (f l) a (f r)
\end{minted}

In combination with \code{Fix}, these definitions are isomorphic to \code{List} and \code{Tree}, in the sense that they still allow us to carry the same information.
Formally, we say that \code{Fix (ListF a)} is isomorphic to \code{List a}, by the means of the following morphisms:
\begin{minted}[escapeinside=@@]{haskell}
    fromList :: List a -> Fix (ListF a)
    fromList Nil = MkFix NilF
    fromList (Cons x xs) = MkFix (ConsF x (fromList xs))
    
    toList :: Fix (ListF a) -> List a
    toList (MkFix NilF) = Nil
    toList (MkFix (ConsF x xs)) = Cons x (toList xs)
\end{minted}

A similar reasoning could be applied to evidence the isomorphism between \code{Tree} and \code{Fix TreeF} with the functions \code{fromTree} and \code{toTree}.
Notice that \code{ListF} and \code{TreeF} are functors over \code{b}, which means that \code{fmap} will not apply the function on the elements, but on the tail of the list.

Finally, \ac{RS} arise from two categorical constructions: algebras and coalgebras.
Given an endofunctor $f$, an $f$-algebra is a pair $(a, h)$ where $a$ is the carrier object and $h : f a \to a$ is the structure map.
The carrier $a$ is the domain in which the functor’s shape is interpreted, and the structure map specifies how one $f$-structured layer is collapsed into a single value.
Operationally, this corresponds to the mechanism underlying folds.

Dually, an $f$-coalgebra is a pair $(a, h)$ with $h : a \to f a$.
Here the carrier $a$ provides the space of seeds from which data will be generated, and the structure map specifies how a seed expands into one $f$-structured layer.
This is the mechanism underlying unfolds.

These constructions supply the abstract basis for recursion schemes: algebras determine how structures are consumed; coalgebras determine how they are produced.
With these definitions, we have all we need to properly define our first recursion scheme.


\subsection{An example Recursion Scheme: catamorphism}

The application of an algebra into a fixed point structure is called \emph{catamorphism}:
\begin{minted}[escapeinside=@@]{haskell}
    cata :: Functor f => (f a -> a) -> Fix f -> a
    cata alg xs = alg (fmap (cata alg) (unfix xs))
\end{minted}
\[
\begin{tikzcd}
\code{Fix f} \arrow[rr, "\code{cata alg}"] \arrow[d, "\code{unfix}"'] 
& {}
& \code{a} \\
\code{f (Fix f)} \arrow[rr, "\code{fmap (cata alg)}"] 
& {}
& \code{f a}  \arrow[u, "\code{alg}"']
\end{tikzcd}
\]

Given an algebra \code{alg}, \code{cata} \emph{peels} the outer layer of the fixed-point \code{data}, maps itself to the nested structure, and applies the algebra to the result. 
In short, the procedure traverses the structure to its deepest layer and applies \code{alg} recursively accumulating the result.
In this sense, \code{cata} is similar to the \code{foldr} function for lists.

Notice that \code{cata} works with any functor, as long as we can create a fixed-point from it, which is true for the \code{ListF} and \code{TreeF} data structures.

As an example, we can implement the \code{sum} function for lists and trees in terms of \code{cata}:
\begin{minted}[escapeinside=@@]{haskell}
    sum :: List Int -> Int
    sum ls = cata sumAlg (fromList ls) where
        -- starting value
        sumAlg NilF = 0
        -- combine the value x to the accumulator
        sumAlf (ConsF x acc) = x + acc
   
    sumTree :: Tree Int -> Int
    sumTree t = cata sumAlg (fromTree t) where
        -- starting value
        sumAlg EmptyF = 0
        -- combine the value x to the accumulators
        sumAlg (NodeF lAcc x rAcc) = x + lAcc + rAcc
\end{minted}

Although the general idea of defining the fixed-point of a data structure and implementing the catamorphism may look like over-complicating standard functions, the end result allows us to focus on much simpler implementations. 
In the special case of a list, we just need to specify the neutral element (\code{NilF}) and how to combine two elements (\code{ConsF~x~y}). 
All the inner mechanics of how the whole list is reduced is abstracted away in the \code{cata} function.




\subsection{Well-known Recursion Schemes}

While catamorphism is one of the simplest and most common \acp{RS}, there are less frequent patterns that hold some useful properties when building recursive programs. 
The most well-known recursion schemes (including the ones already mentioned) are:

\begin{itemize}
    \item \textbf{catamorphism / anamorphism:} also known as folding and unfolding, respectively. The catamorphism aggregates the information stored in the inductive type. The anamorphism generates an inductive type starting from a seed value;
    \item \textbf{accumulation:} this is similar to catamorphism by reducing a structure, but allowing an extra accumulating parameter;
    \item \textbf{paramorphism / apomorphism:} these Recursion Schemes work as catamorphism and anamorphism. However, at every step they allow access to the original downwards structure; 
    \item \textbf{histomorphism / futumorphism:} histomorphism allows access to every previously consumed element from the most to the least recent and futumorphism allows access to the elements that are yet to be generated.
\end{itemize}

And, of course, we can also combine these morphisms creating the \emph{hylomorphism} (anamorphism followed by catamorphism), \emph{metamorphism} (catamorphism followed by anamorphism), and chronomorphism (combination of futumorphism and histomorphism).

In the following, section we will explain some possible ideas on how to exploit these patterns in the context of program synthesis, and show different examples of programs developed using this pattern.

\section{Origami}\label{sec:origami}

The main idea of Origami is to reduce the search space by breaking down the synthesis process into smaller steps.
An overview of the proposed approach is outlined in \autoref{alg:origami}.

\begin{algorithm}[htb]
\caption{Origami Program Synthesis}\label{alg:origami}

    \SetKwFunction{origami}{origami}%
    \SetKwProg{Fn}{\textbf{function}}{\string:}{}%

    \DontPrintSemicolon

    \Fn{\origami{x, y, types}}{
        \DataSty{recScheme} $\gets \operatorname{pickRecursionScheme(types)}$\; \label{line:step1}
        \DataSty{indTypes} $\gets \operatorname{pickInductiveType(types)}$\; \label{line:step2}
        \DataSty{template} $\gets \operatorname{pickTemplate(recScheme, indTypes, types)}$\; \label{line:step3}
        \DataSty{fitness} $\gets \operatorname{createFitnessFunction(template, indTypes)}$\; \label{line:step4}
        \Return $\operatorname{evolveProgramFromTemplate(template, fitness)}$\; \label{line:step5}
    }
\end{algorithm}

In \autoref{line:step1} of the algorithm we determine the recursion scheme of the program. 
Since there are just a few known morphisms and the distribution of use cases for each morphism is highly skewed, some options (from the most naive to more advanced ones) are:

\begin{itemize}
    \item Do the search heuristically (\emph{e.g.}, following a flowchart\footnote{\url{https://hackage.haskell.org/package/recursion-schemes-5.2.2.4/docs/docs/flowchart.svg}})
    \item Run multiple searches in parallel with each one of the templates
    \item Integrate this decision as part of the search (\emph{e.g.}, encode into the chromosome)
    \item Use the type information to pre-select a subset of the templates (see. \autoref{tab:type-to-rec})
    \item Use the description of the problem together with a language model
\end{itemize}

\begin{table}[htb]
    \centering
        \caption{Association between type signatures and its corresponding recursion schemes.}
    \stripedRows
    \begin{tabular}{c|c}
       \toprule
       \textbf{Type signature}  & \textbf{Recursion Scheme} \\
       \midrule
        \code{f a $\to$ b} & catamorphism, accumulation \\
        \code{a $\to$ f b} & anamorphism \\
        \code{a $\to$ b} & hylomorphism \\
        \bottomrule
    \end{tabular}
    \label{tab:type-to-rec}
\end{table}

Specifically to the use of type information, as we can see in \autoref{tab:type-to-rec}, the type signature can constrain the possible recursion schemes, thus reducing the search space of this choice. There are also some specific patterns in the description of the program that can help us choose one of the templates. For example, whenever the problem requires returning the position of an element, we should use accumulation.

After the choice of recursion scheme, in \autoref{line:step2} we choose the appropriate base (inductive) data type. 
The most common choices are natural numbers, lists, and rose trees. 
Besides these choices, one could provide custom data structures if needed. 
This choice could be done employing the same same methods used in \autoref{line:step1}.

\autoref{line:step3} deals with the choice of which specific template of evolvable functions (further explained in the next sections) will specify the parts of the program that must be evolved returning a template function to be filled by the evolutionary process. 
Once this is done, we can build the fitness function (\autoref{line:step4}) that will receive the evolved functions, wrap them into the recursion scheme, and evaluate them using the training data. 
Finally, we run the evolution (\autoref{line:step5}) to find the correct program.

To illustrate the process, let us go through the process to generate a solution to the problem \emph{count-odds} from \ac{PSB1}~\citep{psb1}:


\begin{quote}
\textbf{Count Odds} Given a vector of integers, return the number of integers that are odd, without use of a specific even or odd instruction (but allowing instructions such as modulo and quotient).
\end{quote}

We can start by determining the type signature of this function:

\begin{minted}[escapeinside=@@]{haskell}
    countOdds :: [Int] -> Int
\end{minted}

\begin{enumerate}
    \item[\autoref{line:step1}] As the type signature suggests, we are collapsing a list of values into a value of the same type. So, we should pick one of the \emph{catamorphism} variants. Let us pick the plain catamorphism.
    \item[\autoref{line:step2}] In this step we need to choose a base inductive type. Since the type information tells us we are working with lists, we can use the list functor.
    \item[\autoref{line:step3}] As the specific template we choose the reduction to a value (from a list of integers to a single value).
\end{enumerate}

We are now at this point of the code generation where we depart from the following template:

\begin{minted}[escapeinside=@@]{haskell}
    countOdds :: [Int] -> Int
    countOdds ys = cata alg (fromList ys)

    alg :: ListF Int Int -> Int
    alg xs = case xs of
               NilF -> @\evolv{e1}@
               ConsF x y -> @\evolv{e2}@
\end{minted}

\begin{enumerate}
    \item[\autoref{line:step4}] We still need to fill up the \emph{gaps} \code{e1} and \code{e2} in the code. At this point, the piece of code \code{NilF -> e1} can only evolve to a constant value as it must return an integer and it does not have any integer available in scope. The piece of code \code{ConsF x y -> e2} can only evolve to operations that involve \code{x}, \code{y}, and integer constants.
    \item[\autoref{line:step5}] Finally, the evolution can be run. As the evolution of the functions inside the recursion scheme is well determined by the input-output types of the main function, the final solution should be:

\begin{minted}[escapeinside=@@]{haskell}
    countOdds :: [Int] -> Int
    countOdds xs = cata alg xs

    alg :: ListF Int Int -> Int
    alg xs = case xs of
               NilF -> 0
               ConsF x y -> mod x 2 + y
\end{minted}
\end{enumerate}

\section{Recursion Schemes in PSB1}\label{sec:recursion_in_psb1}

With the purpose of motivating the use of \acp{RS} for \ac{PS} in a practical scenario, we conducted an investigation of \ac{PSB1}~\citep{psb1} in order to find recurrent patterns and recursion schemes.
To identify the different templates that can appear, we manually solved the entire \ac{PSB1} in such a way that the evolvable \emph{gaps} of the recursion schemes have a well determined and concrete function type. 
In this section, we present and analyse each of the patterns we found.
Particularly, we will highlight one example of each template, but the entire set of solutions is available in the appendix (\autoref{cha:origami_solutions}). 
In the following examples, the evolvable parts of the solution are shown in underlined green, making it more evident the number and size of programs that must be evolved by the main algorithm.

It should be noted that we made some concessions in the way some programs were solved. 
In particular, our solutions are only concerned with returning the required values and disregard any IO operations (for instance, \emph{print the result with a string "The results is"}) as we do not see the relevancy in evolving this part of the program at this point. 
This will be part of the full algorithm for a fair comparison with the current state-of-the-art.

\subsection{Catamorphism}

In the previous sections, we gave the definition of a catamorphism. 
Notice that all of the following solutions follow the same main form \code{cata alg (fromList data)}, which changes the input argument into a fixed form of a list and apply the algebra of the catamorphism.
Specifically for the catamorphism, we observed four different templates that we will exemplify in the following, from the simplest to the more complicated approaches. In what follows, we will only present the definition of the \code{alg} function.

\subsubsection{Reducing a structure}

The most common use case of catamorphism is to reduce a structure to a single value, or \code{f a -> a}. 
In this case, the algebra follows a simple function that is applied to each element and combined with the accumulated value. 
This template was already illustrated in the beginning of \autoref{sec:origami} with the example of \emph{countOdds}:

\begin{quote}
\textbf{Count Odds} Given a vector of integers, return the number of integers that are odd, without use of a specific even or odd instruction (but allowing instructions such as modulo and quotient).
\end{quote}

\begin{minted}[escapeinside=@@]{haskell}
    -- required primitives: constant int, mod, +
    alg :: ListF Int Int -> Int
    alg NilF = @\evolv{0}@
    alg (ConsF x acc) = @\evolv{mod x 2 + acc}@
\end{minted}

\begin{quote}
\textbf{Evolvable expressions:} given a function of type \code{[a] -> b}, we need to evolve:
\begin{enumerate}[label=\roman*)]
    \item \code{alg NilF}: an expression of type \code{b};
    \item \code{alg (ConsF x acc)}: an expression of type \code{b}, with the values \code{x :: a} and \code{acc :: b} in scope.
\end{enumerate}
\end{quote}

\subsubsection{Regenerating the structure: mapping}

The higher-order function \code{map} is a \code{fold} that processes and reassembles the structure.
Any function that can be implemented using a map can also be implemented through catamorphism.

\begin{quote}
\textbf{Double Letters} Given a string, print (in our case, return) the string, doubling every letter character, and tripling every exclamation point. All other non-alphabetic and non-exclamation characters should be printed a single time each.
\end{quote}
\begin{minted}[escapeinside=@@]{haskell}
    -- required primitives: if-then-else, (<>), ([])
    -- user provided: constant '!', constant "!!!", isLetter
    alg NilF         = @\evolv{[]}@
    alg (ConsF x xs) = @\evolv{if x == \chLit!}@
                          @\evolv{then \stLit{!!!} <> xs}@
                          @\evolv{else if isLetter x then [x,x] <> xs}@
                                          @\evolv{else x:xs}@

\end{minted}

The main difference from the previous example is that, in this program, at every step the intermediate result (\code{xs}) is a list. Notice that the \code{ConsF} case is still constrained in such a way that we can either insert the character \code{x} somewhere in \code{xs}, or change \code{x} into a string and concatenate to the result.

\begin{quote}
\textbf{Evolvable expressions:} given a function of type \code{[a] -> [b]}, we need to evolve:
\begin{enumerate}[label=\roman*)]
    \item \code{alg NilF}: an expression of type \code{[b]};
    \item \code{alg (ConsF x xs)}: an expression of type \code{[b]}, with the values \code{x :: a} and \code{xs :: [b]} in scope.
\end{enumerate}
\end{quote}

\subsubsection{Generating a function}

Another possible application of catamorphism is in functions that return a function, or in Haskell notation \code{f a -> f a -> b}, which is read as: a function that takes two arguments of type \code{f a} (\emph{e.g.}, a list of values) and returns a value of type \code{b}. This signature is equivalent to its curried form which is \code{f a -> (f a -> b)}: a function that takes a value of type \code{f a} and returns a function that takes an \code{f a} and returns a value of \code{b}. While generating a function that returns a function seems to add complexity, the type constraints can help guiding the synthesis more efficiently than if we were to interpret it as a function of two arguments.

\begin{quote}
\textbf{Super Anagrams} Given strings x and y of lowercase letters, return true if y is a super anagram of x, which is the case if every character in x is in y. To be true, y may contain extra characters, but must have at least as many copies of each character as x does.
\end{quote}

\begin{minted}[escapeinside=@@]{haskell}
    -- required primitives: delete, constant bool
    -- elem, (&&)
    alg NilF = \ys -> @\evolv{True}@
    alg (ConsF x f) = \ys -> @\evolv{elem x ys \&\& f (delete x ys)}@
\end{minted}

For the base case, the empty string (which will also be the end of the first string), we assume that this is a super anagram returning \code{True}.
For the second pattern, we must remember that \code{xs} is supposed to be a function that receives a list and returns a boolean value. 
So, we first check that the second argument is not null, that \code{x} is contained in \code{ys}, and then evaluate \code{xs} passing \code{ys} after removing the first occurrence of \code{x}.
Notice that for the \code{NilF} case we are not limited to returning a constant value, we can apply any function to the second argument that returns a boolean. 
Thus, any function \code{String -> Bool} will work. 
Even though we have more possibilities for the base case, we can grow the tree carefully to achieve a proper solution. 
The same goes for the second case in which we add more possible programs as we have in our possession a char value, a string and a function that process a string.

\begin{quote}
\textbf{Evolvable expressions:} given a function of type \code{[a] -> [a] -> b}, we need to evolve:
\begin{enumerate}[label=\roman*)]
    \item \code{alg NilF}: an expression of type \code{b} with the value \code{ys :: [a]} in scope;
    \item \code{alg (ConsF x xs)}: an expression of type \code{b}, with the values \code{x :: a}, \code{ys :: [a]} and \code{f :: [a] -> b} in scope.
\end{enumerate}
\end{quote}

\subsubsection{Combination of patterns}

More complex programs often combine two or more different tasks represented as functions that return tuples (\code{f a -> (b, c)}).
If both tasks are independent and are implemented with a catamorphism, they are equivalent to applying different functions in every element of the $n$-tuple. 
The evolution process would be the same as per the previous template but we would evolve one function for each output type.

\subsection{Anamorphism}

\emph{Anamorphism} can be seen as the dual of catamorphism, in the sense that it is derivated from a coalgebra:
\begin{minted}[escapeinside=@@]{haskell}
    ana :: Functor f => (a -> f a) -> a -> Fix f
    ana coalg seed = MkFix (fmap (ana coalg) (coalg seed))
\end{minted}
\[
\begin{tikzcd}
\code{a} \arrow[rr, "\code{ana coalg}"] \arrow[d, "\code{coalg}"'] 
& {}
& \code{Fix f} \\
\code{f\,a} \arrow[rr, "\code{fmap (ana coalg)}"] 
& {}
& \code{f (Fix f)} \arrow[u, "\code{MkFix}"']
\end{tikzcd}
\]

In this function, we first apply \code{coalg} to \code{seed}, generating a structure of type \code{f} (usually a singleton), then we map the \code{ana coalg} function to the just generated data to further expand it, finally we enclose it inside a \code{Fix} structure to obtain the fixed point. 
In the context of lists, this procedure is known as \code{unfold} as it departs from one value and unrolls it into a list of values.

Its evolutionary template is composed of the \code{coalg} function, which can either generate a new value and continue the recursion; or choose to terminate the structure, effectively providing a base-case, which ends the recursion. 
The most common case is when there is a predicate that handles deciding when to follow each path: one returning a terminator, and the other the data structure containing one element and one seed value.
For this recursion scheme we only identified a single template in which the first argument is used as the initial seed and any remaining argument (if of the same type) is used as a constant when building the program. 
Of course, during the program synthesis, we may test any permutation of the use of the input arguments.

\begin{quote}
\textbf{For Loop Index} Given 3 integer inputs start, end, and step, print the integers in the sequence $n_0$ = start, $n_i$ = $n_i-1$ + step for each $n_i$ < end, each on their own line.
\end{quote}

\begin{minted}[escapeinside=@@]{haskell}
    -- required primitives: (==), (+)
    forLoopIndex :: Int -> Int -> Int -> [Int]
    forLoopIndex start end step = toList (ana coalg @\evolv{start}@)
      where
        coalg seed = case @\evolv{seed == end}@ of
                       True -> NilF
                       False -> ConsF seed @\evolv{(seed + step)}@
\end{minted}

In this program, the \code{start} argument is the starting seed of the anamorphism and the \code{step} and \code{end} are used when defining each case. 
The \code{case} predicate must evolve a function that takes the seed as an argument and returns a boolean. 
To evolve such a function we are limited to the logical and comparison operators. 
As the type of the seed is well determined, we must compare it with values of the same type, which can be constants or one of the remaining arguments. 
After that, we must evolve two programs, one that creates the element out of the seed (a function of type \code{Int -> Int}) and the generation of the next seed.

\begin{quote}
\textbf{Evolvable expressions:} given a function of type \code{a\textsubscript1 -> a\textsubscript2 -> $\ldots$ -> a\textsubscript n -> [b]}, we need to evolve:
\begin{enumerate}[label=\roman*)] 
\item initial seed: an expression of type \code{b}, with all the arguments in scope;
\item predicate: and expression of type \code{Bool}, with \code{seed :: b} and all the arguments in scope;
\item next seed: an expression of type \code{b}, with \code{seed :: b} and all the arguments in scope;
\end{enumerate}
\end{quote}

\subsection{Hylomorphism}

Hylomorphism is the fusion of both catamorphism and anamorphism:
\begin{minted}{haskell}
    hylo :: Functor f => (f b -> b) -> (a -> f a) -> a -> b
    hylo alg coalg = alg . fmap (cata alg . ana coalg) . coalg
\end{minted}
\[
\begin{tikzcd}[column sep=6em]
\code{a} \arrow[rr, "\code{hylo alg coalg}"] \arrow[d, "\code{coalg}"'] 
& {}
& \code{b} \\
\code{f\,a} \arrow[r, "\code{fmap (ana coalg)}"] 
\arrow[rr, bend right=15, "\code{fmap (cata alg . ana coalg)}"'] 
& \code{f (Fix f)} \arrow[r, "\code{fmap (cata alg)}"] 
& \code{f\,b} \arrow[u, "\code{alg}"']
\end{tikzcd}
\]

Given both an algebra and a coalgabra, \code{hylo} uses an initial seed to generate a structure with the coalgebra, and immediately consumes it with the algebra.
This template works the same as evolving the functions for both schemes.

\begin{quote}
\textbf{Collatz Numbers} Given an integer, find the number of terms in the Collatz (hailstone) sequence starting from that integer.
\end{quote}

\begin{minted}[escapeinside=@@]{haskell}
    -- required primitives: constant int, (==)
    -- (+), (*), mod, div
    alg NilF = @\evolv{1}@
    alg (ConsF x acc) = @\evolv{1 + acc}@
    
    coalg x = case @\evolv{x == 1}@ of
                True -> NilF
                False -> ConsF x @\evolv{(if mod x 2 == 0}@
                                  @\evolv{then div x 2 }@
                                  @\evolv{else div (3*x + 1) 2)}@
\end{minted}

The single input argument is used as the initial seed.
In the coalgebra, if the current seed is equal to $1$ the process terminates, otherwise we generate the next hailstone number.
The algebra in this case is simply the \code{length+1}, counting the number of generated values and offsetting it by $1$ in the \code{NilF} case, to account for the value $1$ that was dropped during the anamorphism.

\begin{quote}
\textbf{Evolvable expressions:} given a function of type \code{a -> b}, we need to evolve:
\begin{enumerate}[label=\roman*)] 
\item \code{alg NilF}: an expression of type \code{b};
\item \code{alg (ConsF x xs)}: an expression of type \code{b}, with the values \code{x :: a} and \code{acc :: b} in scope.
\item \code{coalg} predicate: and expression of type \code{Bool}, with \code{x :: b} in scope;
\item \code{coalg} next seed: an expression of type \code{b}, with \code{x :: b} in scope;
\end{enumerate}
\end{quote}

\subsection{Accumulation}

In some situations our solution needs to accumulate the results and gain access to the partial results. 
For this purpose we can implement \emph{accumulation}, which requires an algebra and an accumulator strategy \code{st}.
In this diagram, we use the notation \code{st $\square$ p} as a shorthand for \code{\textbackslash x -> st x p}.
\begin{minted}{haskell}
    accu :: Functor f => (forall x. f x -> p -> f (x, p)) ->
        (f a -> p -> a) -> Fix f -> p -> a
    accu st alg xs p = alg (fmap (uncurry (accu st alg)) (st (unfix xs) p)) p
\end{minted}
\[
\begin{tikzcd}[column sep=10em]
\code{Fix f} \arrow[d, "\code{unfix}"'] \arrow[rr, "\code{accu st alg $\square$ p}"] &                                                    & \code{a                      }  \\
\code{f (Fix f)} \arrow[r, "\code{st $\square$ p}"]                           & \code{f (Fix f, p)} \arrow[r, "\code{fmap (accu st alg) $\square$ p}"] & \code{f a} \arrow[u, "\code{alg $\square$ p}"']
\end{tikzcd}
\]

This pattern first traverses the structure from the top, using the accumulator strategy \code{st} to annotate each element of type \code{a} with some additional information: the accumulating parameter of type \code{p}.
Then, it uses the algebra \code{f a -> p -> a} in from bottom to top, just like catamorphism, but allowing access to the accumulating parameter \code{p} at that level.
This template must be carefully used because it can add an additional degree of freedom through \code{p}, since this can be of any type, not limited by any of the main program types, thus expanding the search space. 
To avoid such a problem, we will use accumulations in very specific use-cases as described in the following sub-sections.

\subsubsection{Indexing data}

Whenever the problem requires the indexing of the data structure, we can use the accumulating parameter to store the index of each value of the structure and, afterwards, use this information to process the final solution. 
With this template, the parameter should be of type \code{Int}.

\begin{quote}
\textbf{Last Index of Zero} Given a vector of integers, at least one of which is 0, return the index of the last occurrence of 0 in the vector.
\end{quote}

\begin{minted}[escapeinside=@@]{haskell}
    -- required primitives: if-then-else, (+), (==)
    -- (<>), constant int, Maybe, Last
    lastIndexZero :: [Int] -> Int
    lastIndexZero xs = accu st alg (fromList xs) @\evolv{0}@
      where
        st NilF s = NilF
        st (ConsF x xs) s = ConsF x (xs, @\evolv{s+1}@)

        alg NilF i = @\evolv{\negative1}@
        alg (ConsF x acc) i = @\evolv{if x == 0 \&\& acc == -1}@
                               @\evolv{then i}@
                               @\evolv{else acc}@
\end{minted}

The accumulator strategy of this program has the purpose of indexing our list, starting from 0 and incrementing it by 1 at every step.
When the list is indexed, we build the result from the bottom up by signaling that we have not found a zero by initially returning \code{-1}. 
Whenever \code{x == 0} and the currently stored index is \code{-1}, the program returns the index stored in that level (\code{s}). 
Otherwise, it just returns the current \code{acc}.

\begin{quote}
    
\textbf{Evolvable expressions:} given a function of type \code{[a] -> b}, we need to evolve:
\begin{enumerate}[label=\roman*)] 
\item initial parameter: an expression of type \code{Int};
\item updated parameter: an expression of type \code{Int}, with \code{s :: Int} in scope;
\item \code{alg NilF s}: an expression of type \code{b}, with \code{i :: Int} in scope;
\item \code{alg (ConsF x xs)}: an expression of type \code{b}, with \code{i :: Int}, \code{x :: a} and \code{acc :: b} in scope;
\end{enumerate}
\end{quote}

Naturally, this pattern only requires accumulation as a means to build an indexed list.
If we allow our solution to have an indexed list structure in the first place, this pattern is trivially transformed into a catamorphism:

\begin{minted}[escapeinside=@@]{haskell}
    data IListF a b = INilF | IConsF Int a b deriving Functor
    toIndexedList :: [a] -> Fix (IListF a)
    toIndexedList xs = go 0 xs where
        go _ [] = MkFix INilF
        go i (x:xs) = MkFix (IConsF i x (go (i+1) xs))
            
    -- required primitives: if-then-else, (+), (==)
    -- (<>), constant int, Maybe, Last
    lastIndexZero :: [Int] -> Int
    lastIndexZero xs = cata alg (toIndexedList xs)
      where
        alg INilF = @\evolv{\negative1}@
        alg (IConsF i x acc) = @\evolv{if x == 0 \&\& acc == -1}@
                               @\evolv{then i}@
                               @\evolv{else acc}@
\end{minted}

\subsubsection{A combination of catamorphisms}

In some cases the recursive function is equivalent to the processing of two or more catamorphisms, with a post-processing step that combines the results. 
If all the operations involved are associative, this can be expressed in terms of an accumulation, as it doesn't matter whether we perform the computation top to bottom or bottom to top.

A simple example is the average of the values of a vector, in which we need to sum the values and count the length of the vector, combining both final results with the division operator. 
This template of accumulation constrains the type of the accumulator to a tuple of the returning type of the program.

\begin{quote}
\textbf{Vector Average} Given a vector of floats, return the average of those floats. Results are rounded to 4 decimal places.
\end{quote}

\begin{minted}[escapeinside=@@]{haskell}
    -- required primitives: (+), (/)
    vecAvg :: [Double] -> Double
    vecAvg xs = accu st alg (fromList xs) @\evolv{(0.0, 0.0)}@
      where
        st NilF _ = NilF
        st (ConsF x xs) (s1, s2) = ConsF x (xs, (@\evolv{s1 + x, s2 + 1)}@)

        alg NilF (s1, s2) = @\evolv{s1 / s2}@
        alg (ConsF x acc) s = acc
\end{minted}
We illustrate this solution by splitting the accumulator expression into two distinct expression, one for each element of the tuple. 
While \code{s1 + x} accumulates the sum of the values of the list, \code{s2 + 1} increments the accumulator by one at every step. 
In this template, the final solution is the combination of the values at the final state of the accumulator, thus in the \code{alg} function we just need to evolve a function of the elements of the state.

\begin{quote}
    
\textbf{Evolvable expression:} given a function of type \code{[a] -> b}, we need to evolve:
\begin{enumerate}[label=\roman*)] 
\item initial value: an expression of type \code{(b, b)};
\item updated seeds: an expression of type \code{(b, b)}, with \code{x :: a}, \code{xs :: [a]}, \code{s1 :: b} and \code{s2 :: b} in scope;
\item post-processing: an expression of type \code{b}, with \code{s1 :: b} and \code{s2 :: b} in scope.
\end{enumerate}
\end{quote}

Naturally, this could also be converted into two catamorphisms if we allow for a post-processing step of type \code{b -> b -> b} (assuming that the reducing operation is associative):

\begin{minted}[escapeinside=@@]{haskell}
    -- required primitives: (+), (/)
    vecAvg :: [Double] -> Double
    vecAvg xs = f (cata alg1 (fromList xs)) (cata alg2 (fromList xs))
      where
        f s1 s2 = s1 / s2

        alg1 NilF = 0.0
        alg1 (ConsF x acc) = acc + x
        
        alg2 NilF = 0.0
        alg2 (ConsF x acc) = acc + 1
\end{minted}

\subsection{Summarizing PSB1}

\begin{figure}[htb]
    \centering
    \input{images/plots/dist_morphisms}
    \caption{Distribution of recursion schemes used to solve the full set of PSB1 problems.}
    \label{fig:dist-morphisms}
\end{figure}
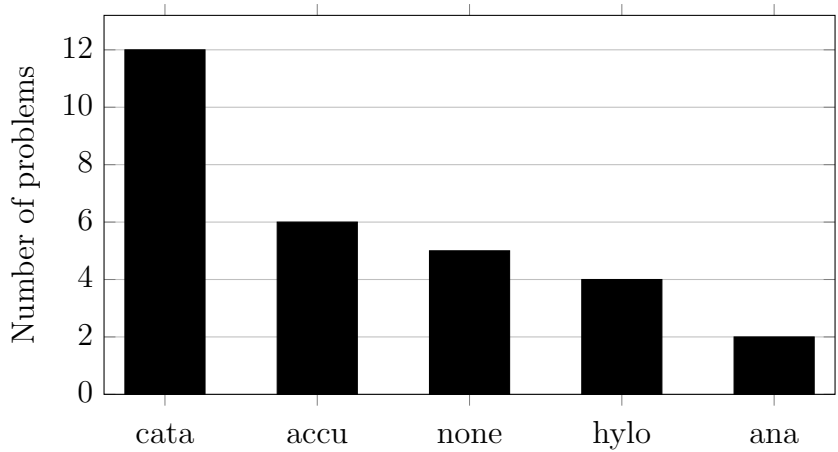

From the $29$ problems considered here and implemented by a human programmer, $17\%$ were trivial enough and did not require any recursion scheme; $41\%$ were solved using catamorphism; $20\%$ of used accumulation (although we were required to constrain the accumulator function). 
Anamorphism accounted for only $7\%$ of the problems and hylomorphism for $14\%$. 
The distribution of the usage of each recursion scheme is shown in \autoref{fig:dist-morphisms}.

In the next section we report a simple experiment with a subset of these problems as a proof-of-concept of our approach.

\section{Origami Proof-of-Concept Results}\label{sec:preliminary}

The main objective of this chapter is to introduce the ideas of using recursion schemes to solve programming challenges and to verify whether the current benchmark problems can be solved using this approach. 
In this section, we will show how using the catamorphism template can help improve the overall performance of a \ac{GP} approach. 

For this purpose, we adapted \ac{HOTGP} to generate only the necessary \emph{evolvable} part of a catamorphism (in here, implemented as a \code{foldr}) and tested most of the benchmarks that can be solved by this specific template. 
Specifically, we asked \ac{HOTGP} to generate an expression for the \code{alg (ConsF x xs)} pattern. 
For the \code{alg NilF}, we used default \emph{empty} values depending on the data-type: $0$ for \code{Int} and \code{Float}, \code{False} for \code{Bool}, the space character for \code{Char}, and empty lists for lists and strings. 
Naturally, this is a simplification of the template, as all of the benchmarks we are interested in happen to use these values for the null pattern. 
In order to properly generate the recursion patterns, this part of the function should also be considered in the evolution.

We set the maximum depth of the tree to 5, as the expressions we want to generate are always smaller than that. This also has an effect in keeping the generated programs more readable. All the other parameters use the same values described in the previous chapter.
To position Origami within the current literature, we compare the obtained results against the same algorithms \ac{HOTGP} was compared to in \autoref{sec:hotgp_results}.

Furthermore, we removed the benchmarks that would need the algorithm to output a function (Mirror Image, Vectors Summed, and Grade), as this is not currently supported by \ac{HOTGP}.
For completeness, we also tested the benchmarks that can be solved by accumulations (even though this adaptation does not support it) to show that once committed to a template (\emph{e.g.}, catamorphism), the algorithm cannot find the correct solution if it requires a different template (\emph{e.g.}, accumulation).

\autoref{tab:functionset} shows the function set we used for the presented experimental evaluation, while \autoref{tab:userset} lists the set of functions and constants we assume that should be provided by the user, as they are contained in the problem description. 
Some of them can be replaced by \code{case-of} instructions (e.g., \code{isVowel, isLetter, scrabbleScore}), which can increase the difficulty of obtaining a solution. 

\begin{table}[t]
    \centering
    \caption{Function set used for solving the PSB1 benchmark problems. }
    \stripedRows
    \begin{tabular}{rp{\dimexpr 0.75\linewidth-2\tabcolsep}}
       \toprule
       Type class  & Functions \\
       \midrule
        Numbers & \code{fromIntegral, +, -, *, /, \^{}} \code{div, quot, mod, rem} \code{abs, min, max}\\
        Logical & \code{<, <=, >, >=, ==, /=, \&\&, not, ||} \\
        Lists & \code{cons, snoc, <>, head, tail, init, last, null, length, delete, elem}\\
        Tuple & \code{fst, snd} \\
        Map & \code{findMap, insertWith} \\
        General purpose & \code{if-then-else, case, uncurry, fromEnum, toEnum, id} \\
        \bottomrule
    \end{tabular}
    \label{tab:functionset}
\end{table}

\begin{table}[t]
    \centering
    \caption{Function and constant set assumed to be provided by the user. All of these are explicitly mentioned in the problem description.}
    \stripedRows
    \begin{tabular}{rp{\dimexpr 0.75\linewidth-2\tabcolsep}}
       \toprule
       Type  & Functions/Constants \\
       \midrule
        \code{Int -> Bool} & \code{(< 1000), (>= 2000)}\\
        \code{String} & \code{"small", "large", "!!!", "ABCDF", "ay"} \\
        \code{Char} & \code{'!', ' ', '\textbackslash n'} \\
        \code{Int} & 0, 1, 64 \\
        \code{Char -> Bool} & \code{isVowel, isLetter} \\
        \bottomrule
    \end{tabular}
    \label{tab:userset}
\end{table}

\afterpage{
\begin{landscape}
\begin{table}[p]
    \centering
    \caption{Percentage of runs that returned a perfect solution on the validation set. The bottom part of the table summarizes the result as the number of times each algorithm had the highest percentage, and in how many problems the percentage was greater or equal to a certain threshold. The benchmarks marked with $*$ are only solvable with accumulation.}
\stripedRows
\begin{tabular}{lrrrrrrrrrr}
    \toprule
    Benchmark & Origami & \ac{HOTGP} & \ac{DSLS} & \ac{UMAD} & PushGP & \ac{G3P} & \ac{CBGP} & \ac{G3P+} & \ac{G3P}hs & \ac{G3P}py \\
    \midrule
    checksum*& 0 & -- & 1 & \bestResult{5} & 0 & 0 & -- & 0 & -- & -- \\
    count-odds & \bestResult{100} & 50 & 11 & 12 & 8 & 12 & 0 & 3 & -- & -- \\
    double-letters & \bestResult{94} & 0 & 50 & 20 & 6 & 0 & -- & 0 & -- & -- \\
    last-index-of-zero* & 0 & 0 & \bestResult{62} & 56 & 21 & 22 & 10 & 44 & 0 & 2 \\
    negative-to-zero & \bestResult{100} & \bestResult{100} & 82 & 82 & 45 & 63 & 99 & 13 & 0 & 66 \\
    replace-space-with-newline & 60 & 38 & \bestResult{100} & 87 & 51 & 0 & 0 & 16 & -- & -- \\
    scrabble-score & \bestResult{100} & -- & 31 & 20 & 2 & 2 & -- & 1 & -- & -- \\
    string-lengths-backwards & \bestResult{100} & 89 & 95 & 86 & 66 & 68 & -- & 18 & 0 & 34 \\
    syllables & \bestResult{84} & 0 & 64 & 48 & 18 & 0 & -- & 39 & -- & -- \\
    vector-average* & 0 & 80 & 88 & \bestResult{92} & 16 & 5 & 88 & 0 & 4 & 0 \\
\midrule
\textbf{\# of Best Results} & 6 & 1 & 2 & 2 & 0 & 0 & 0 & 0 & 0 \\
\midrule
    $\mathbf{=100}$ & \bestResult{4} & 1 & 1 & 0 & 0 & 0 & 0 & 0 & 0 & 0 \\
    $\mathbf{\geq75}$ & \bestResult{6} & 3 & 4 & 4 & 0 & 0 & 2 & 0 & 0 & 0 \\
    $\mathbf{\geq50}$ & \bestResult{7} & 4 & \bestResult{7} & 5 & 2 & 2 & 2 & 0 & 0 & 1 \\
    $\mathbf{>0}$ & 7 & 5 & \bestResult{10} & \bestResult{10} & 9 & 6 & 3 & 7 & 1 & 3 \\
    \bottomrule
    \end{tabular} 
    \label{tab:origami_results}
\end{table}

\end{landscape}
}

Analyzing the results depicted in \autoref{tab:origami_results}, one can notice that when comparing the standard \ac{HOTGP} with Origami, the latter always obtains an equal or better number of perfect runs, except on the accumulation benchmarks. Not only that, but the number of problems that are always solved increased from $1$ to $4$, and those higher than $75\%$ increased from $3$ to $6$, a significant improvement in success rate. When compared to related work, out of the $7$ solvable benchmarks, Origami had the best results in $6$ of them. The only exception being the \emph{replace-space-with-newline}. Overall, once we choose the correct template, the synthesis step becomes simpler.

\section{Final Remarks}\label{sec:origami_outro}

Our main hypothesis with this chapter is that, by starting the program synthesis fixing one of the recursion schemes, we simplify the process of program synthesis. 
For this purpose we used a general set of benchmarks widely used in the literature. 
Within this benchmark suite, we observed that, in most cases, the evolvable part of the programs becomes much simpler, to the point of being trivial. 
However, some of them require a pre-processing of the input arguments with some general use functions (such as \code{zip}) to keep this simplicity, or an adaptation to the output type as to return a function instead of a value. 

In some cases, more complicated functions can be evolved with the help of a human interaction by asking additional information such as \emph{when should the recursion stop?}. 
Also, in many cases, the type signature of each one of the evolvable programs already constrains the search space. 
For example, the pattern \code{alg NilF} must return a value of the return type of the program without using any additional information, thus the space is constrained to constant values of the return type.

Analysing the minimal function set required to solve all these problems, one can formulate a basic idea about the adequate choice based on the signature of the main function and on any user-provided type/function. 

Also,  every problem in the benchmark could be solved by just a handful of patterns.
As the problems become more difficult and other patterns emerge, we can resort to more advanced recursions such as \emph{dynomorphism} when dealing with dynamic programming problems, for example. 
Also, none of these problems required a recursive pattern with a base structure different from a list. 
In the future, we plan to test other benchmarks and introduce new ones that require different structures to test our approach.

One challenge to this approach is how to treat the templates containing multiple evolvable parts. 
For example, anamorphism requires the evolution of three expressions: one that generates the next element, one to generate the next seed, and one predicate to check for the stop condition. 
We will consider a multi-gene approach~\citep{searson2010gptips} or a collaborative co-evolution strategy~\citep{soule2008improving,grefenstette1996methods}.

As a final consideration, we highlight the fact that most of the programs can be further simplified if we annotate the output type with monoids. 
In functional programming, and Haskell in particular, monoids are a class of types that have an identity value (\code{mempty}) and a binary operator (\code{<>}) such that \code{mempty <> a = a <> mempty = a}. 
With these definitions we can replace many of the functions and constants described in \autoref{tab:functionset}~and~\autoref{tab:userset} with \code{mempty} and \code{<>}, reducing the search space.

%% file: images/plots/dist_morphisms.tex
\begin{tikzpicture}
\begin{axis}[
    ybar,
	bar width=30pt,
    width=.7\linewidth,
    legend cell align={left},
	legend style={
            /tikz/nodes={anchor=base},
            at={(0.03,0.94)},
            anchor={north west}
        },
    ylabel={Number of problems},
    symbolic x coords={
        cata,
        accu,
        none,
        hylo,
        ana,
    },
    xtick=data,
    ymin=0,
    ymajorgrids,
    typeset ticklabels with strut,
    height=16em,
    /pgf/number format/.cd,
        use comma,
        1000 sep={},
    xticklabel style = {text width=2.25cm,align=center},
    ]
\addplot[fill=black] plot coordinates {
    (cata, 12)
    (accu, 6)
    (ana,2)
    (hylo,4)
    (none,5)
};
\end{axis}
\end{tikzpicture}

%% file: text/bananas.tex
\chapter{Going Bananas!}
\label{cha:bananas}

\autoref{cha:origami} introduced Origami as a technique for synthesizing programs by leveraging common recursive patterns as described by \acp{RS}. 
We showed that, for \ac{PSB1}~\citep{psb1}, each problem can be fully solved choosing one of four \acp{RS} (cata, accu, ana, or hylo).
However, these results were simply a proof-of-concept, and preliminary experiments were conducted as a way to show the feasibility of the algorithm. 

In this chapter, we present the first full implementation of the Origami algorithm.
This implementation aims to evaluate Origami in its complete form against \ac{PSB1}, while also enabling potential improvements as well experiments in other benchmarks, in the future.

The remainder of this chapter is organized as follows.
\autoref{sec:bananas_methods} presents and discusses implementation details of Origami.
In \autoref{sec:bananas_results}, we report and analyze the experimental results of this implementation in \ac{PSB1}, comparing it to the state of the art.
Finally, \autoref{sec:bananas_outro} presents closing thoughts, highlighting the main developments of this chapter and discussing current limitations.

We highlight that the contents of this chapter have been previously presented at the $34^{\mathit{th}}$ Brazilian Conference on Intelligent Systems (BRACIS)~\citep{bananas}.
A considerable portion of the text presented here is an adaptation or extension of the text presented in that paper.

\section{Origami Program Synthesis}
\label{sec:bananas_methods}

Origami's implementation follows a Koza-style Genetic Programming~\citep{koza1992genetic} (tree representation). The main distinctions to traditional approaches are the introduction of immutable nodes (ensuring a certain Recursion Scheme); and the type-safety of the genetic operators using the same approach taken by \ac{HOTGP}.

The implementation is based on \emph{patterns}, which are used to represent different \acp{RS}.
A pattern is composed of immutable nodes and a set of evolvable slots that, when replaced with expressions, can be evaluated.
The immutable nodes describe the main definition of the \ac{RS} (see \autoref{sec:bananas_patterns}) and are fixed once we choose the pattern, while the evolvable slots represent the inner mechanisms that need to be synthesized to correspond to the expected behavior described by the dataset.
These slots have a well-defined output type (inferred from the problem description), and a well-defined set of bindings to which the expression has access.

In this chapter we focus on the six different patterns that comprehend the minimal set required to solve \ac{PSB1}.
All the manual solutions are publicly available on \href{https://github.com/folivetti/origami-programming/}{GitHub}\footnote{\label{fnote:github}\href{https://github.com/folivetti/origami-programming/}{https://github.com/folivetti/origami-programming/}}, and are referred to as \emph{canonical} solutions in the remainder of this text.
When a pattern is used by the canonical solution to solve a problem, we also refer to it as the \emph{canonical pattern} for that problem.
Naturally, Origami is not limited to these patterns, and more could be included as needed.
The next section details these six patterns.

\subsection{Patterns}
\label{sec:bananas_patterns}

\newcommand{\slot}[1]{\textcolor{blue}%
{\code{\textit{\underline{slot\textsubscript{#1}}}}}}

\newcommand{\argVal}[1]{%
\code{arg{\textsubscript{#1}}}%
}

\newcommand{\argType}[1]{%
\code{i{\textsubscript{#1}}}%
}

\newcommand{\scopeitem}[2]{
\item \code{#1}~::~\code{#2}
}

\newenvironment{scope}
{%
    with scope \{\begin{itemize*}[
    itemjoin={;},
    label={} 
    ]
}
{%
    \end{itemize*}%
    \}%
}

\newcommand{\argScope}{
    \item \argVal{0}~::~\argType{0}
    $\ldots$
    \argVal{n}~::~\argType{n}
}

\subsubsection{NoScheme}

This is the simplest pattern in Origami, as it does not employ any recursion at all.
It is represented by the following code:

\begin{minted}[escapeinside=@@]{haskell}
f @\argVal{0}@ ... @\argVal{n}@ = @\slot{1}@
\end{minted}

This pattern has just a single slot, which has all the arguments in scope and returns a value of the same type as the output of the program.
Its main use is to accommodate for problems that do not require any recursion.

\subsubsection{Catamorphism over Indexed List}

This pattern captures the most common \ac{RS} observed in \ac{PSB1}, and arguably in practical scenarios as well, \ie folding a list from the right. 
This pattern is commonly used in Haskell as \code{foldr}.
In Meijer-notation~\citep{meijer1991functional}, this would be represented by the banana brackets $\llparenthesis b, \oplus \rrparenthesis$, where $b$ is the initial value and $\oplus$ is the combining function.
In the context of Origami, it can be represented as:

\begin{minted}[escapeinside=@@]{haskell}
f @\argVal{0}@ ... @\argVal{n}@ = cata alg @\argVal{0}@ where
      alg INil = @\slot{1}@
      alg (ICons i x acc) = @\slot{2}@
\end{minted}

In a problem with arguments of type \code{\argType{0} $\ldots$ \argType{n}} and of output type \code{o}, where \code{\argType{0}~$\equiv$~[e]}%
\footnote{The notation \code{\argType{0}~$\equiv$~[e]} is a restriction such that \code{\argType{0}} can be decomposed into the type \code{[e]}, which is the type of a list with elements of some type \code{e}.}%
, this pattern's slots are typed as follows:

\begin{itemize}
    \item \code{\slot{1}~::~o}, with nothing in scope;
    \item \code{\slot{2}~::~o},
    \begin{scope}
        \scopeitem{i}{Int}
        \scopeitem{x}{e}
        \scopeitem{acc}{o}
        \argScope
    \end{scope}.
\end{itemize}

This pattern will be referred simply as \emph{Cata} in the remainder of this thesis.

\subsubsection{Curried Catamorphism over Indexed List}
This pattern captures a common variation of the Catamorphism, and can be represented by the following code:

\begin{minted}[escapeinside=@@]{haskell}
f @\argVal{0}@ @\argVal{1}@ = cata alg @\argVal{0}@ @\argVal{1}@ where
      alg INil = \ys -> @\slot{1}@
      alg (ICons i x f) = \ys -> @\slot{2}@
\end{minted}

As a problem of type \code{\argType{0}~->~\argType{1}~->~o} can also be seen in its curried form as \code{\argType{0}~->~(\argType{1}~->~o)}, we can employ Catamorphism to accumulate a \emph{function} over the first argument, and then apply this function to the second argument.
This is useful, as mentioned in \autoref{cha:origami}, when we need to apply a Catamorphism over the zip of two lists.

In a problem with arguments of type \code{\argType{0}, \argType{1}}\footnote{Notice that this pattern can only be applied to problems with exactly two arguments.} and of output type \code{o}, where \code{\argType{0}~$\equiv$~[e]}, this pattern's slots are typed as follows:

\begin{itemize}
    \item \code{\slot{1}~::~o},
    \begin{scope}
        \scopeitem{ys}{\argType{1}}
    \end{scope};
    
    \item \code{\slot{2}~::~o}, 
    \begin{scope}
        \scopeitem{i}{Int}
        \scopeitem{x}{e}
        \scopeitem{f}{\argType{1}~->~o}
        \scopeitem{ys}{\argType{1}}
    \end{scope}.
\end{itemize}

For brevity, this will be referred to as simply \emph{CurriedCata}.
Note that both this and the previous pattern use Indexed Linked List as the data structure, allowing the program to access each element's index and value.
For the remaining patterns we employ a regular list since it is enough to solve their problems (as shown by the canonical solutions presented in~\autoref{cha:origami_solutions}).

\subsubsection{Anamorphism to a List}
This pattern is commonly used in Haskell as \code{unfold}, which is used to generate a list. In Meijer-notation~\citep{meijer1991functional}, this would be represented by the concave lenses $\lbparen g, p\rbparen$
where $g$ is the generator function, and $p$ is the predicate. In the context of Origami, it can be represented by the following code:

\begin{minted}[escapeinside=@@]{haskell}
f @\argVal{0}@ ... @\argVal{n}@ = ana coalg @\slot{1}@ where
      coalg seed = if @\slot{2}@ then [] 
                   else @\slot{3}@ : @\slot{4}@
\end{minted}

In a problem with arguments of type \code{\argType{0} $\ldots$ \argType{n}} and of output type \code{o}, where \code{o~$\equiv$~[e]}, this pattern's slots are typed as follows:

\begin{itemize}
    \item \code{\slot{1}~::~\argType{0}}, 
    \begin{scope}
        \argScope
    \end{scope};
    
    \item \code{\slot{2}~::~Bool}, 
    \begin{scope}
        \scopeitem{seed}{\argType{0}}
        \argScope
    \end{scope};
    
    \item \code{\slot{3}~::~e},
    \begin{scope}
        \scopeitem{seed}{\argType{0}}
        \argScope 
    \end{scope};

    \item \code{\slot{4}~::~\argType{0}},
    \begin{scope}
        \scopeitem{seed}{\argType{0}}
    \end{scope}.
\end{itemize}

Note that while we do not enforce \code{\argVal{0}} to be used in \slot{1}, it must be of the same type as \code{\argVal{0}}, as all of the solutions for \ac{PSB1} respected this constraint.
For brevity, this will be referred to as simply \emph{Ana} in the rest of this thesis.

\subsubsection{Accumulation over a List}
This pattern captures using an accumulation strategy before using a \code{foldr}, and can be represented by the following code:

\begin{minted}[escapeinside=@@]{haskell}
f @\argVal{0}@ ... @\argVal{n}@ = accu st alg @\argVal{0}@ @\slot{1}@
      where
        st [] s = []
        st (x : xs) s = x : (xs, @\slot{2}@)
        alg [] s = @\slot{3}@
        alg (x : acc) s = @\slot{4}@
\end{minted}

In a problem with arguments of type \code{\argType{0} $\ldots$ \argType{n}} and of output type \code{o}, where \code{\argType{0}~$\equiv$~[e]}, \textbf{and given a type \code{a}}, this pattern's slots are typed as follows:

\begin{itemize}
    \item \code{\slot{1}~::~a}, 
    \begin{scope}\argScope\end{scope};
    
    \item \code{\slot{2}~::~a},
    \begin{scope}
        \scopeitem{x}{e}
        \scopeitem{xs}{[e]}
        \scopeitem{s}{a}
        \argScope
    \end{scope};
    
    \item \code{\slot{3}~::~o},
    \begin{scope}
        \scopeitem{s}{a}
        \argScope
    \end{scope};

    \item \code{\slot{4}~::~o},
    \begin{scope}
        \scopeitem{x}{e}
        \scopeitem{acc}{o}
        \scopeitem{s}{a}
        \argScope
    \end{scope}.
\end{itemize}

This is the first pattern whose types are not fully determined by the type of the arguments and the expected output type: the accumulator type \code{a}.
Types such as this will be referred to as \emph{unbound types}.
To keep the implementation simple, we assume unbound types are known and provided by the user.
The exploration of different types is an interesting challenge that warrants dedicated research.
This pattern will be referred to as simply \emph{Accu} in the rest of this thesis.

\subsubsection{Hylomorphism through a List}
This pattern captures an Anamorphism followed by a Catamorphism, such as applying \code{foldr} to the result of \code{unfold}, in Haskell.
In Meijer-notation~\citep{meijer1991functional}, this would be represented by the envelopes $\left\llbracket (c, \oplus), (g, p)\right\rrbracket$.
In Origami, it is represented by the following code:

\begin{minted}[escapeinside=@@]{haskell}
f @\argVal{0}@ ... @\argVal{n}@ = hylo alg coalg @\argVal{0}@ where
      coalg seed = if @\slot{1}@ then [] else @\slot{2}@ : @\slot{3}@
      alg [] = @\slot{4}@
      alg (x : acc) = @\slot{5}@
\end{minted}

In a problem with arguments of type \code{\argType{0} $\ldots$ \argType{n}} and of output type \code{o}, \textbf{and given a type \code{a}}, this pattern's slots are typed as follows:

\begin{itemize}
    \item \code{\slot{1}~::~Bool},
    \begin{scope}
        \scopeitem{seed}{\argType{0}}
        \argScope
    \end{scope};
    
    \item \code{\slot{2}~::~a},
    \begin{scope}
        \scopeitem{seed}{\argType{0}}
        \argScope 
    \end{scope};
    
    \item \code{\slot{3}~::~\argType{0}},
    \begin{scope}
        \scopeitem{seed}{\argType{0}}
        \argScope 
    \end{scope};

    \item \code{\slot{4}~::~o}, with nothing in scope;
    
    \item \code{\slot{5}~::~o}, 
    \begin{scope}
        \scopeitem{x}{a}
        \scopeitem{acc}{o}
        \argScope 
    \end{scope}.
\end{itemize}

This pattern also contains an unbound type: the intermediary list has elements of type \code{a}.
This pattern will be referred to as \emph{Hylo}.

\subsection{Genetic Programming}

Origami synthesizes the evolvable slots using \iac{GP}~\citep{koza1992genetic} algorithm. 
Since the patterns require more than a single slot, we represent each solution as a collection of programs represented as expression trees, in a multi-gene representation~\citep{searson2010gptips}.
Each element of this collection corresponds to one of the slots.

The \ac{GP} starts with an initial random population of $1\,000$ individuals, and iterates by applying either crossover to a pair of parents, or mutation to a single parent, generating $1\,000$ new individuals in total.
The entire population is replaced by the offspring population.

The initial population is generated using a \emph{ramped half-and-half} method, where half of the individuals are generated using the \emph{full} method and half using the \emph{grow} method.
The maximum depth for each method varies between $1$ and $5$.
The parental selection is performed using a tournament selection of size $10$.

Following a simple \ac{GP} algorithm, in Origami the mutation randomly selects one of the evolvable slots, then picks one point in the tree at random to be replaced by a new subtree generated at random using the grow method, with a maximum depth of $5 - d_{\mathit{current}}$.
Crossover also starts by picking one of the slots at random, then performing one of these two actions with equal probability: 
i) swap the entire slot of one parent with the same slot of the other parent; 
ii) swap two subtrees of the same output type from each parent.

\subsection{Set of functions}

In \ac{HOTGP} (\autoref{cha:hotgp}), the selected functions focused on providing a minimal set of operations, including \acp{HOF}, that would enable the synthesis of programs under the functional programming paradigm.

With Origami, however, the main focus is assessing \acp{RS} as the only means of synthesizing recursive programs. 
Therefore, we designed our function-set to avoid implicitly recursive functions, like \code{map}, \code{filter}, \code{sum}, and \code{product}.
We acknowledge that this might remove shortcuts and potentially make the synthesis of certain problems harder. 
Notice that the recursion happens in the immutable nodes that describe the \acp{RS}, so the recursion is provided rather than evolved.
Additionally, the set of operations includes functions equivalent to those used by other methods, in particular those implemented by PushGP~\citep{helmuth2018program}. 
As a result, Origami has a larger set of operations than \ac{HOTGP}.
The full set of functions is presented in~\autoref{tab:grammar}.
Once an execution is finished, the champion's slots are refined using the same procedure that was used in \ac{HOTGP} (\autoref{sec:code_refinements}, sans the hand-written simplification rules).
To refine a tree, we pick the root node and check if replacing it with any of its children leads to a correctly-typed solution with an equal or better fitness.
If so, we replace it with the best child; otherwise, we keep the original node.
This process continues recursively, traversing the tree and greedily replacing nodes with their children when needed.
This procedure applies Occam's Razor to choose a simpler solution~\citep{helmuth2017improving}, making sure the fitness in the training set is never worse.

\section{Experimental Results}
\label{sec:bananas_results}

To evaluate our approach we conducted experiments to perform an automatic search for different patterns in the \ac{PSB1}~\citep{psb1} context.
For each of the $29$ datasets, we sequentially tried each pattern in increasing order of complexity: NoScheme; Cata, if \code{\argVal{0}} is a list; CurriedCata, if the problem has two arguments and \code{\argVal{0}} is a list; Ana, if the return type is a list; Accu, if \code{\argVal{0}} is a list; Hylo.

\begin{table}[t!]
    \caption{The complete set of functions available for Origami. Each dataset only had access to the functions that involved its allowed types according to~\citet{psb1}.}
    \label{tab:grammar}
    \centering 
    \scriptsize
\stripedRows
    \begin{tabular}{p{.45\linewidth}|p{.45\linewidth}}
    \toprule
        Operations & Types \\
        \midrule
        
\code{add, sub, mult, div, quot, mod, rem, min, max} & \code{Int -> Int -> Int} \\
\code{abs, succ, pred } & \code{Int -> Int} \\
\code{add, sub, mult, div, min, max} & \code{Float -> Float -> Float} \\
\code{abs, sqrt, sin, cos, succ, pred} & \code{Float -> Float} \\
\code{fromIntegral} & \code{Int -> Float} \\
\code{floor, ceiling, round} & \code{Float -> Int} \\
\code{lt, gt, gte, lte} & \code{Int -> Int -> Bool} \\
\code{lt, gt, gte, lte} & \code{Float -> Float -> Bool} \\
\midrule
\code{and, or} & \code{Bool -> Bool -> Bool} \\
\code{not} & \code{Bool -> Bool} \\
\code{if} & \code{Bool -> a -> a -> a} \\
\code{eq, neq} & \code{a -> a -> Bool} \\
\midrule
\code{show} & \code{a -> [Char]} \\
\code{charToInt} & \code{Char -> Int} \\
\code{intToChar} & \code{Int -> Char} \\
\code{isLetter, isSpace, isDigit} & \code{Char -> Bool} \\
\midrule
\code{length} & \code{[a] -> Int} \\
\code{cons, snoc} & \code{a -> [a] -> [a]} \\
\code{mappend} & \code{[a] -> [a] -> [a]} \\
\code{elem} & \code{a -> [a] -> Bool} \\
\code{delete} & \code{a -> [a] -> [a]} \\
\code{null} & \code{[a] -> Bool} \\
\code{head, last} & \code{[a] -> a} \\
\code{tail, init} & \code{[a] -> [a]} \\
\code{zip} & \code{[a] -> [b] -> [(a, b)]} \\
\code{replicate} & \code{Int -> a -> [a]} \\
\code{enumFromThenTo} & \code{Int -> Int -> Int -> [Int]} \\
\code{reverse} & \code{[a] -> [a]} \\
\code{splitAt} & \code{Int -> [a] -> ([a], [a])} \\
\code{intercalate} & \code{[a] -> [a] -> [a]} \\
\midrule
\code{fst} & \code{(a, b) -> a} \\
\code{snd} & \code{(a, b) -> b} \\
\code{mkPair} & \code{a -> b -> (a, b)} \\
\midrule
\code{apply} & \code{(a -> b) -> a -> b} \\
\midrule
\code{singleton} & \code{a -> b -> Map a b} \\
\code{insert} & \code{a -> b -> Map a b -> Map a b} \\
\code{insertWith} & \code{((b, b) -> b) -> a -> b -> Map a b -> Map a b} \\
\code{fromList} & \code{[(a, b)] -> Map a b} \\
         \bottomrule
    \end{tabular}
\end{table}

For each dataset, we executed $30$ seeds of each pattern starting from the simplest and testing other patterns if none of the seeds succeeded in finding a solution (\ie the success rate was $0\%$).
Each seed followed the instructions provided by \ac{PSB1}, using the recommended number of training and test instances, and included the fixed edge cases in the training data, as well as using the fitness functions described in~\citep{psb1}.
We also made the same adaptations to the benchmarks as in \autoref{cha:hotgp} and \autoref{cha:origami}, which are similar to the ones made by \citet{copilot}. 
Specifically, we changed the input of the \grade benchmark from \code{(\argVal{0}, \argVal{1}, \argVal{2}, \argVal{3}, \argVal{4})} to \code{([(\argVal{0}, 'A'), (\argVal{1}, 'B'), (\argVal{2}, 'C'), (\argVal{3}, 'D')], \argVal{4})}; and, since we only generate pure programs, we adapted the output to return the results instead of printing them on: \checksum, \digits, \evenSquares, \forLoopIndex, \grade, \pigLatin, \replaceSpaceWithNewline, \stringDifferences, \syllables, and \wordStats.

Note that we deliberately placed the patterns with unbound types at the end of the sequence.
Therefore, the unbound type in both Accu and Hylo is only decided after all other patterns have failed.
For the benchmarks that Origami failed to find a solution with the other patterns, we applied one of these two patterns choosing the type that was known to be correct according to the canonical solutions.
For the cases in which the canonical solutions did not use Accu or Hylo,  we chose a reasonable type as needed (see~\autoref{tab:unbound_types}).

\begin{table}[b]
    \centering
    \caption{The chosen types for the unbound types in Accu and Hylo. The type is colored in blue when the decision was guided by the canonical solution.}
    \label{tab:unbound_types}
\stripedRows
    \begin{tabular}{lll}
    \toprule
       \textbf{Dataset} & Accu & Hylo  \\
       \midrule
        \checksum & \canonical\code{Int} & \code{Int} \\
        \collatzNumbers & -- & \canonical\code{Int} \\
        \digits & -- & \code{Int} \\
        \pigLatin & -- & \code{[Char]} \\
        \stringDifferences & \code{Int} & \code{(Char, Char)} \\
        \sumOfSquares & -- & \canonical\code{Int} \\
        \vectorAverage & \canonical\code{(Float, Int)} & \code{Int} \\
        \wallisPi & -- & \canonical\code{Float} \\
        \wordStats & \canonical\code{((Int, Int), (Int, Int))} & \code{[Char]} \\
        \xWordLines & \canonical\code{Int} & \code{[Char]} \\
         \bottomrule
    \end{tabular}
\end{table}

The maximum tree depth was set to $5$ for each slot. As Origami is based in \ac{HOTGP}, which was empirically shown to be robust to changes to the crossover rate, we set it to the same value as \ac{HOTGP} ($50\%$).
We allowed a maximum of $300\,000$ evaluations with an early stop whenever the algorithm finds a perfectly accurate solution according to the training data.
For patterns in which termination is not guaranteed, namely Ana and Hylo, a maximum number of iterations was imposed (empirically set to $10\,000$). 
Non-termination is also an issue that can occur with  CurriedCata. Specifically, Origami was synthesizing solutions with the slot \code{alg (Cons i x f) = \textbackslash ys -> f (f ys)}, essentially creating a ``fork bomb''.
To tackle this, a maximum execution budget is enforced when the evaluation of a single iteration of a slot executes more than $10\,000$ operations. In this case, the program is assigned an infinitely bad fitness. This limit was reached by less than $0.5\%$ of the individuals.

\autoref{tab:results_by_pattern} shows the percentage of executions in which Origami was able to synthesize a solution that completely solved the test set (\ie success rate).
Origami found a solution for all of the problems that were canonically solved by NoScheme as well as Cata.
Surprisingly, it was also able to synthesize a solution for 
\forLoopIndex by using NoScheme, even though the canonical solution used Ana, and for \grade by using Cata when the canonical solution used CurriedCata.
Nonetheless, we also ran these problems with their canonical patterns and discovered Origami was also able to synthesize solutions, albeit less often.
Moreover, Origami was able to find the solutions for $3$ out of the $4$ canonical CurriedCata problems, and $2$ out of the $3$ Ana problems.
Accu and Hylo, however, appear to be the most difficult patterns to synthesize, as no solution for problems that canonically involve these patterns was found.

\begin{table}[t!]
\caption{
Success rates obtained by Origami for each pattern in each benchmark. 
The ``Best'' column shows the highest success rate  for that benchmark across all patterns, which is also underlined.
We also show in blue the pattern of the canonical solution.
}
\label{tab:results_by_pattern}
    \centering
\stripedRows
\begin{tabular}{lrrrrrr|r}
\toprule
\textbf{Dataset} & NoScheme & Cata & CurriedCata & Ana & Accu & Hylo & \textbf{Best} \\
\midrule
\checksum & 0 & 0 & -- & -- & \canonical{0} & 0 & 0 \\
\collatzNumbers & 0 & -- & -- & -- & -- & \canonical{0} & 0 \\
\compareStringLengths & \canonical{\bestResult{90}} & -- & -- & -- & -- & -- & 90 \\
\countOdds & 0 & \canonical{\bestResult{40}} & -- & -- & -- & -- & {40} \\
\digits & 0 & -- & -- & \canonical{0} & -- & 0 & 0 \\
\doubleLetters & 0 & \canonical{\bestResult{3}} & -- & -- & -- & -- & {3} \\
\evenSquares & 0 & -- & -- & \canonical\bestResult{3} & -- & -- & {3} \\
\forLoopIndex & \bestResult{90} & -- & -- & \canonical{67} & -- & -- & {90} \\
\grade & 0 & \bestResult{100} & \canonical{10} & -- & -- & -- & {100} \\
\lastIndexOfZero & 0 & \canonical\bestResult{70} & -- & -- & -- & -- & {70} \\
\median & \canonical\bestResult{97} & -- & -- & -- & -- & -- & {97} \\
\mirrorImage& \canonical\bestResult{93} & -- & -- & -- & -- & -- & {93} \\
\negativeToZero & 0 & \canonical\bestResult{87} & -- & -- & -- & -- & {87} \\
\numberIo & \canonical\bestResult{100} & -- & -- & -- & -- & -- & {100} \\
\pigLatin & 0 & \canonical0 & -- & -- & -- & 0 & 0 \\
\replaceSpaceWithNewline & 0 & \canonical\bestResult{3} & -- & -- & -- & -- & {3} \\
\scrabbleScore & 0 & \canonical\bestResult{100} & -- & -- & -- & -- & {100} \\
\smallOrLarge & \canonical\bestResult{53} & -- & -- & -- & -- & -- & {53} \\
\smallest & \canonical\bestResult{100} & -- & -- & -- & -- & -- & {100} \\
\stringDifferences & 0 & 0 & \canonical{0} & -- & 0 & 0 & 0 \\
\stringLengthsBackwards & 0 & \canonical\bestResult{97} & -- & -- & -- & -- & {97} \\
\sumOfSquares & 0 & -- & -- & -- & -- & \canonical0 & 0 \\
\superAnagrams & 0 & 0 & \canonical\bestResult{73} & -- & -- & -- & {73} \\
\syllables & 0 & \canonical\bestResult{7} & -- & -- & -- & -- & {7} \\
\vectorAverage & 0 & 0 & -- & -- & \canonical{0} & 0 & 0 \\
\vectorsSummed & 0 & 0 & \canonical\bestResult{20} & -- & -- & -- & {20} \\
\wallisPi & 0 & -- & -- & -- & -- & \canonical{0} & 0 \\
\wordStats & 0 & 0 & -- & -- & \canonical{0} & 0 & 0 \\
\xWordLines & 0 & 0 & 0 & -- & \canonical{0} & 0 & 0 \\
\bottomrule
\end{tabular}
\end{table}

Considering the $4$ canonical Accu problems, \checksum and \wordStats are historically hard, with few methods ever finding a solution.
The same can be said for Hylo in the \wallisPi and \collatzNumbers problems.

In \vectorAverage, the canonical solution involved using Accu to compute both the sum and the count as a pair in the \code{st} slots, and using the \code{alg} slots to perform the division as a post-processing step, finally obtaining the average.
The solution that got closer to the intended result was the following:
\begin{minted}[escapeinside=@@]{haskell}
accu st alg @\argVal{0}@ (last @\argVal{0}@, length @\argVal{0}@)
  where
    st [] s = []
    st (x : xs) s = x : (xs, s)
    alg [] s = min 0 (last @\argVal{0}@)
    alg (x : acc) s = acc + ((max (x - acc) x) / (snd s))
\end{minted}

Origami took a different approach from the canonical solution, by storing the length of the input in the second element of the tuple while having no use for the first element.
The \code{st} section had no other purpose than to transmit this pre-processing step to the \code{alg} section.
This solution got a perfect score during training but failed in testing for certain cases.
If we were to replace \code{min 0 (last \argVal{0})} by \code{0} and \code{max (x - acc) x} by \code{x}, then this solution would be correct.


The Hylo solution for \sumOfSquares employed \code{coalg} to generate a list of all the numbers from $0$ to \argVal{0}, and then used \code{alg} to square each number and sum them.
Even though this was the simplest Hylo solution, its $5$ different slots incur an increased search space in relation to other patterns, which seems challenging for the algorithm.

\autoref{tab:vs_hotgp} compares Origami's results to \ac{HOTGP}'s.
There was a substantial increase ($>30$) in the success rate in $6$ problems.
In the $17$ problems where the absolute difference is $<30$, we highlight \syllables, \doubleLetters and \evenSquares problems, as those were problems for which \ac{HOTGP} was not able to synthesize a solution, whereas Origami was successful at least once.
The two problems with a more noticeable decrease are \replaceSpaceWithNewline and \vectorAverage.
These can be explained by the change in function-set between the two algorithms, as \ac{HOTGP}'s solutions were arguably simpler due to having \code{map} and \code{filter} for \replaceSpaceWithNewline and \code{sum} for \vectorAverage.
In a practical scenario, the inclusion of these functions would likely lead to a correct solution but, as previously noted, removing them was a conscious decision to enable the proper assessment of the impact of Recursion Schemes in PS.
It would also allow for composite solutions, such as using Ana with a \code{map} inside instead of relying on Hylo to find the entire pattern, which might be easier to synthesize.

\begin{table}[t!]
    \caption{Origami's success rates compared to \ac{HOTGP}'s on solved problems. The $\Delta$ column shows the relative success rate of Origami with respect to \ac{HOTGP}.}
    \label{tab:vs_hotgp}
    \centering
\stripedRows
\small
\begin{tabular}{lrrr||lrrr}
\toprule
\scriptsize  Dataset & \scriptsize Origami & \scriptsize \ac{HOTGP} & $\Delta$ & \scriptsize  Dataset & \scriptsize Origami & \scriptsize \ac{HOTGP} & $\Delta$  \\
 \midrule
\scrabbleScore & 100 & 0 & 100 & \numberIo & 100 & 100 & 0 \\
\mirrorImage & 93 & 1 & 92 & \sumOfSquares & 0 & 1 & -1 \\
\superAnagrams & 73 & 0 & 73 & \median & 97 & 99 & -2 \\
\lastIndexOfZero & 70 & 0 & 70 & \smallOrLarge & 53 & 59 & -6 \\
\grade & 100 & 37 & 63 & \compareStringLengths & 90 & 100 & -10 \\
\forLoopIndex & 90 & 59 & 31 & \countOdds & 40 & 50 & -10 \\
\stringLengthsBackwards & 97 & 89 & 8 & \negativeToZero & 87 & 100 & -13 \\
\syllables & 7 & 0 & 7 & \vectorsSummed & 20 & 37 & -17 \\
\doubleLetters & 3 & 0 & 3 & \replaceSpaceWithNewline & 3 & 38 & -35 \\
\evenSquares & 3 & 0 & 3 & \vectorAverage & 0 & 80 & -80 \\
\smallest & 100 & 100 & 0 \\
\bottomrule
\end{tabular}
\end{table}

\begin{table}[t!]
    \caption{Success rate with the best values underlined, with the ratio of victories of each algorithm against Origami by the amount of tested problems. The last rows display the amount of problems with a success rate above a certain \% for each method.}
    \label{tab:vs_all}
    \centering
    \renewcommand{\arraystretch}{1.25}
   
\scriptsize
\stripedRows
\begin{tabularx}{\textwidth}{X|rrrrrrrrrr}
 \toprule
 \rowcolor{white}
 \scriptsize	\textbf{Dataset} &  \scriptsize	Origami & \scriptsize	HOTGP &\scriptsize	 DSLS & \scriptsize	UMAD & \scriptsize	PushGP & \scriptsize	CBGP & \scriptsize	G3P & \scriptsize	G3P+ & \scriptsize	G3Phs & \scriptsize	G3Ppy \\
 \midrule
{\checksum} & 0 & 0 & \bestResult{18} & 5 & 0 & -- & 0 & 0 & -- & -- \\


{\compareStringLengths} & 90 & \bestResult{100} & 51 & 42 & 7 & 22 & 2 & 0 & 5 & 0 \\

{\countOdds} & 40 & \bestResult{50} & 11 & 12 & 8 & 0 & 12 & 3 & -- & -- \\

{\digits} & 0 & 0 & \bestResult{28} & 11 & 7 & 0 & 0 & 0 & -- & -- \\

{\doubleLetters} & 3 & 0 & \bestResult{50} & 20 & 6 & -- & 0 & 0 & -- & -- \\

{\evenSquares} & \bestResult{3} & 0 & 2 & 0 & 2 & -- & 1 & 0 & -- & -- \\

{\forLoopIndex} & \bestResult{90} & 59 & 5 & 1 & 1 & 0 & 8 & 6 & -- & -- \\

{\grade} & \bestResult{100} & 37 & 2 & 0 & 4 & -- & 31 & 31 & -- & -- \\

{\lastIndexOfZero} & \bestResult{70} & 0 & 65 & 56 & 21 & 10 & 22 & 44 & 0 & 2 \\

{\median} & 97 & \bestResult{99} & 69 & 48 & 45 & 98 & 79 & 59 & 96 & 21 \\

{\mirrorImage} & 93 & 1 & 99 & \bestResult{100} & 78 & \bestResult{100} & 0 & 25 & -- & -- \\

{\negativeToZero} & 87 & \bestResult{100} & 82 & 82 & 45 & 99 & 63 & 13 & 0 & 66 \\

{\numberIo} & \bestResult{100} & \bestResult{100} & 99 & \bestResult{100} & 98 & \bestResult{100} & 94 & 83 & 99 & \bestResult{100} \\

{\pigLatin} & 0 & 0 & 0 & 0 & 0 & -- & 0 & \bestResult{3} & -- & -- \\

{\replaceSpaceWithNewline} & 3 & 38 & \bestResult{100} & 87 & 51 & 0 & 0 & 16 & -- & -- \\

{\scrabbleScore} & \bestResult{100} & 0 & 31 & 20 & 2 & -- & 2 & 1 & -- & -- \\

{\smallOrLarge} & 53 & \bestResult{59} & 22 & 4 & 5 & 0 & 7 & 9 & 4 & 0 \\

{\smallest} & \bestResult{100} & \bestResult{100} & 98 & \bestResult{100} & 81 & \bestResult{100} & 94 & 73 & \bestResult{100} & 89 \\

{\stringLengthsBackwards} & \bestResult{97} & 89 & 95 & 86 & 66 & -- & 68 & 18 & 0 & 34 \\

{\sumOfSquares} & 0 & 1 & 25 & \bestResult{26} & 6 & -- & 3 & 5 & -- & -- \\

{\superAnagrams} & \bestResult{73} & 0 & 4 & 0 & 0 & -- & 21 & 0 & 5 & 38 \\

{\syllables} & 7 & 0 & \bestResult{64} & 48 & 18 & -- & 0 & 39 & -- & -- \\

{\vectorAverage} & 0 & 80 & \bestResult{97} & 92 & 16 & 88 & 5 & 0 & 4 & 0 \\

{\vectorsSummed} & 20 & 37 & 21 & 9 & 1 & \bestResult{100} & 91 & 21 & 68 & 0 \\

{\xWordLines} & 0 & 0 & \bestResult{91} & 59 & 8 & -- & 0 & 0 & -- & -- \\
\midrule
Win ratio vs. Origami & &
$\dfrac{9}{25}$ & 
$\dfrac{10}{25}$ & 
$\dfrac{9}{25}$ & 
$\dfrac{7}{29}$ & 
$\dfrac{5}{14}$ & 
$\dfrac{3}{29}$ & 
$\dfrac{5}{29}$ & 
$\dfrac{2}{11}$ & 
$\dfrac{0}{11}$ \\
\midrule

   $\mathbf{= 100\%}$ & \bestResult{4} & \bestResult{4} & 1 & 3 & 0 & \bestResult{4} & 0 & 0 & 1 & 1  \\
   $\mathbf{\geq 75\%}$ & \bestResult{10} & 7 & 8 & 7 & 3 & 7 & 4 & 1 & 3 & 2 \\
   $\mathbf{\geq 50\%}$ & \bestResult{13} & 10 & 13 & 9 & 5 & 7 & 6 & 3 & 4 & 3 \\
   $\mathbf{\geq 25\%}$ & 14 & 13 & \bestResult{16} & 13 & 7 & 7 & 7 & 7 & 4 & 5 \\
   $\mathbf{> 0\%}$ & 19 & 15 & \bestResult{24} & 21 & 22 & 9 & 17 & 17 & 8 & 7 \\
   \bottomrule
   \end{tabularx}
\end{table}
To assess how Origami fares with relation to the best methods found in the literature, we compare its results to those obtained by different versions of PushGP~\citep{psb1, dsls, umad}, \acf{G3P}~\citep{g3p}, the extended grammar version of G3P (here called \ac{G3P+})~\citep{g3pe}, \acf{CBGP}~\citep{pantridge2022functional}, \ac{G3P} with Haskell and Python grammars (G3Phs and G3Ppy)~\citep{garrow2022functional}, as well as with Github Copilot~\citep{copilot}. 
In some of those works, only a subset of the problems was chosen, often avoiding the most difficult ones (\emph{i.e.}, not previously solved by any other method), or problems not solvable by the proposed method itself.
The results are reported in \autoref{tab:vs_all} (``--''~indicates the authors did not test their method on that specific problem).

Among the other \ac{GP} algorithms, the best performer (\ac{DSLS}) achieves a higher success rate than Origami in only 10 problems.
Considering that they both get a $0$ in \pigLatin, Origami outperforms \ac{DSLS} in 14 problems.
Notably, Origami frequently outperforms \ac{CBGP} and the \ac{G3P} variants.
It also has the highest number of problems solved with $100\%$, $\geq75\%$, and $\geq50\%$, and is second-place in $\geq25\%$.
When we consider problems to which Origami found at least one solution, we note that it outperforms \ac{HOTGP}, \ac{CBGP}, and all the \ac{G3P} variations, placing Origami at the fourth place. 
It is also worth noting that
Origami outperforms \ac{HOTGP} in both the number of best results and amount of problems above all thresholds, which demonstrates it is a substantial improvement over \ac{HOTGP}.

We also compare with the results obtained using Copilot on \ac{PSB1}, as reported by~\citet{copilot}. In that paper, the authors tested Copilot with a different formulation of the program synthesis problem: instead of receiving example of input-outputs, they made the problem description and the function signature available to Copilot.
This input format can be more difficult to process, as it requires the extraction of useful information from a textual description, but contains additional information that may be implicit in the input-output format. 
As shown in \autoref{tab:vs_copilot}, out of the $29$ problems, Copilot had better results in $14$ of them, and
Origami is equivalent or outperforms Copilot in $15$ problems. 
While Copilot solves more problems (at least once) than Origami, it struggles with consistency, and does not achieve $100\%$ success rate on any of the tested problems. 

\afterpage{
\begin{table}[p]
    \caption{Comparison between Origami and Copilot. Success rates with the best values underlined, with the ratio of victories of each algorithm against Origami by the amount of tested problems. The last rows display the amount of problems with a success rate above a certain \% for each method.}
    \label{tab:vs_copilot}
    \centering
    \renewcommand{\arraystretch}{1.25}
   
\scriptsize
\stripedRows
\begin{tabular}{l|rr}
 \toprule
 \rowcolor{white}
 \scriptsize	\textbf{Dataset} &  \scriptsize	Origami & \scriptsize Copilot\\
 \midrule
{\checksum} & 0 & \bestResult{89} \\

{\collatzNumbers} & 0 & \bestResult{73} \\

{\compareStringLengths} & \bestResult{90} & 70 \\

{\countOdds} & 40 &  \bestResult{98} \\


{\doubleLetters} & 3 & \bestResult{88} \\

{\evenSquares} & 3 & \bestResult{11} \\

{\forLoopIndex} & \bestResult{90} & 72 \\

{\grade} & \bestResult{100} & 84 \\

{\lastIndexOfZero} & \bestResult{70} & 61 \\

{\median} & 97 & 79 \\

{\mirrorImage} & 93 & 70 \\

{\negativeToZero} & 87 & 99 \\

{\numberIo} & \bestResult{100} & 93 \\

{\pigLatin} & 0 & \bestResult{54} \\

{\replaceSpaceWithNewline} & 3 & 87 \\

{\scrabbleScore} & \bestResult{100} & 35 \\

{\smallOrLarge} & 53 & 51 \\

{\smallest} & \bestResult{100} & 66 \\

{\stringLengthsBackwards} & \bestResult{97} &  60 \\

{\sumOfSquares} & 0 & \bestResult{90} \\

{\superAnagrams} & \bestResult{73} & 55 \\

{\syllables} & 7 & \bestResult{96} \\

{\vectorAverage} & 0 & \bestResult{92} \\

{\vectorsSummed} & 20 & \bestResult{87} \\

{\xWordLines} & 0 & \bestResult{1} \\
\midrule
Win ratio vs. Origami & &
$\dfrac{14}{29}$ \\

\midrule
$\mathbf{= 100\%}$ & \bestResult{4} & 0 \\
$\mathbf{\geq 75\%}$ & 10 & \bestResult{12} \\
$\mathbf{\geq 50\%}$ & 13 & \bestResult{22} \\
$\mathbf{\geq 25\%}$ & 14 & \bestResult{23} \\
$\mathbf{> 0\%}$ & 19 & \bestResult{26} \\
\bottomrule
\end{tabular}
\end{table}
}

\section{Final Remarks}
\label{sec:bananas_outro}

This chapter presents the first full implementation of Origami, a \ac{GP} algorithm proposed in \autoref{cha:origami}, and builds on its previous work, \ac{HOTGP}.
Origami's main differential is the use of \acp{RS}, well-known constructs in functional programming that enable recursive algorithms to be defined in a unified manner.
The main motivation for using these in the PS context is enabling recursive programs to be synthesized in a controlled manner, without sacrificing expressiveness.

We evaluate our approach in the 29 problems in the \ac{PSB1} dataset, which is known to be solvable by just a handful of \acp{RS}.
In general, Origami performs better than other similar methods, synthesizing the correct solution more often than other methods in most problems.
It was also able to obtain the highest count of problems with success rate $=100\%$, $\geq75\%$ and $\geq50\%$ among the \ac{GP} methods.
Furthermore, Origami achieved comparable results to Github Copilot, solving some problems that the \ac{LLM} achieved $0\%$ score. We should stress that the problem formulation is different for both approaches, indicating that combining \ac{LLM} with \ac{GP} and \acp{RS} could be beneficial to improve the results.
These experimental results suggest that using Recursion Schemes to guide the search is a promising research avenue.

Currently, the main challenge of Origami appears to be dealing with harder \acp{RS}, such as Accumulation and Hylomorphism. 
Different evolutionary mechanisms, such as other selection methods and mutation/crossover operators, should be evaluated in this context to understand if they can positively impact the search process.

%% file: text/acdc.tex
\chapter{AC/DC}
\label{cha:acdc}

\newcommand{\pct}[2]{%
    $\fpeval{round(#1/#2*100,0)}\%$%
}
\newcommand{\fracpct}[2]{%
  $\frac{#1}{#2}$~(\pct{#1}{#2})%
}

\newcommand{\acdc}{\ac{AC/DC}\xspace}

\renewcommand{\origami}{Origami\xspace}
\newcommand{\origamipoc}{Origami\textsubscript{POC}\xspace} 
\newcommand{\origamibananas}{Origami\textsubscript{BNNS}\xspace} 
\newcommand{\origamidsls}{Origami\textsubscript{DSLS}\xspace} 
\newcommand{\origamiacdc}{Origami\textsubscript{AC/DC}\xspace} 

After introducing the concept of \origami in \autoref{cha:origami}, \autoref{cha:bananas} presented its first full implementation (hereafter, \origamibananas).
This version was capable of synthesizing recursive programs across multiple recursion schemes and consistently outperformed other \ac{GP} methods.

Nevertheless, one limitation observed in \origamibananas is its lack of population diversity, as it frequently spends considerable time exploring solutions that specialize in a similar subset of the training examples. 
Even with syntactic differences, they had a similar prediction error throughout all the examples.

In this chapter, we introduce a new diversity mechanism called \acdc that aims to remove part of the individuals specialized in the same test cases while replacing them with new randomly generated programs, allowing a more efficient exploration of the search space (see \autoref{sec:acdc}).
Beyond that, we also implement and evaluate:
\begin{itemize}
    \item A new \ac{RS} (histomorphism), as well as variations of catamorphism for other data-structures (namely \code{Map} and \code{Set});
    \item A new selection strategy: \acf{DSLS}~\citep{dsls};
    \item A less restrictive function set, so that \origami can synthesize more expressive programs;
    \item A more comprehensive experimental evaluation by evaluating \origami in two new benchmark suites, namely \ac{PSB2}~\citep{psb2} and \ac{PolyPSB}~\citep{polypsb};
\end{itemize}

The remainder of this chapter is organized as follows.
In \autoref{sec:acdc}, we introduce and describe \acdc, which is the main contribution of this chapter.
In \autoref{sec:extensions}, we present other proposed extensions to the Origami algorithm.
In \autoref{sec:results}, we empirically evaluate this version of Origami and analyze the results.

We highlight that the contents of this chapter constitute a paper which is yet to be published.
A considerable portion of the text presented here is an adaptation or extension of the text presented in the paper.

\section{\acl{AC/DC}}
\label{sec:acdc}

The main contribution of this chapter is a proposal of a two-part diversification procedure called \acf{AC/DC}.

\acf{DC} groups individuals sharing identical error vectors, and removes all of them except the one with the fewest nodes (\ie \emph{culling}, since smaller programs potentially have a higher generalization capability~\citep{helmuth2017improving}).
Using the error vector removes individuals that specialize in the same examples, even if their outputs differ in the incorrect predictions.
This procedure is executed every $g_{\mathit{dc}}$ generations.

As \ac{DC} consistently reduces the population count, new individuals are randomly generated, using the same procedure employed during the initialization, to replace the removed ones.
This aims to periodically introduce new genetic material, mitigating premature convergence and leading to a more diverse population and more comprehensive exploration of the search space.

A complementary step involves applying an \acf{AC} procedure to every individual in the population every \gac generations.
Previously, \origamibananas and \ac{HOTGP} applied simplification (\ie \emph{abridging}) to the best individual it found at the end of the search, only as a post-processing step (see \autoref{sec:bananas_methods}).
This procedure first performs a behavior-keeping change by evaluating branches which do not rely on any input, replacing them with constant leaves (\ie \emph{clipping}).
Then, it applies the local search procedure previously presented in \autoref{sec:code_refinements} (\autoref{alg:ls}), which tries to replace each node by each of its children, reevaluating the program and keeping changes that do not reduce the fitness value.

Extending this clipping procedure to the entire population improves the fitness of certain individuals, as some parts of the program may actually increase the error (see \autoref{sec:code_refinements}).
Furthermore, as it eliminates segments of the program that do not contribute to the overall fitness, making it part of the evolution creates space for mutation and crossover operators to explore more variations without exceeding the maximum size constraints.

As a result of the application of \ac{AC}, the chance that some individuals share the same error vector increases. 
This, in turn, creates an opportunity for \ac{DC} which will then eliminate redundant solutions and introduce novel genetic material into the population.
Therefore, \ac{AC} should be executed at least before every \ac{DC}. 
In other words, it is desirable that $\gdc = k \cdot \gac$ for some $k \in \mathbb{N}^+$.
An empirical analysis for the choice of the \gac and \gdc values is shown in \autoref{sec:acdc_hyper_empirical}.

\section{Extensions to Origami}
\label{sec:extensions}
This section proposes enhancements to the \origami algorithm: 
three new patterns (including a new \ac{RS}),
a new selection strategy,
and an extended function set.

\subsection{New Patterns}
\label{sec:new_patterns}

For the current work, we manually solved two additional benchmarks:
\begin{itemize}
    \item \textbf{\acs{PSB2}}~\citep{psb2}: The \acf{PSB2} is the successor to \ac{PSB1}, introducing more challenging problems;
    \item \textbf{\acs{PolyPSB}}~\citep{polypsb}: A benchmark that was introduced to highlight the capabilities of \ac{CBGP} in handling polymorphic data and functions, which is also supported by \origami. Since this benchmark lacks an official name, we unofficially refer to it as the \acf{PolyPSB}.
\end{itemize}
The canonical solutions for these two new benchmarks are also available on \href{https://github.com/folivetti/origami-programming/}{GitHub}\footnote{
\href{https://github.com/folivetti/origami-programming/}{https://github.com/folivetti/origami-programming}
}.
They required one previously unused \ac{RS} (histomorphism), as well as catamorphism variations for $2$ different data structures: \code{Set} and \code{Map}, which can be encoded into $3$ additional patterns.
These additions do not incur any changes to the algorithm, as \origami is capable of supporting any \ac{RS} in general.

\subsubsection{Catamorphism over a Set}

This pattern is very similar to the regular Cata, but traversing a \code{Set} instead of an indexed list:

\begin{minted}[escapeinside=@@]{haskell}
f @\argVal{0}@ ... @\argVal{n}@ = 
    cata alg (fromSet @\argVal{0}@) where
      alg SNilF = @\slot{1}@
      alg (SConsF x acc) = @\slot{2}@
\end{minted}

In a problem with arguments of type \code{\argType{0} $\ldots$ \argType{n}} and of output type \code{o}, where \code{\argType{0}~$\equiv$~Set e}%
\footnote{The notation \code{\argType{0}~$\equiv$~Set e} is a restriction such that \code{\argType{0}} can be decomposed into the type \code{Set e}, which is the type of a set with elements of some type \code{e}.}%
, the slots are typed as follows:

\begin{itemize}
    \item \code{\slot{1}~::~o}, with nothing in scope;
    \item \code{\slot{2}~::~o},
    \begin{scope}
        \scopeitem{x}{e}
        \scopeitem{acc}{o}
        \argScope
    \end{scope}.
\end{itemize}

This pattern will be referred simply as \emph{Cata Set} in the remainder of this work.

\subsubsection{Catamorphism over a Map}

Similarly, this pattern represents folding over a Map, while allowing simultaneous access for the current key and value:

\begin{minted}[escapeinside=@@]{haskell}
f @\argVal{0}@ ... @\argVal{n}@ = 
    cata alg (fromMap @\argVal{0}@) where
      alg MNilF = @\slot{1}@
      alg (MConsF k v acc) = @\slot{2}@
\end{minted}

In a problem with arguments of type \code{\argType{0} $\ldots$ \argType{n}} and of output type \code{o}, where \code{\argType{0}~$\equiv$~ Map a b}, the slots are typed as follows:

\begin{itemize}
    \item \code{\slot{1}~::~o}, with nothing in scope;
    \item \code{\slot{2}~::~o},
    \begin{scope}
        \scopeitem{k}{a}
        \scopeitem{v}{b}
        \scopeitem{acc}{o}
        \argScope
    \end{scope}.
\end{itemize}

This pattern will be referred simply as \emph{Cata Map} in the remainder of this work.

\subsubsection{Histomorphism over Indexed List}

This pattern is a generalization of catamorphism, giving it access to the entire history of the recursion so far.
It is specifically useful in problems that have the need to compare the current element with the previous elements in some way.
It is represented by the following:

\begin{minted}[escapeinside=@@]{haskell}
f @\argVal{0}@ ... @\argVal{n}@ = histo alg (fromList @\argVal{0}@)
    where
      alg INilF = @\slot{1}@
      alg (IConsF i x table) = @\slot{2}@
      where
        acc = extract table
        tableAsList = tbl2List table
\end{minted}

When compared to the regular Cata, the only difference is that instead of being able to access just the last value returned by \code{alg}, it has access to \code{table}, containing the entire history of the values returned by it.

This table is traditionally encoded in a representation that allows for any data structure to record its history.
Since this pattern only deals with indexed lists, for convenience, we introduced two bindings: \code{tableAsList} has the previously computed elements as a regular Haskell linked-list; and \code{acc} contains the latest value returned by \code{alg}.

In a problem with arguments of type \code{\argType{0} $\ldots$ \argType{n}} and of output type \code{o}, where \code{\argType{0}~$\equiv$~[e]}, the slots are typed as follows:

\begin{itemize}
    \item \code{\slot{1}~::~o}, with nothing in scope;
    \item \code{\slot{2}~::~o},
    \begin{scope}
        \scopeitem{i}{Int}
        \scopeitem{x}{e}
        \scopeitem{acc}{o}
        \scopeitem{tableAsList}{[e]}
        \argScope
    \end{scope}.
\end{itemize}

This pattern will be referred simply as \emph{Histo} in the remainder of this work.

\subsection{Selection mechanism}
\label{sec:dsls}

We also consider the \acf{DSLS}~\citep{dsls} as a selection strategy, instead of the tournament selection that was previously used in \origamibananas.
This change was motivated by its positive impact on the PushGP algorithm, which could potentially be translated to \origami.

Lexicase Selection aims to maintain a diverse population, in terms of the subset of examples that each individual solves.
The goal is to combine genetic material from programs that successfully solve different subsets of the examples, hoping that the resulting offspring solves both subsets simultaneously.

The standard Lexicase Selection algorithm~\citep{lexicase} iterates through the examples in shuffled order, at each step selecting only the individuals that have the lowest error for the current example. 
This process stops when only one individual remains, or when the examples are exhausted (in which case a random individual is selected from the remaining ones).

In contrast, \ac{DSLS}~\citep{dsls} is a variation that shuffles a smaller random subset of examples, instead of the entire dataset.
This approach makes selection not only faster, as fewer programs need to be evaluated, but empirically leads to better results.
In experimental evaluations on PushGP, \ac{DSLS} was reported as a substantial improvement over standard Lexicase Selection, being robust to a wide range of subsample sizes.
Evidence suggests that its success is due to allowing more individuals to be evaluated, as well as submitting them through a ``changing environment''~\citep{dsls}, in which the set of examples is constantly changing.

In the context of this work, \ac{DSLS} complements the selection pressure from \acdc, which promotes a diverse population that fails in different cases and increases the chance of producing solutions that cover complementary errors.

\subsection{Function set}

The previous chapters involving \origami aimed to assess whether representing recursion using \acp{RS} was beneficial for \ac{PS}.
For that intent, \origamibananas needed to synthesize solutions using \acp{RS} as the \emph{only} means of recursion.
Thus, its function set was purposefully restricted, not allowing most implicitly recursive functions (such as \code{sum} and \code{product}).
If \origamibananas needed to perform those operations, it would need to synthesize them purely through the \acp{RS}, which it was able to do with varying degrees of success.

In the current chapter, we aim to test \origami as a feature-complete synthesizer, which means that it should have access to the set of functions that are commonly provided by the Haskell standard library.
We highlight, however, that a function set can be undesirably large, making the search space larger than it needs to be.
Furthermore, certain functions can trivialize some problems, leading to an unfair comparison to other methods.
To achieve a balance, we provide a function set that is intended to be equivalent to the ones provided by other synthesizers, such as PushGP~\citep{psb1} and \ac{CBGP}~\citep{pantridge2022functional}, using the most similar functions available in the Haskell standard library.
The function set is provided in detail in \autoref{tab:function_set}.

\section{Experimental Results}
\label{sec:results}

This chapter also extends previous chapters by performing a more comprehensive experimental evaluation of Origami.
While the previous evaluation (\autoref{sec:bananas_results}) already tackled \ac{PSB1}~\citep{psb1} ($29$ problems), we now incorporate both \ac{PSB2}~\citep{psb2} ($25$ problems) and \ac{PolyPSB}~\citep{polypsb} ($17$ problems), with a total of $71$ problems. 
Each problem prescribes which type of operations should be available, as well as providing ways to obtain or generate input/output examples.
They also define a small number to be used as the count of examples that are visible to the synthesizer (train cases), as well as a larger number indicating how many examples are hidden, reserved to evaluate the generalization capabilities of the synthesized program (test cases).

Our evaluation is centered on the assessment of the new version of \origami, denoted \origamiacdc. 
This variant integrates the entire set of improvements presented in this chapter and will be used by most of the experiments.
However, to allow verifying the contribution of the \acdc procedures alone (see \autoref{sec:ablation}), we also introduce a separate version of the algorithm, \origamidsls, which contains just the extensions presented on \autoref{sec:extensions}, and does not contain \acdc.

The main experimental evaluation follows the same strategy used to evaluate \origamibananas in \autoref{sec:bananas_results}.
For each problem, we perform $30$ independent runs (initialized with different seeds) for each pattern in the following order:
\begin{enumerate*}
    \item NoScheme;
    \item Cata List/Set/Map;
    \item CurriedCata;
    \item Histo;
    \item Ana;
    \item Accu;
    \item Hylo.
\end{enumerate*}

\begin{table}[p]
    \centering
    \caption{The function set available for \origamidsls.}
    \rowcolors{2}{gray!25}{white}
    \label{tab:function_set}
\scriptsize
\begin{tabular}{p{.5\linewidth}|p{.46\linewidth}}
    \toprule
    \textbf{Operations} & \textbf{Types} \\
    \midrule

    \code{addInt, subInt, multInt, divInt, quotInt, modInt, remInt, minInt, maxInt} & \code{Int -> Int -> Int} \\
    \code{absInt, incrementInt, decrementInt} & \code{Int -> Int} \\
    \code{sumInt, productInt} & \code{[Int] -> Int} \\
    \midrule

    \code{addFloat, subFloat, multFloat, divFloat, minFloat, maxFloat, powFloat} & \code{Float -> Float -> Float} \\
    \code{absFloat, sqrt, sin, cos, incrementFloat, decrementFloat} & \code{Float -> Float} \\
    \code{sumFloat, productFloat} & \code{[Float] -> Float} \\
    \midrule

    \code{ltInt, gtInt, gteInt, lteInt} & \code{Int -> Int -> Bool} \\
    \code{ltFloat, gtFloat, gteFloat, lteFloat} & \code{Float -> Float -> Bool} \\
    \code{fromIntegral} & \code{Int -> Float} \\
    \code{floor, ceiling, round} & \code{Float -> Int} \\
    \midrule

    \code{and, or} & \code{Bool -> Bool -> Bool} \\
    \code{not} & \code{Bool -> Bool} \\
    \code{if} & \code{Bool -> a -> a -> a} \\
    \code{eq, neq} & \code{a -> a -> Bool} \\
    \midrule

    \code{showInt} & \code{Int -> [Char]} \\
    \code{showFloat} & \code{Float -> [Char]} \\
    \code{showBool} & \code{Bool -> [Char]} \\
    \code{showChar} & \code{Char -> [Char]} \\
    \code{charToInt, digitToInt} & \code{Char -> Int} \\
    \code{intToChar} & \code{Int -> Char} \\
    \code{isLetter, isSpace, isDigit} & \code{Char -> Bool} \\
    \code{toLower, toUpper} & \code{Char -> Char} \\
    \midrule

    \code{length} & \code{[a] -> Int} \\
    \code{cons, snoc, mappend, delete} & \code{a -> [a] -> [a]} \\
    \code{elem} & \code{a -> [a] -> Bool} \\
    \code{null} & \code{[a] -> Bool} \\
    \code{head, last} & \code{[a] -> a} \\
    \code{tail, init} & \code{[a] -> [a]} \\
    \code{zip} & \code{[a] -> [b] -> [(a,b)]} \\
    \code{replicate} & \code{Int -> a -> [a]} \\
    \code{enumFromThenTo} & \code{Int -> Int -> Int -> [Int]} \\
    \code{take, drop} & \code{Int -> [a] -> [a]} \\
    \code{takeWhile} & \code{(a -> Bool) -> [a] -> [a]} \\
    \code{any, all} & \code{(a -> Bool) -> [a] -> Bool} \\
    \code{reverse} & \code{[a] -> [a]} \\
    \code{splitAt} & \code{Int -> [a] -> ([a], [a])} \\
    \code{intercalate} & \code{[a] -> [[a]] -> [a]} \\
    \midrule

    \code{fst} & \code{(a, b) -> a} \\
    \code{snd} & \code{(a, b) -> b} \\
    \code{pair} & \code{a -> b -> (a, b)} \\
    \midrule

    \code{apply} & \code{(a -> b) -> a -> b} \\
    \midrule

    \code{singleton} & \code{k -> v -> Map k v} \\
    \code{insert, insertWith} & \code{k -> v -> Map k v -> Map k v} \\
    \code{lookup} & \code{k -> Map k v -> Maybe v} \\
    \code{fromList} & \code{[(k, v)] -> Map k v} \\
    \code{keys} & \code{Map k v -> [k]} \\
    \midrule

    \code{fromList} & \code{[a] -> Set a} \\
    \code{toList} & \code{Set a -> [a]} \\
    \code{insert, delete} & \code{a -> Set a -> Set a} \\
    \code{member} & \code{a -> Set a -> Bool} \\
    \code{union, intersection, difference} & \code{Set a -> Set a -> Set a} \\
    \midrule

    \code{gtOrd, ltOrd, gteOrd, lteOrd} & \code{Ord a => a -> a -> Bool} \\
    \bottomrule
\end{tabular}
\end{table}

The execution only advances to the next pattern when none of the $30$ seeds was able to find a solution that solves the test dataset.
This simulates a realistic scenario, where there would be no previous information regarding which scheme is capable or better suited to solve the problem, so all of them need to be tested.
Also, a pattern is only attempted if the input/output type of the problem matches the one expected by the pattern.
For example, CurriedCata will only be executed for problems with two arguments where the first one is a list.
Note that it is trivial to decide which Cata variation to use, as it is uniquely determined by the type of the first argument.
This specific order was chosen so simpler patterns are given preference, while also avoiding the decision of the unbound types in Accu and Hylo when another pattern solves the problem.


\subsection{\acdc Hyperparameters}
\label{sec:acdc_hyper_empirical}

Before carrying out the main evaluation, we empirically determined the values for \gac and \gdc, which are the execution frequencies of the \ac{AC} and \ac{DC} procedures, respectively (see \autoref{sec:acdc}).
We executed $10$ independent runs of \origamiacdc (with different seeds) on a subset of $4$ problems: 
\emph{fizz-buzz} (\ac{PSB2}), 
\emph{mirror-image}, 
\emph{small-or-large}, and
\emph{vector-average} (\ac{PSB1}).

\begin{figure}[H]
    \centering
    \begin{subfigure}[t]{.6\linewidth}
        \centering
        \includegraphics[width=\linewidth]{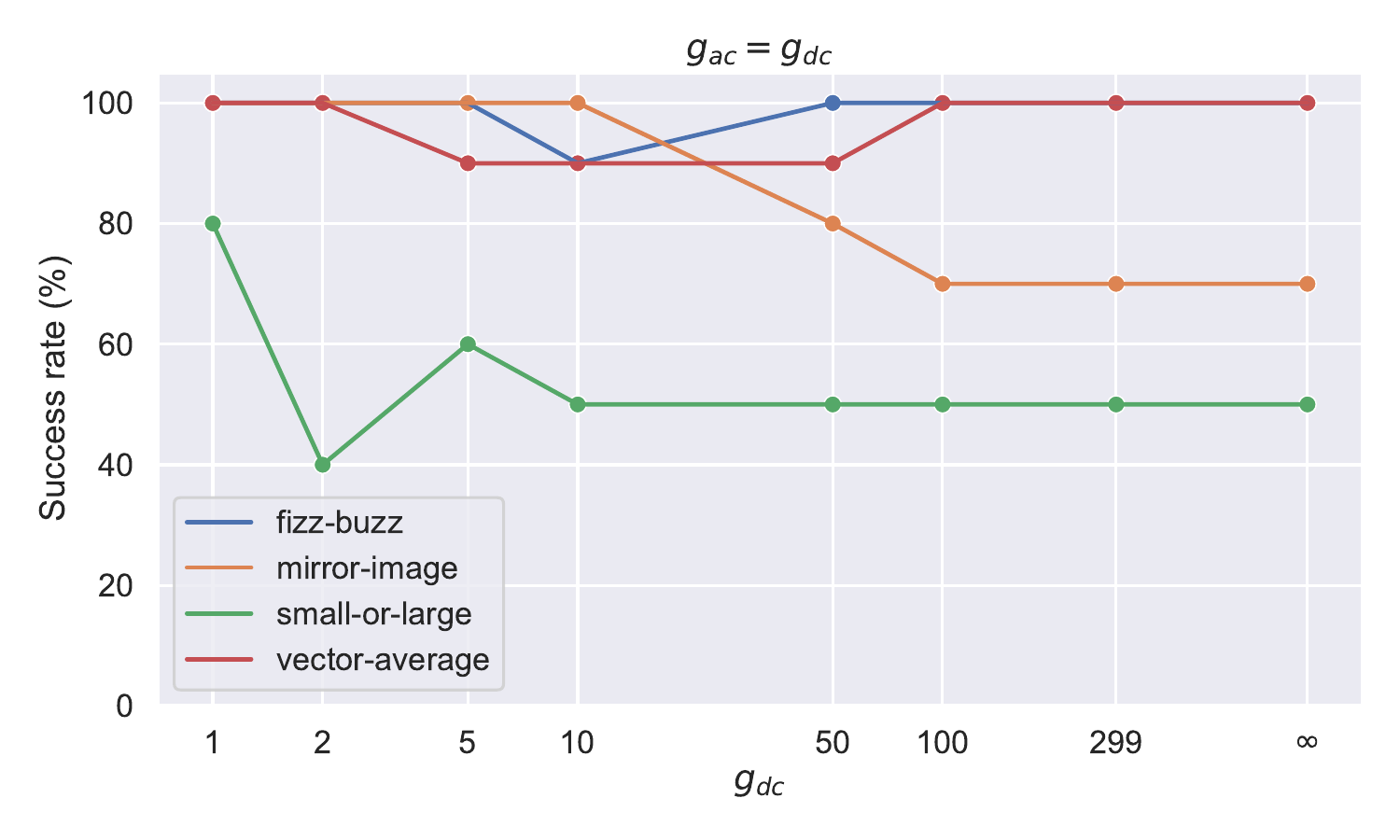}
        \caption{AC and DC executed at equal intervals.}
        \label{fig:a_equals_d}
    \end{subfigure}
    
    \begin{subfigure}[t]{.6\linewidth}
        \centering
        \includegraphics[width=\linewidth]{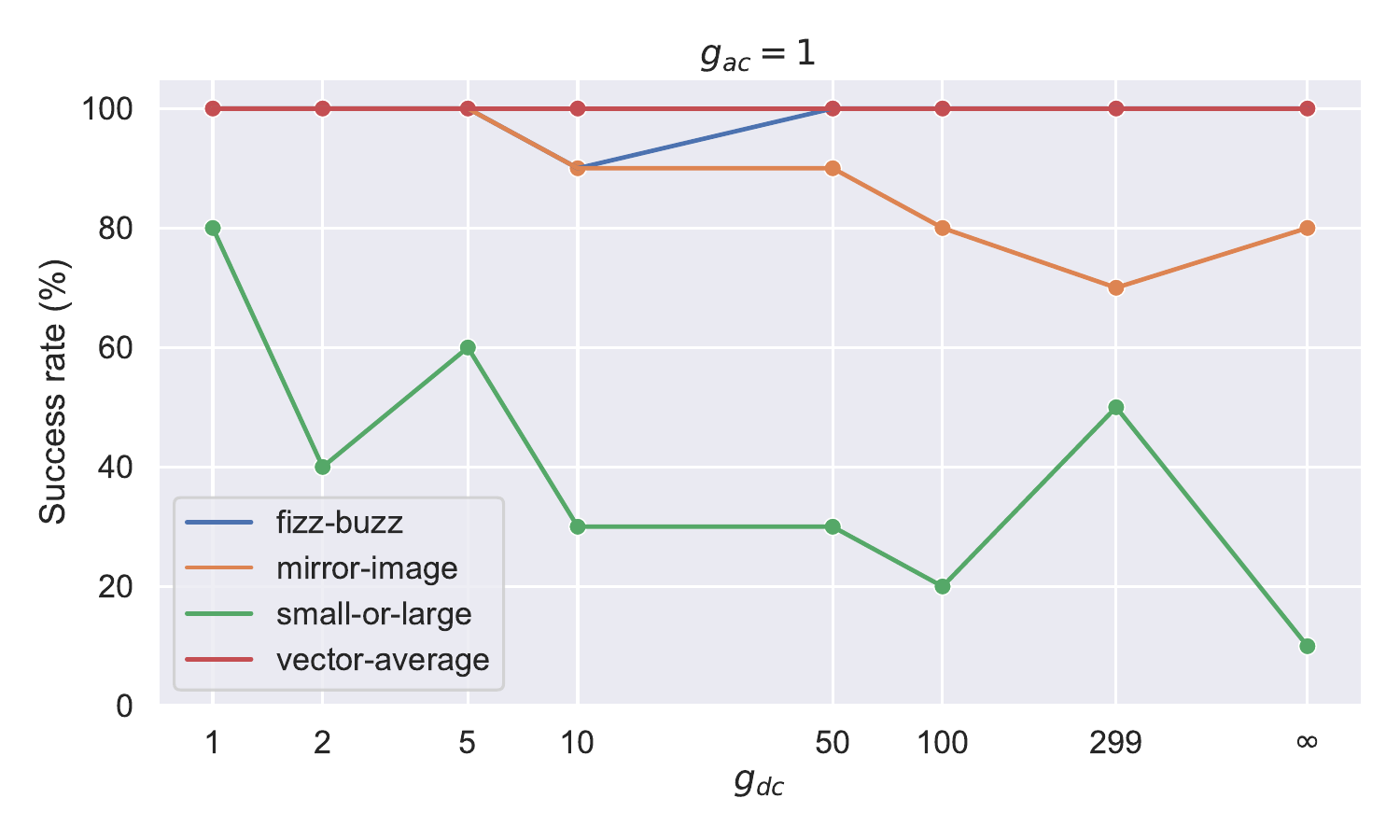}
        \caption{AC executed every generation, DC executed at varied intervals.}
        \label{fig:a_equals_1}
    \end{subfigure}

    \caption{Success rates under different AC and DC configurations. A value of $1$ indicates execution every generation, $2$ every other generation, $299$ only once just before the final ($300^{\mathit{th}}$) generation, and $\infty$ indicates the procedure is never executed.}
    \label{fig:grid_search}
\end{figure}

In the first experiment, we tested the values $\left\{1,2,5,10,50,100,299,\infty\right\}$ with \gac = \gdc.
\autoref{fig:a_equals_d} shows the \emph{success rates} of each evaluation, which measures the percentage of executions (in this case, $10$) for which \origamiacdc synthesized a solution that successfully handles all the test cases.

The overall peak at $\gdc=1$ indicates that the best performance is achieved when both were set to $1$, meaning that \ac{AC} and \ac{DC} were applied every generation.
Next, we fixed $\gac=1$ and tested different values for \gdc (\autoref{fig:a_equals_1}).
The results confirmed that performing \ac{AC} and \ac{DC} at every generation ($\gac=\gdc=1$) is the configuration that empirically seems to obtain the best performance.

\subsection{Results by Scheme}

\newcommand{\pushgp}{PushGP\textsubscript{0}\xspace}
\newcommand{\umad}{PushGP\textsubscript{UMAD}\xspace}
\newcommand{\dsls}{PushGP\textsubscript{DSLS}\xspace}

We executed \origamiacdc following the experimental framework described earlier.
\autoref{tab:by_scheme_psb1}, \autoref{tab:by_scheme_psb2}, \autoref{tab:by_scheme_polypsb} show the success rates for each pattern in the tested problems.
Underlined values show the highest rate for a given problem, and blue values indicate its canonical pattern --- that is, the pattern that was used in the manual implementation presented as the canonical solution.

The bottom rows summarize, for each pattern, the number of problems whose canonical solution use this pattern (as ``Total canonical'') and how many of these problems were solved at least once using that pattern (as ``Solved canonical'').
They also report the number of problems solved using a given pattern, but whose canonical solution corresponded to a different pattern (as ``Solved non-canonical'').




\newcommand{\needToRun}{0}
\newcommand{\stillRunning}[1]{#1}

\begin{table}[b!]
\caption{
Success Rates of each pattern on \ac{PSB1} for \origamiacdc.
}
\label{tab:by_scheme_psb1}
\centering
\rowcolors{2}{gray!25}{white}
\footnotesize
\hspace*{-1em}
\begin{tabular}{lrrrrrrr|r}
\toprule
\textbf{Dataset} & NoScheme & Cata List & CurriedCata & Histo & Ana & Accu & Hylo & \textbf{Best} \\
\midrule
checksum & 0 & \needToRun & \needToRun & \needToRun & -- & \canonical{\bestResult{70}} & -- & 70 \\
\stillRunning{collatz-numbers} & 0 & -- & -- & -- & -- & -- & \canonical{0} & 0 \\
compare-string-lengths & \canonical{\bestResult{97}} & -- & -- & -- & -- & -- & -- & 97 \\
count-odds & 0 & \canonical{\bestResult{100}} & -- & -- & -- & -- & -- & 100 \\
digits & 0 & -- & -- & -- & \canonical{0} & -- & \needToRun & 0 \\
double-letters & 0 & \canonical{\bestResult{63}} & -- & -- & -- & -- & -- & 63 \\
\stillRunning{even-squares} & 0 & -- & -- & -- & \canonical{\bestResult{38}} & -- & -- & 38 \\
for-loop-index & 77 & -- & -- & -- & \canonical{\bestResult{97}} & -- & -- & 97 \\
grade & 0 & \bestResult{100} & \canonical{97} & -- & -- & -- & -- & 100 \\
last-index-of-zero & 0 & \canonical{\bestResult{93}} & -- & -- & -- & -- & -- & 93 \\
median & \canonical{\bestResult{100}} & -- & -- & -- & -- & -- & -- & 100 \\
mirror-image & \canonical{\bestResult{100}} & -- & -- & -- & -- & -- & -- & 100 \\
negative-to-zero & 0 & \canonical{\bestResult{100}} & -- & -- & -- & -- & -- & 100 \\
number-io & \canonical{\bestResult{100}} & -- & -- & -- & -- & -- & -- & 100 \\
pig-latin & 0 & \canonical{0} & -- & \needToRun & \needToRun & \needToRun & \needToRun & 0 \\
replace-space-with-newline & 0 & \canonical{\bestResult{20}} & -- & -- & -- & -- & -- & 20 \\
scrabble-score & 0 & \canonical{\bestResult{100}} & -- & -- & -- & -- & -- & 100 \\
small-or-large & \canonical{\bestResult{67}} & -- & -- & -- & -- & -- & -- & 67 \\
smallest & \canonical{\bestResult{100}} & -- & -- & -- & -- & -- & -- & 100 \\
string-differences & 0 & \needToRun & \canonical{0} & \needToRun & \needToRun & \needToRun & \needToRun & 0 \\
string-lengths-backwards & 0 & \canonical{\bestResult{100}} & -- & -- & -- & -- & -- & 100 \\
sum-of-squares & 0 & -- & -- & -- & -- & -- & \canonical{\bestResult{30}} & 30 \\
super-anagrams & 0 & \needToRun & \canonical{\bestResult{63}} & -- & -- & -- & -- & 63 \\
syllables & 0 & \canonical{\bestResult{80}} & -- & -- & -- & -- & -- & 80 \\
vector-average  & \bestResult{100} & -- & -- & -- & -- & \canonical{87} & -- & 100 \\
vectors-summed & 0 & \needToRun & \canonical{\bestResult{87}} & -- & -- & -- & -- & 87 \\
\stillRunning{wallis-pi} & 0 & -- & -- & -- & -- & -- & \canonical{0} & 0 \\
\stillRunning{word-stats}  & 0 & \needToRun & -- & \needToRun & -- & \canonical{0} & \needToRun & 0 \\
\stillRunning{x-word-lines} & 0 & \needToRun & -- & \needToRun & \needToRun & \canonical{0} & \needToRun & 0 \\
\midrule
Total canonical & 6 & 9 & 4 & 0 & 3 & 4 & 3 \\
Solved canonical & 6 & 8 & 3 & 0 & 2 & 2 & 1 \\
Solved non-canonical & 2 & 1 & 0 & 0 & 0 & 0 & 0 \\
\bottomrule
\end{tabular}
\end{table}

\begin{table}[tbp]
\caption{
Success Rates of each pattern on \ac{PSB2} for \origamiacdc.}
\label{tab:by_scheme_psb2}
\centering
\rowcolors{2}{gray!25}{white}
\footnotesize
\begin{tabular}{lrrrrrrr|r}
\toprule
\textbf{Dataset} & NoScheme & Cata List & CurriedCata & Histo & Ana & Accu & Hylo & \textbf{Best} \\
\midrule
basement & 0 & \needToRun & -- & \needToRun & -- &  \canonical{\bestResult{40}} & -- & 40 \\
bouncing-balls & 0 & -- & -- & -- & -- & -- & \canonical{0} & 0 \\
bowling & 0 & \needToRun & -- & \needToRun & -- & \needToRun & \needToRun & 0 \\
camel-case & 0 & \canonical{0} & -- & \needToRun & -- & \needToRun & \needToRun & 0 \\
coin-sums & 0 & \needToRun & \canonical{\bestResult{87}} & -- & -- & -- & -- & 87 \\
cut-vector & 0 & \needToRun & -- & \needToRun & -- & \canonical{\needToRun} & \needToRun & 0 \\
dice-game & \canonical{\bestResult{20}} & -- & -- & -- & -- & -- & -- & 20 \\
find-pair & 0 & \needToRun & -- & \canonical{\bestResult{20}} & -- & -- & -- & 20 \\
fizz-buzz & \canonical{\bestResult{87}} & -- & -- & -- & -- & -- & -- & 87 \\
fuel-cost & 0 & \canonical{\bestResult{100}} & -- & -- & -- & -- & -- & 100 \\
gcd & 0 & -- & -- & -- & -- & -- & \canonical{\bestResult{23}} & 23 \\
indices-of-substring & 0 & \needToRun & \needToRun & \canonical{0} & \needToRun & \needToRun & \needToRun & 0 \\
leaders & 0 & \bestResult{40} & -- & \canonical{37} & -- & -- & -- & 40 \\
luhn & 0 & \canonical{0} & -- & \needToRun & -- & \needToRun & \needToRun & 0 \\
mastermind & 0 & \needToRun & \canonical{\needToRun} & \needToRun & -- & \needToRun & \needToRun & 0 \\
middle-character & \canonical{\bestResult{33}} & -- & -- & -- & -- & -- & -- & 33 \\
paired-digits & 0 & \needToRun & -- & \canonical{\bestResult{60}} & -- & -- & -- & 60 \\
shopping-list & 0 & \needToRun & \canonical{0} & \needToRun & -- & \needToRun & \needToRun & 0 \\
snow-day & 0 & -- & -- & -- & -- & -- & \canonical{\bestResult{10}}  & 0 \\
solve-boolean & 0 & 0 & -- & 0 & -- & \needToRun & \needToRun & 0 \\
spin-words & 0 & \needToRun & -- & \needToRun & \needToRun & \canonical{0} & \needToRun & 0 \\
\stillRunning{square-digits} & 0 & -- & -- & -- & \needToRun & -- & \canonical{0} & 0 \\
substitution-cipher & 0 & \needToRun & -- & \needToRun & \needToRun & \needToRun & \needToRun & 0 \\
twitter & 0 & \canonical{0} & -- & \needToRun & \needToRun & \needToRun & \needToRun & 0 \\
vector-distance & 0 & \needToRun & \needToRun & \needToRun & -- & \needToRun & \needToRun & 0 \\
\midrule
Total canonical & 3 & 4 & 3 & 4 & 0 & 3 & 4  \\
Solved canonical & 3 & 1 & 1 & 3 & 0 & 1 & 2  \\
Solved non-canonical & 0 & 1 & 0 & 0 & 0 & 0 & 0  \\
\bottomrule
\end{tabular}

\end{table}

\begin{table}[tbp]
\caption{Success Rates of each pattern on \ac{PolyPSB} for \origamiacdc.}
\label{tab:by_scheme_polypsb}
\rowcolors{2}{gray!25}{white}
\centering
\footnotesize
\begin{tabular}{lrrrrr|r}
\toprule
\textbf{Dataset} & NoScheme & Cata List & Cata Set & Cata Map & Hylo & \textbf{Best} \\
\midrule
area-of-rectangle & \canonical{\bestResult{100}} & -- & -- & -- & -- & 100 \\
centimeters-to-meters & \canonical{\bestResult{97}} & -- & -- & -- & -- & 97 \\
count-true & 0 & \canonical{\bestResult{100}} & -- & -- & -- & 100 \\
filter-bounds & 0 & -- & \canonical{\bestResult{77}} & -- & -- & 77 \\
first-index-of-true & 97 & \canonical{\bestResult{100}} & -- & -- & -- & 100 \\
get-vals-of-key & 0 & \canonical{\bestResult{100}} & -- & -- & -- & 100 \\
max-applied-fn & 0 & -- & -- & -- & \canonical{\bestResult{7}} & 7 \\
min-key & -- & -- & -- & \canonical{\bestResult{97}} & -- & 97 \\
\stillRunning{set-cartesian-product} & \needToRun & -- & \needToRun & -- & \needToRun & 0 \\
set-symmetric-difference & \canonical{\bestResult{100}} & -- & -- & -- & -- & 100 \\
sets-with-element & 0 & -- & \canonical{\bestResult{100}} & -- & -- & 100 \\
simple-encryption & 0 & \canonical{\bestResult{100}} & -- & -- & -- & 100 \\
sum-2-vals & \canonical{\bestResult{100}} & -- & -- & -- & -- & 100 \\
sum-2-vals-polymorphic & \canonical{\bestResult{100}} & -- & -- & -- & -- & 100 \\
sum-2d & \canonical{\bestResult{97}} & -- & -- & -- & -- & 97 \\
sum-vector-vals & 0 & \canonical{\bestResult{100}} & -- & -- & -- & 100 \\
time-sheet & 0 & \canonical{\bestResult{73}} & -- & -- & -- & 73 \\
\midrule
Total canonical & 6 & 6 & 2 & 1 & 1 \\
Solved canonical & 6 & 6 & 2 & 1 & 1 \\
Solved non-canonical & 1 & 0 & 0 & 0 & 0 \\
\bottomrule
\end{tabular}

\end{table}

Overall, \origamiacdc demonstrates that it is able to synthesize solutions using all the provided patterns.

Specifically, it found a solution at least once in all of the $15$ problems that are canonically solved by NoScheme.
Moreover, in $3$ problems (\emph{vector-average}, \emph{for-loop-index}, \emph{first-index-of-true}), a NoScheme solution was found even though it was not the canonical pattern.
This is not completely unexpected, as our goal while writing the canonical solutions was to employ \acp{RS} as much as possible. However, with the extended function set available, sometimes the implicit recursion of other functions (such as \code{sum} and \code{takeWhile}) is enough for the problem to be solved in NoScheme.
Nevertheless, for a more complete assessment of \origamiacdc, it was still evaluated on those problems using their canonical patterns, being able to solve all of them at least once.

For the next pattern in our sequence, Cata, \fracpct{18}{22} of the canonical problems were solved, even for the new data structures (\code{Map} and \code{Set}) in \ac{PolyPSB}.
Like previously described, \emph{grade} and \emph{leaders} were solved using Cata, as well as with their canonical CurriedCata and Histo patterns, respectively.
For the remaining patterns, \origamiacdc solves 
\fracpct{3}{4} of Histo, 
\fracpct{2}{3} of Ana, 
\fracpct{4}{7} of CurriedCata,
\fracpct{3}{7} of Accu,
and \fracpct{3}{8} of the Hylo problems.

There is an apparent correlation between the chosen pattern and its success rate.
In general, some patterns solve fewer problems than others, and also present lower success rates, \ie for the subset of problems they are able to solve, they do so with low consistency.
This trend is supported by the previous assessment of \origamibananas, and can likely be attributed to a combination of two main factors:

\begin{itemize}
    \item \textbf{Inherent problem complexity:} Patterns such as Accu, Ana, and Hylo tend to solve more difficult problems, while NoScheme addresses simpler problems.
    Evidence for this is that other methods in the literature present a similar behavior for those problems (see \autoref{sec:vs_others});
    \item \textbf{Pattern complexity:} Some patterns inherently make solutions harder to evolve.
    For example, CurriedCata solutions require an accumulated function, making programs more sensitive to small changes. 
    Histo involves more bindings, while Accu and Ana have four slots and Hylo has six, which significantly expands the search space. 
    In addition, unlike Cata, Accu, and Histo, which have guaranteed termination, Ana and Hylo can produce infinite recursions, increasing the difficulty of finding correct solutions.
\end{itemize}

\subsection{Ablation Study}
\label{sec:ablation}

In this section, we perform an Ablation Study~\citep{sheikholeslami2019ablation}: a controlled experiment in which components of a system are systematically removed to evaluate their individual impact on overall performance. 
By comparing the $3$ different versions of \origami --- \origamibananas, \origamidsls, and \origamiacdc --- this study allows us to assess the effect of \acdc independently of the other enhancements introduced in this chapter.


The best results obtained for each method in \ac{PSB1} are presented in \autoref{tab:vs_origamis}.
For most of the problems, both \origamiacdc and \origamidsls represent a considerable improvement over \origamibananas, showing the positive overall impact of the new function set and selection method.
Among these, we highlight the improvements on 
\emph{checksum}, 
\emph{count-odds}, 
\emph{double-letters}, 
\emph{even-squares}, 
\emph{replace-space-with-newline}, 
\emph{small-or-large}, 
\emph{syllables}, 
\emph{sum-of-squares}, 
\emph{vector-average} and 
\emph{vectors-summed}.
Specifically, the major difference in performance in \emph{vector-average} stems from the introduction of the \code{sum} operator, making it solvable using NoScheme instead of the canonical Accu.
The only problem with a lower success rate is \emph{super-anagrams}, where \origamiacdc got a marginally lower success rate than \origamibananas, even though \origamidsls presented a considerable improvement.

Comparing the results from \origamiacdc to \origamidsls illustrates the effects of applying \acdc to the evolution. In this benchmark, \origamiacdc improves \fracpct{8}{29} problems, with a mean\footnote{All ratio means we report are computed as geometric means.} improvement of $1.62\times$. 
The most pronounced improvement is \emph{checksum}, whose success rate increased from $23\%$ to $70\%$ (a delta of $47\%$, corresponding to a $3\times$ improvement).  
Conversely, \origamidsls achieves higher success rates in \fracpct{4}{29} problems, specifically \emph{compare-string-lengths}, \emph{replace-space-with-newline}, \emph{super-anagrams}, and \emph{vectors-summed}, with \origamiacdc representing a mean decrease of $0.77\times$.  
As a whole, \origamiacdc yields more improvements than regressions on this dataset, indicating that the addition of \acdc generally benefits the evolutionary process.

\begin{table}[t!]
\caption{Success rates of different versions of \origami in \ac{PSB1}.}
\label{tab:vs_origamis}
\rowcolors{2}{gray!25}{white}
\centering
\footnotesize
\begin{tabular}{lrrr}
\toprule
\textbf{Dataset} & \origamiacdc & \origamidsls & \origamibananas \\
\midrule
checksum & \bestResult{70} & 23 & 0 \\
compare-string-lengths & 97 & \bestResult{100} & 90 \\
count-odds & \bestResult{100} & \bestResult{100} & 40 \\
double-letters & \bestResult{63} & 60 & 3 \\
\stillRunning{even-squares} & \bestResult{38} & 13 & 3 \\
for-loop-index & \bestResult{97} & 90 & 90 \\
grade & \bestResult{100} & \bestResult{100} & \bestResult{100} \\
last-index-of-zero & \bestResult{93} & \bestResult{93} & 70 \\
median & \bestResult{100} & \bestResult{100} & 97 \\
mirror-image & \bestResult{100} & 87 & 93 \\
negative-to-zero & \bestResult{100} & \bestResult{100} & 87 \\
number-io & \bestResult{100} & \bestResult{100} & \bestResult{100} \\
replace-space-with-newline & 20 & \bestResult{33} & 3 \\
scrabble-score & \bestResult{100} & \bestResult{100} & \bestResult{100} \\
small-or-large & \bestResult{67} & 40 & 53 \\
smallest & \bestResult{100} & \bestResult{100} & \bestResult{100} \\
string-lengths-backwards & \bestResult{100} & \bestResult{100} & 97 \\
sum-of-squares & \bestResult{30} & 13 & 0 \\
super-anagrams & 63 & \bestResult{90} & 73 \\
syllables & \bestResult{80} & 73 & 7 \\
vector-average & \bestResult{100} & \bestResult{100} & 0 \\
vectors-summed & 87 & \bestResult{97} & 20 \\
\bottomrule
\end{tabular}

\end{table}

\subsection{Comparison to other GP methods}
\label{sec:vs_others}

To evaluate the performance of \origamiacdc, we compare its results to those of established GP methods from the literature, including:
\begin{itemize}
    \item \textbf{PushGP} variants:
    \begin{itemize}
        \item \textbf{\pushgp}, the baseline algorithm for program synthesis, which evolves programs in the stack-based Push language~\citep{psb1, psb2};
        \item \textbf{\umad}, which replaces its standard mutation with \ac{UMAD}~\citep{umad};
        \item \textbf{\dsls}, which employs \ac{DSLS}~\citep{dsls} (see \autoref{sec:dsls});
    \end{itemize}
    \item \textbf{\ac{CBGP}}, Code-Building Genetic Programming, a typed functional GP system that supports polymorphism and \acp{HOF}~\citep{pantridge2020code,pantridge2022functional,polypsb}.
\end{itemize}


\begin{table}[tbp]
\caption{Success rates of GP Methods in \ac{PSB1}.}
\label{tab:vs_others_psb1}
\rowcolors{2}{gray!25}{white}
\centering
\footnotesize
\begin{tabular}{lrrrrr}
\toprule
 & \origamiacdc & \dsls & \umad & \pushgp & CBGP \\
\midrule
checksum & \bestResult{70} & 18 & 5 & 0 & -- \\
compare-str.-len. & \bestResult{97} & 51 & 42 & 7 & 22 \\
count-odds & \bestResult{100} & 11 & 12 & 8 & 0 \\
digits & 0 & \bestResult{28} & 11 & 7 & 0 \\
double-letters & \bestResult{63} & 50 & 20 & 6 & -- \\
\stillRunning{even-squares} & \bestResult{38} & 2 & 0 & 2 & -- \\
for-loop-index & \bestResult{97} & 5 & 1 & 1 & 0 \\
grade & \bestResult{100} & 2 & 0 & 4 & -- \\
last-index-of-zero & \bestResult{93} & 65 & 56 & 21 & 10 \\
median & \bestResult{100} & 69 & 48 & 45 & 98 \\
mirror-image & \bestResult{100} & 99 & \bestResult{100} & 78 & \bestResult{100} \\
negative-to-zero & \bestResult{100} & 82 & 82 & 45 & 99 \\
number-io & \bestResult{100} & 99 & \bestResult{100} & 98 & \bestResult{100} \\
replace-sp.-with-nl. & 20 & \bestResult{100} & 87 & 51 & 0 \\
scrabble-score & \bestResult{100} & 31 & 20 & 2 & -- \\
small-or-large & \bestResult{67} & 22 & 4 & 5 & 0 \\
smallest & \bestResult{100} & 98 & \bestResult{100} & 81 & \bestResult{100} \\
string-len.-back. & \bestResult{100} & 95 & 86 & 66 & -- \\
sum-of-squares & \bestResult{30} & 25 & 26 & 6 & -- \\
super-anagrams & \bestResult{63} & 4 & 0 & 0 & -- \\
syllables & \bestResult{80} & 64 & 48 & 18 & -- \\
vector-average & \bestResult{100} & 97 & 92 & 16 & 88 \\
vectors-summed & 87 & 21 & 9 & 1 & \bestResult{100} \\
\stillRunning{x-word-lines} & 0 & \bestResult{91} & 59 & 8 & -- \\
\midrule
$\mathbf{= 100\%}$ & \bestResult{10} & 1 & 3 & 0 & 4 \\
$\mathbf{\geq 75\%}$ & \bestResult{15} & 8 & 7 & 3 & 7 \\
$\mathbf{\geq 50\%}$ & \bestResult{20} & 13 & 9 & 5 & 7 \\
$\mathbf{\geq 25\%}$ & \bestResult{21} & 16 & 13 & 7 & 7 \\
$\mathbf{> 0\%}$ & 22 & \bestResult{24} & 21 & 22 & 9 \\
\bottomrule
\end{tabular}

\end{table}

\begin{table}[tbp]
\caption{Success rates of GP Methods in \ac{PSB2}.}
\label{tab:vs_others_psb2}
\centering
\rowcolors{2}{gray!25}{white}
\footnotesize
\begin{tabular}{lrrrr}
\toprule
 & \origamiacdc & \dsls & \pushgp \\
\midrule
basement & \bestResult{40} & 2 & 1 \\
bouncing-balls & 0 & \bestResult{3} & 0 \\
camel-case & 0 & \bestResult{4} & 1 \\
coin-sums & \bestResult{87} & 39 & 2 \\
dice-game & \bestResult{20} & 1 & 0 \\
find-pair & \bestResult{20} & \bestResult{20} & 4 \\
fizz-buzz & \bestResult{87} & 74 & 25 \\
fuel-cost & \bestResult{100} & 67 & 50 \\
gcd & \bestResult{23} & 20 & 8 \\
indices-of-substring & 0 & \bestResult{4} & 0 \\
leaders & \bestResult{40} & 0 & 0 \\
middle-character & 33 & \bestResult{79} & 57 \\
paired-digits & \bestResult{60} & 17 & 8 \\
snow-day & \bestResult{10} & 7 & 4 \\
solve-boolean & 0 & \bestResult{5} & \bestResult{5}  \\
\stillRunning{square-digits} & 0 & 2 & 0 \\
substitution-cipher & 0 & \bestResult{86} & 60 \\
twitter & 0 & \bestResult{52} & 31 \\
\midrule
$\mathbf{= 100\%}$ & \bestResult{1} & 0 & 0 \\
$\mathbf{\geq 75\%}$ & \bestResult{3} & 2 & 0 \\
$\mathbf{\geq 50\%}$ & 4 & \bestResult{5} & 3 \\
$\mathbf{\geq 25\%}$ & \bestResult{7} & 6 & 5 \\
$\mathbf{> 0\%}$ & 11 & \bestResult{17} & 13 \\
\bottomrule
\end{tabular}
\end{table}

\begin{table}[tbp]
\caption{Success rates of GP Methods in \ac{PolyPSB}.}
\label{tab:vs_others_polypsb}
\centering
\rowcolors{2}{gray!25}{white}
\footnotesize
\begin{tabular}{lrr}
\toprule
 & \origamiacdc & CBGP \\
\midrule
area-of-rectangle & \bestResult{100} & 59 \\
centimeters-to-meters & \bestResult{97} & 92 \\
count-true & \bestResult{100} & \bestResult{100} \\
filter-bounds & \bestResult{77} & 13 \\
first-index-of-true & \bestResult{100} & \bestResult{100} \\
get-vals-of-key & \bestResult{100} & 12 \\
max-applied-fn & 7 & \bestResult{24} \\
min-key & \bestResult{97} & 31 \\
set-symmetric-difference & \bestResult{100} & 50 \\
sets-with-element & \bestResult{100} & 4 \\
simple-encryption & \bestResult{100} & 96 \\
sum-2-vals & \bestResult{100} & 94 \\
sum-2-vals-polymorphic & \bestResult{100} & \bestResult{100} \\
sum-2d & 97 & \bestResult{100} \\
sum-vector-vals & \bestResult{100} & 16 \\
time-sheet & \bestResult{73} & 2 \\
\midrule
$\mathbf{= 100\%}$ & \bestResult{10} & 4 \\
$\mathbf{\geq 75\%}$ & \bestResult{14} & 7 \\
$\mathbf{\geq 50\%}$ & \bestResult{15} & 9 \\
$\mathbf{\geq 25\%}$ & \bestResult{15} & 10 \\
$\mathbf{> 0\%}$ & \bestResult{16} & \bestResult{16} \\
\bottomrule
\end{tabular}
\end{table}

\autoref{tab:vs_others_psb1}, \autoref{tab:vs_others_psb2}, \autoref{tab:vs_others_polypsb} show the results when comparing to other state-of-the-art methods.


Across all $3$ benchmarks, \origamiacdc has the highest $= 100\%$, $\geq75\%$, and $\geq25\%$ rates when compared to the other methods.
It also has the highest $\geq50\%$ rates on \ac{PSB1} and \ac{PolyPSB}, and is second to \dsls on \ac{PSB2}.
For the $>0\%$ category, it ties with CBGP on \ac{PolyPSB}, and is second to \dsls on both \ac{PSB1} and \ac{PSB2}.


Overall, in \fracpct{13}{71} of the problems across all benchmarks, \origamiacdc is the only method to obtain $100\%$ success rate:
\begin{itemize*}[label={}, itemjoin={,}, itemjoin*={, and}]
    \item \emph{count-odds}%
    \item \emph{grade}%
    \item \emph{scrabble-score}%
    \item \emph{string-lengths-backwards}%
    \item \emph{vector-average}%
    \item \emph{fuel-cost}%
    \item \emph{area-of-rectangle}%
    \item \emph{get-vals-of-key}%
    \item \emph{set-symmetric-difference}%
    \item \emph{sets-with-element}%
    \item \emph{simple-encryption}%
    \item \emph{sum-2-vals}%
    \item \emph{sum-vector-vals}%
\end{itemize*}%
.
Moreover, \origamiacdc is the only method to achieve $100\%$ success rate on a \ac{PSB2} problem, in \emph{fuel-cost}.


In \ac{PSB1}, \origamiacdc achieves the highest success rate (including ties) in \fracpct{25}{29} problems, with lower rates only in \emph{digits}, \emph{replace-space-with-newline}, \emph{vectors-summed}, and \emph{x-word-lines}.  
In \ac{PSB2}, it achieves the highest success rate (including ties) in \fracpct{10}{25} problems, with \dsls obtaining higher rates in $8$ problems.
The remaining $7$ problems have not been solved by any method to date.
In \ac{PolyPSB}, it achieves the highest success rate (including ties) in \fracpct{15}{17} problems, as CBGP has the highest rate in \emph{sum-2d} and \emph{max-applied-fn}.


Across the three benchmarks, \origamiacdc strictly outperformed all other methods in \fracpct{37}{71} problems, with an average improvement of $3.35\times$ over the second-best method (excluding the \emph{leaders} problem, in which \origamiacdc is the only method to achieve a nonzero success rate).  
In \fracpct{20}{71} problems, it tied with other approaches as the best performer, showing no improvement. 
This group includes the \emph{find-pair} problem (where \origamiacdc tied with \dsls at a $20\%$ success rate), the $13$ problems that were not solved by any method, and $6$ problems in which \origamiacdc and at least one other method both achieved $100\%$ success rates.  
\origamiacdc was outperformed in \fracpct{14}{71} problems. 
Among these, $9$ problems were solved by other methods but not by \origamiacdc. 
In the $5$ problems \origamiacdc solved but did not achieve the highest success rate, the top method obtained a $2.19\times$ higher success rate on average.

\subsection{Comparison against LLMs}

\acp{LLM} have been recently tested in many variations of the \ac{PS} task. 
In one of such works~\citep{copilot}, GitHub Copilot is evaluated in solving both \ac{PSB1} and \ac{PSB2}. 
However, comparing these results to \origamiacdc is not straightforward, as we need to consider the risk of \emph{data contamination}, and the fact that Copilot has access to the problem textual description rather than input-output examples.

First, data contamination means that the benchmarks used for the evaluation, as well as their intended solutions, may have been included in the corpora used to train these LLMs.
Thus, their performance might simply reflect the memorization of such solutions, rather than a genuine capability of solving novel tasks in a generalizable fashion.
This concern is supported by the literature, with evidence that LLM performances tend to be higher on problems that were public before the training data cutoff date~\citep{liu2024no}.
As the information regarding the dataset used for training these models is not publicly available, there is no way to verify whether or not that is the case for GitHub Copilot and these benchmarks.

Secondly, the fundamental difference in the \emph{type of information} each method receives makes them arguably operate on different tasks.
\origami, as a traditional PBE algorithm, receives a set of structured input-output examples.
In contrast, LLMs operate on natural language and therefore need a textual description of the problem.
These two types of specification are not only distinct in format, but also in the actual task they describe.
A textual description often contains implicit context and information that is not available in the input-output pairs, which LLMs can effectively leverage when synthesizing the program.
On the other hand, input-output examples offer a low-level, more detailed description of the expected behavior, which is especially valuable in edge cases.

With those challenges in mind, a comparison between \origamiacdc and Copilot is shown in \autoref{tab:vs_copilot_psb1} and \autoref{tab:vs_copilot_psb2}.

%

\begin{landscape}
\begin{minipage}{0.45\textwidth}
\captionof{table}{Success rates of \origamiacdc and Copilot in \ac{PSB1}.}
\label{tab:vs_copilot_psb1}
\centering
\rowcolors{2}{gray!25}{white}
{\footnotesize
\renewcommand*{\arraystretch}{0.75}
\begin{tabular}{lrr}
\toprule
 & \origamiacdc & Copilot \\
\midrule
checksum & 70 & \bestResult{89} \\
collatz-numbers & 0 & \bestResult{73} \\
compare-string-lengths & \bestResult{97} & 70 \\
count-odds & \bestResult{100} & 98 \\
double-letters & 63 & \bestResult{88} \\
\stillRunning{even-squares} & \bestResult{38} & 11 \\
for-loop-index & \bestResult{97} & 72 \\
grade & \bestResult{100} & 84 \\
last-index-of-zero & \bestResult{93} & 61 \\
median & \bestResult{100} & 79 \\
mirror-image & \bestResult{100} & 70 \\
negative-to-zero & \bestResult{100} & 99 \\
number-io & \bestResult{100} & 93 \\
pig-latin & 0 & \bestResult{54} \\
replace-space-with-newline & 20 & \bestResult{87} \\
scrabble-score & \bestResult{100} & 35 \\
small-or-large & \bestResult{67} & 51 \\
smallest & \bestResult{100} & 66 \\
string-differences & 0 & \bestResult{9} \\
string-lengths-backwards & \bestResult{100} & 60 \\
sum-of-squares & 30 & \bestResult{90} \\
super-anagrams & \bestResult{63} & 55 \\
syllables & 80 & \bestResult{96} \\
vector-average & \bestResult{100} & 92 \\
vectors-summed & 87 & \bestResult{87} \\
x-word-lines & 0 & \bestResult{1} \\
\midrule
$\mathbf{= 100\%}$ & \bestResult{11} & 0 \\
$\mathbf{\geq 75\%}$ & \bestResult{16} & 12 \\
$\mathbf{\geq 50\%}$ & 20 & \bestResult{22} \\
$\mathbf{\geq 25\%}$ & 21 & \bestResult{23} \\
$\mathbf{> 0\%}$ & 22 & \bestResult{26} \\
\bottomrule
\end{tabular}
}
%
\end{minipage}
\hfill
\begin{minipage}{0.45\textwidth}
\captionof{table}{Success rates of \origamiacdc and Copilot in \ac{PSB2}.}
\label{tab:vs_copilot_psb2}
\centering
\rowcolors{2}{gray!25}{white}
{\footnotesize
\renewcommand*{\arraystretch}{0.75}
\begin{tabular}{lrr}
\toprule
 & \origami & Copilot \\
\midrule
basement & 40 & \bestResult{95} \\
camel-case & 0 & \bestResult{31} \\
coin-sums & \bestResult{87} & 12 \\
cut-vector & 0 & \bestResult{1} \\
dice-game & \bestResult{20} & 0 \\
find-pair & 20 & \bestResult{41} \\
fizz-buzz & 87 & \bestResult{89} \\
fuel-cost & \bestResult{100} & 97 \\
gcd & 23 & \bestResult{80} \\
indices-of-substring & 0 & \bestResult{82} \\
leaders & 40 & \bestResult{67} \\
luhn & 0 & \bestResult{6} \\
middle-character & 33 & \bestResult{98} \\
paired-digits & 60 & \bestResult{88} \\
shopping-list & 0 & \bestResult{75} \\
snow-day & \bestResult{10} & \bestResult{10} \\
spin-words & 0 & \bestResult{96} \\
\stillRunning{square-digits} & 0 & \bestResult{55} \\
substitution-cipher & 0 & \bestResult{78} \\
twitter & 0 & \bestResult{89} \\
vector-distance & 0 & \bestResult{79} \\
\midrule
$\mathbf{= 100\%}$ & \bestResult{1} & 0 \\
$\mathbf{\geq 75\%}$ & 3 & \bestResult{12} \\
$\mathbf{\geq 50\%}$ & 5 & \bestResult{14} \\
$\mathbf{\geq 25\%}$ & 7 & \bestResult{16} \\
$\mathbf{> 0\%}$ & 9 & \bestResult{20} \\
\bottomrule
\end{tabular}
}
%
\end{minipage}
\end{landscape}

The results in \ac{PSB1} show that the performance of both algorithms is broadly comparable, with a similar number of problems solved in the $>0\%$, $\geq25\%$, $\geq50\%$, and $\geq75\%$ categories. 
The main distinction appears in the $=100\%$ category, where \origamiacdc achieves perfect success rates in 11 problems, while Copilot does not reach $100\%$ in any case.
Similarly, in \ac{PSB2} benchmark, Copilot achieves higher counts in most categories ($>0\%$, $\geq25\%$, $\geq50\%$, and $\geq75\%$) while \origamiacdc leads only in the $=100\%$.

Overall, a larger number of problems are solved at least once ($>0\%$) by Copilot~(\fracpct{46}{54}) than \origamiacdc~(\fracpct{31}{54}), across both benchmarks. 
However, Copilot is not able to solve any problem with $100\%$ success rate.
This indicates that its higher problem coverage comes with less reliability, as it occasionally fails to produce a solution even for simple problems.

\begin{figure}[tb]
    \centering
    \begin{subfigure}{.48\linewidth}
        \includegraphics[width=\linewidth]{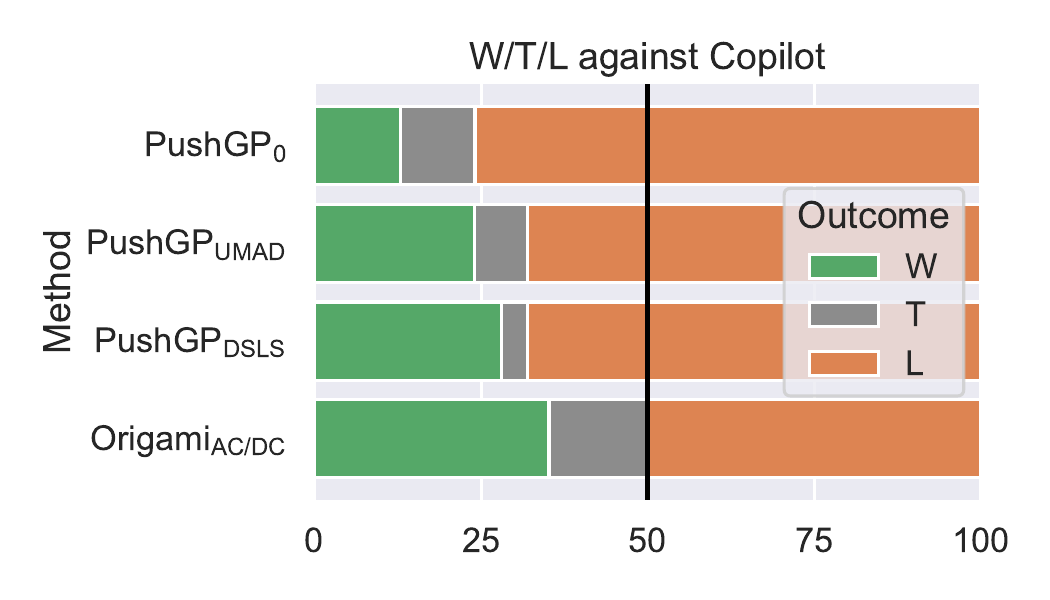}
        \caption{Across \ac{PSB1} and \ac{PSB2}}
    \end{subfigure}
    \hfill
    \begin{subfigure}{.48\linewidth}
        \includegraphics[width=\linewidth]{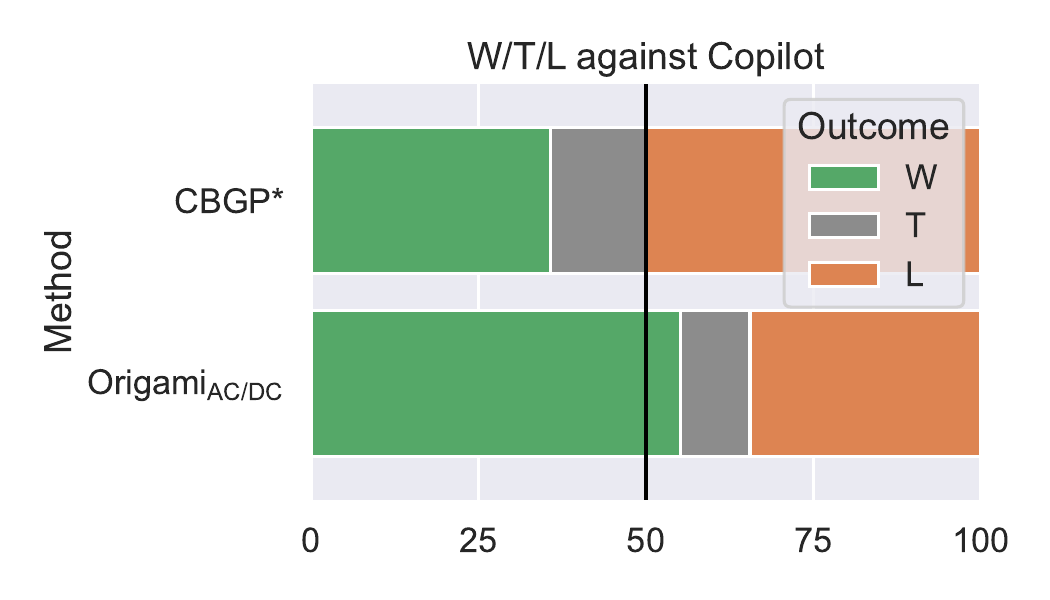}
        \caption{Across \ac{PSB1}}
        \label{fig:wtl_copilot_psb1}
    \end{subfigure}
    
    \caption{Percentage of problems won (W), tied (T), and lost (L) by each method compared to Copilot, across \ac{PSB1} and \ac{PSB2}. 
    Here, W indicates that a given method achieved a higher success rate than Copilot, T indicates equal success rates, and L indicates a lower success rate.  
    CBGP* indicates partial results, as it was evaluated only on \ac{PSB1}.
    A reference vertical line is drawn at 50\%.
    }
    \label{fig:all_vs_copilot}
\end{figure}

We also consider which method has a higher success rate on a problem-by-problem basis.
\autoref{fig:all_vs_copilot} presents a comparison of \origamiacdc and other GP methods against Copilot.

Considering both benchmarks, \origamiacdc 
wins against Copilot in \fracpct{19}{54},
ties in \fracpct{8}{54},
and loses in \fracpct{27}{54}.
This represents the highest win rate among the methods evaluated on both \ac{PSB1} and \ac{PSB2}.

As \ac{CBGP} was only evaluated on \ac{PSB1}, which was shown to be the less challenging of the two, \autoref{fig:wtl_copilot_psb1} shows an isolated comparison on just this benchmark.
In this scenario, \origamiacdc achieves a W/T/L record of $16/3/10$ (ties are omitted in the table when both methods scored 0).
This corresponds to a $55\%$ win rate over Copilot, which is higher than \ac{CBGP} under the same conditions.
In contrast, \origamiacdc achieves $3/17/5$ against Copilot on \ac{PSB2}, or a $12\%$ win rate, indicating that Copilot performs strongly on this harder benchmark.

\section{Final Remarks}
\label{sec:acdc_outro}

This main contribution of this chapter is a novel procedure called \acf{AC/DC}.
This procedure aims at a more efficient exploration of the search space, as it leads to a more diverse population and creates more opportunities for mutation and crossover.
This chapter also continues the development of \origami, by applying the \ac{DSLS} selection method, expanding the set of functions available, and introducing $3$ new patterns that allow it to synthesize a new \ac{RS} (histomorphism), as well as variations of catamorphism for Maps and Sets.

We evaluated this improved version of \origami (\origamiacdc) in $3$ different benchmarks, two of which were not considered in previous work.
Overall, \origamiacdc demonstrates that it is able to synthesize solutions using all the provided patterns.
We observed a strong correlation between the canonical scheme used by each problem and whether \origamiacdc can solve them at least once, as well as its success rate.
Most notably, \origamiacdc is able to solve all problems that canonically use NoScheme, and solves $\geq50\%$ of the problems in all other \acp{RS}.

The changes brought by this chapter also achieved a considerable improvement over the previous version of \origami (\origamibananas), having a higher success rate in all of the \ac{PSB1} problems.
\acdc also led to meaningful improvements in several problems in \ac{PSB1}, with a very reduced number of regressions.

When compared to the state of the art of \ac{GP}, \origamiacdc achieved the highest number of problems solved with success rates of $=100\%$, $\geq75\%$, and $\geq25\%$ on all benchmarks, and is either tied or in second place in $>0\%$.
In $18\%$ of the problems across the entire benchmark, \origamiacdc is the only method to obtain $100\%$ success rate, being the first known method to achieve this in a \ac{PSB2} problem.

\origamiacdc also presents a performance that is competitive to state-of-the-art LLMs, such as GitHub Copilot, especially in \ac{PSB1}, with the highest overall win-rate over Copilot when compared to other \ac{GP} methods.

For future work, we believe that a deeper investigation into some of the \ac{PSB2} problems might lead to insights into what might be preventing \origami from finding solutions more frequently, or from finding solutions at all.
These could result in strategies to improve the overall success rate of \origami across all benchmarks and patterns.

Another current limitation is that \origamiacdc goes through the \acp{RS} in a linear fashion, exhausting all the allowed runs in a pattern before advancing to the next one.
This could be done more effectively by, for example, applying heuristics that would bias the search towards the most likely pattern.
Or, more interestingly, applying a co-evolutionary approach, with one population per pattern being evolved in parallel.
Exchanging genetic material among the populations has the potential to make the search more effective overall, constituting a promising avenue of work.

%% file: text/outro.tex
\chapter{Conclusion}
\label{cha:outro}

This text aimed to contribute mainly to the fields of \ac{PS}, \ac{GP} and \ac{FP}.
Its main objective was applying \ac{FP} concepts with strong types, including Recursion Schemes, to \ac{PS}.
This was motivated by the benefits this combination brings to the table, which could potentially lead to improvements in the \ac{PS} process.
Therefore, in \autoref{cha:intro}, we defined the following as our problem statement.

\textbf{Problem Statement:}\emph{
    \ac{PS} poses challenges in terms of complexity and cost, primarily attributed to the vast search space involved.
    This study assesses whether the integration of functional programming principles, strong typing, and recursion schemes can mitigate these challenges, potentially leading to a more efficient synthesis, and identifies  potential drawbacks associated with this approach.
}

In order to properly contextualize this work, we presented a general theoretical background of the aforementioned fields.
We conducted a literature review, describing related work and detailing well-known methods which have been largely influential, identifying potential improvement opportunities.
Then, two novel \ac{PS} methods were proposed.

\ac{HOTGP} extends on the standard \ac{GP} algorithm by employing $\lambda$-functions and higher-order functions.
These abstractions are native to \ac{FP} and enable the algorithm to synthesize short solutions to problems that would require increased complexity in an imperative setting.
Furthermore, \ac{HOTGP} has a simplification procedure which leads to shorter programs, which are both more readable and generalize better.
We positioned this method with the current literature by evaluating it and comparing it to state-of-the-art methods.
Overall, \ac{HOTGP} obtained good results, being competitive with the literature and showing considerable improvements in certain problems.
The synthesized solutions were comparable to the hand-crafted ones, in terms of size and complexity.

Our second proposed method is called Origami, and explores the knowledge of the already established and deeply researched Recursion Schemes field.
We argue that, by having access to a handful of pre-established patterns, an algorithm would be able to navigate the search space of recursive programs more effectively, providing a better way of synthesizing functional programs.

We first conducted an analysis of the most widely used \ac{PS} benchmark, \ac{PSB1}~\citep{psb1}, by manually implementing solutions using Recursion Schemes and were able to identify recurrent patterns on the solutions.
We conducted initial experiments by adapting \ac{HOTGP} to evolve the non-constant part of the most common pattern, catamorphism, in order to assess the validity of this algorithm.
The preliminary results show that, when using the correct pattern, an algorithm can quickly synthesize a correct solution.

Following, we presented the first complete implementation of Origami, using a multi-gene approach that uses patterns and slots.
We evaluated our approach in \ac{PSB1}, finding that, in general, Origami performs better than other similar methods, synthesizing the correct solution more often than other methods in most problems.
It was also able to obtain the highest count of problems with success rate $=100\%$, $\geq75\%$ and $\geq50\%$ among the \ac{GP} methods.
It also achieved comparable results to Github Copilot, solving some problems that the \ac{LLM} could not solve. 

Finally, we also proposed a novel procedure called \ac{AC/DC}.
This procedure aims at a more efficient exploration of the search space, as it leads to a more diverse population and creates more opportunities for mutation and crossover.
We applied this procedure to Origami, and also continued its development by applying the \ac{DSLS} selection method, expanding the set of functions available, and introducing $3$ new patterns that allow it to synthesize a new RS (histomorphism), as well as variations of catamorphism for Maps and Sets.
This improved version (\origamiacdc) was evaluated in $3$ different benchmarks, two of which were not considered previously.
Experimental results show that, together, these advancements synthesize solutions using all provided RSs, achieving a considerable improvement over its previous version by raising success rates on every problem. 
Compared to other similar methods in the literature, it has the highest count of problems solved with success rates of $100\%$, $\geq75\%$, and $\geq25\%$ across all benchmarks. 
In $18\%$ of all benchmark problems it stands as the only method to reach $100\%$ success rate, being the first known approach to achieve it on any problem in the General Program Synthesis Benchmark 2.
It also demonstrates competitive performance to \acp{LLM}, achieving the highest overall win-rate against Copilot among all GP methods.

\section{Limitations}

This section addresses some current limitations of Origami, which we believe warrant more research and could potentially lead to meaningful scientific contributions.


Currently, both implementations of Origami go through the RSs in a linear fashion, exhausting all the allowed runs in a pattern before advancing to the next one.
This could be done more effectively by, for example, applying heuristics that would bias the search towards the most likely pattern.
Or, more interestingly, by applying a co-evolutionary approach.

In terms of co-evolution, one option is to have one population for each \emph{slot} that we want to evolve.
To exemplify, consider we want to evolve a solution using catamorphism and hylomorphism.
These two schemes share the same evolvable expressions for the algebra, so it could be evolved once and evaluated in both schemes at the same time.

Another approach would be to keep the multi-gene representation, and provide each \emph{pattern} with its own dedicated population, in which individuals will compete amongst themselves and be evaluated with the same fitness function.

Regardless, here we envision a significant opportunity to also share genetic information between different populations.
For instance, crossover could be applied in two individuals of different populations.
Additionally, some patterns only require minor adaptations to be converted into other patterns.
For example, we can easily convert an accumulation into a catamorphism by discarding the \code{st} expressions, and replacing any reference to the second parameter of \code{alg} with freshly generated expressions, and a catamorphism into an accumulation by generating expressions for \code{st}, and injecting the second parameter of \code{alg} into some part of the expression.
We can also trivially combine a catamorphism and an anamorphism into a hylomorphism, given that their types match.
All of these are ideas that could lead to interesting results, and can only be explored by an algorithm that employs these \acp{RS}.

Additionally, in the text, we have already shown that, from the types of the problem alone, we can discard some patterns that will never be able to provide results of the desired type.
However, more information is also available in the example set itself, which can allow us to further narrow down some patterns.
For example, a problem of type \code{[a] -> [b]} can be solved by catamorphism, but could be solved by other schemes.
However, if we verify that every output has the same length of the input, this is a strong indicator that we want the catamorphism, specifically one that represents a \code{map} operation.
Similarly, in a problem of type \code{[a] -> [a]}, if the elements are always in the same order, and the output is always equal or shorter than the input, this could also indicate that we want a catamorphism that represents a \code{filter} operation.
These are invariants that can be derived from the dataset itself, and could be used to bias the search in such a way that it searches for certain schemes more than others.

Also from the training dataset, we could potentially derive information that trivializes the search for specific slots.
For example, suppose we are searching for a Cata solution, and the examples provide the expected output for the empty list.
We can always guarantee that, if a solution that uses the Cata pattern exists, its expression for the \code{NilF} slot must be the output for the empty list.
In this situation, as this output is already provided by the dataset, the search can focus just on the \code{ConsF}, effectively reducing the search space in half.
The exact same reasoning can also be applied for Accu and Histo.


An additional limitation is that Origami does not address the issue of making sure that a co-algebra, such as those used in anamorphism or in hylomorphism, terminates.
We currently impose a fixed arbitrary memory consumption limit and discard solutions as soon as they exceed it.
In general, as termination is undecidable, there will always be instances where it is impossible to determine whether a program halts. 
Nevertheless, it may be possible to identify specific cases where termination can be formally proved or disproved, or at the very least apply heuristics that reduce the frequency of non-terminating programs.

Another idea that could improve on how quickly we find solutions is that some constants are more often used than others, such as \code{0}, \code{1}, \code{-1}, \code{""}, \code{[]}, etc. 
This knowledge could be incorporated into the GP process, by prioritizing those constants and trying all of them before moving to generating a random constant.

Another potential avenue of work is Interactive Synthesis \citet{helmuth2023human}, in which we start from a largely reduced number of examples.
As the synthesis progresses and a solution is found, the synthesizer asks the user to check if the program generalized correctly by generating unseen examples.
These examples are generated heuristically, aiming to allow the user to quickly identify if that is the correct program or not.
If the examples show that the solution is actually incorrect, they are included in the train set and the search continues.
While this is not a problem that literature benchmarks have, as the scarcity of examples is artificial, such a system would take Origami to a production-ready level.

%% file: Apendices/origami_solutions.tex



\chapter{Canonical solutions to PSB1 problems}\label{cha:origami_solutions}

Here we present the complete set of solutions to the \ac{PSB1} benchmark.
All of these solutions were written by the authors in order to get a better understanding of the recurrent patterns that could be used to solve the benchmark.

\section{No recursion}

In this section we outline the solutions to the easiest problems that do not require any recursive scheme. We should note that by wrapping the inputs of these problems in a list with a single element, we could fit them into the discussed recursive patterns. However, it would just add unnecessary complications to the code.

Even though the solutions to these problems are trivial,  for the sake of completeness, we give the description of each problem followed by their solutions. By their simplicity, these solutions dispense the need for additional explanations.

\subsection{Number IO} 
\emph{Given an integer and a float, print their sum.}

\begin{minted}[escapeinside=@@]{haskell}
    -- required primitives: fromIntegral, (+)
    numberIO :: Int -> Double -> Double
    numberIO x y = fromIntegral x + y
\end{minted}

\subsection{Small or Large} 
\emph{Given an integer n, print “small” if n < 1000 and “large” if n >= 2000 (and nothing if 1000 <= n < 2000).}

\begin{minted}[escapeinside=@@]{haskell}
    -- required primitives: multi-way if
    -- provided primitives: (< 1000), (>= 2000),
    -- "small", "large"
    smallOrLarge :: Int -> String
    smallOrLarge x = if | (< 1000) x -> "small"
                        | (>= 2000) x -> "large"
                        | otherwise -> ""
\end{minted}

\subsection{Compare String Lengths} 
\emph{Given three strings n1, n2, and n3, return true if length(n1) < length(n2) < length(n3), and false otherwise.}

\begin{minted}[escapeinside=@@]{haskell}
    -- required primitives: length, (<), (&&)
    strLengths :: String -> String -> String -> Bool
    strLengths a b c = length a < length b
                     && length b < length c
\end{minted}

\subsection{Median}
\emph{Given 3 integers, print their median.}

\begin{minted}[escapeinside=@@]{haskell}
    -- required primitives: min, max, (+), (-)
    medianOfThree :: Int -> Int -> Int -> Int
    medianOfThree a b c = (a + b + c) - (min a (min b c))
                        - (max a (max b c))
\end{minted}

\subsection{Smallest} 
\emph{Given 4 integers, print the smallest of them.}

\begin{minted}[escapeinside=@@]{haskell}
    -- required primitives: min
    smallest :: Int -> Int -> Int -> Int -> Int
    medianOfThree a b c d = min a (min b (min c d))
\end{minted}

\section{Catamorphisms}

Here we report the problems that were solved by using catamorphism.
The next three problems have already been presented and dissected in the text.
For the sake of completeness, we replicate them here:

\subsection{Count Odds} 
\emph{Given a vector of integers, return the number of integers that are odd, without use of a specific even or odd instruction (but allowing instructions such as modulo and quotient).}

\begin{minted}[escapeinside=@@]{haskell}
    -- required primitives: constant int, mod, +
    countOdds :: [Int] -> Int
    countOdds = cata alg . fromList where
        alg NilF = @\evolv{0}@
        alg (ConsF x acc) = @\evolv{mod x 2 + acc}@
\end{minted}

\subsection{Double Letters} 
\emph{Given a string, print (in our case, return) the string, doubling every letter character, and tripling every exclamation point. All other non-alphabetic and non-exclamation characters should be printed a single time each.}
\begin{minted}[escapeinside=@@]{haskell}
    -- required primitives: if-then-else, (<>), ([])
    -- user provided: constant '!', constant "!!!", isLetter
    doubleLetters :: String -> String
    doubleLetters = cata alg . fromList where
        alg NilF         = @\evolv{[]}@
        alg (ConsF x xs) = @\evolv{x == \chLit!}@
                              @\evolv{then \stLit{!!!} <> xs}@
                              @\evolv{else if isLetter x then [x,x] <> xs}@
                                              @\evolv{else x:xs}@

\end{minted}

\subsection{Super Anagrams} 
\emph{Given strings x and y of lowercase letters, return true if y is a super anagram of x, which is the case if every character in x is in y. To be true, y may contain extra characters, but must have at least as many copies of each character as x does.}

\begin{minted}[escapeinside=@@]{haskell}
    -- required primitives: delete, constant bool
    -- elem, (&&)
    superAnagram :: String -> (String -> Bool)
    superAnagram = cata alg . fromList where
        alg NilF = \ys -> @\evolv{True}@
        alg (ConsF x f) = \ys -> @\evolv{elem x ys \&\& f (delete x ys)}@
\end{minted}

\subsection{String Lengths Backwards}
\emph{Given a vector of strings, print the length of each string in the vector starting with the last and ending with the first.}

\begin{minted}[escapeinside=@@]{haskell}
    -- required primitives: length, snoc
    strLenBack :: [String] -> [Int]
    strLenBack = cata alg . fromList where
        alg NilF = @\evolv{[]}@
        alg (ConsF x xs) = @\evolv{snoc (length x) xs}@
\end{minted}

The initial condition is simply the empty list and we simply append the length of each element to the end of the resulting list (\code{snoc}).

\subsection{Negative To Zero}
\emph{Given a vector of integers, return the vector where all negative integers have been replaced by 0.}

\begin{minted}[escapeinside=@@]{haskell}
    -- required primitives: max, (:), int constants
    negativeToZero :: [Int] -> [Int]
    negativeToZero = cata alg . fromList where
        alg NilF = @\evolv{[]}@
        alg (ConsF x xs) = @\evolv{max 0 x : xs}@
\end{minted}

The initial condition is the neutral element of a list and we just cons (\code{:}) maximum between 0 and each element to the resulting list.

\subsection{Scrabble Score}
\emph{Given a string of visible ASCII characters, return the Scrabble score for that string. Each letter has a corresponding value according to normal Scrabble rules, and non-letter characters are worth zero.}

\begin{minted}[escapeinside=@@]{haskell}
    -- required primitives: findMap, constant int, (+)
    -- user provided primitives: scores, a hash table
    -- mapping the characters to their scores
    scrabbleScore :: String -> Int
    scrabbleScore = cata alg . fromList where
       alg NilF = @\evolv{0}@
       alg (ConsF x xs) = @\evolv{getScore x + xs}@
\end{minted}

This solution assumes that the user can provide a function that returns the correspondence between any character to a score. 

\subsection{Mirror Image} 
\emph{Given two vectors of integers, return true if one vector is the reverse of the other, and false otherwise.}

\begin{minted}[escapeinside=@@]{haskell}
    -- required primitives: init, last, null, not, (&&), (==)
    mirrorImage :: [Int] -> ([Int] -> Bool)
    mirrorImage = cata alg . fromList where
        alg NilF ys = @\evolv{null ys}@
        alg (ConsF x xs) ys = @\evolv{(not.null) ys \&\& last ys == x \&\& xs (init ys)}@
\end{minted}
For this problem we must return a function that takes the second list and returns a boolean. As such, for the \code{NilF} case we return whether the argument is null signaling that all of the elements were consumed. In the \code{ConsF} case we must check if we still have elements to consume, compare the last element to \code{x} and apply \code{xs} to the list \code{ys} without the last value.

\subsection{Vectors Summed}
\emph{Given two equal-sized vectors of integers, return a vector of integers that contains the sum of the input vectors at each index.}

\begin{minted}[escapeinside=@@]{haskell}
    -- required primitives: tail, head, (+), (:)
    sumOfVecs :: [Int] -> ([Int] -> [Int])
    sumOfVecs = cata alg . fromList where
        alg NilF ys = @\evolv{[]}@
        alg (ConsF x xs) ys = @\evolv{if null ys then []}@
                                 @\evolv{else (x + head ys : xs (tail ys))}@
\end{minted}
When there is no element to consume from the first list, we simply return the empty list, as there is nothing to add. When we do have at least one element to consume, we check if we still have an element of the second list, if we do, we add those two together and concatenate to the remainder of the list.

\subsection{Grade} 
\emph{Given 5 integers, the first four represent the lower numeric thresholds for achieving an A, B, C, and D, and will be distinct and in descending order. The fifth represents the student’s numeric grade. The program must print Student has a X grade., where X is A, B, C, D, or F depending on the thresholds and the numeric grade.}

\begin{minted}[escapeinside=@@]{haskell}
    -- required primitives: constant "ABCDF", if-then-else
    -- (<), head, tail.
    grade :: [(Double, Char)] -> (Double -> Char)
    grade = cata alg . fromList where
        alg NilF n = @\evolv{\chLit{F}}@
        alg (ConsF x xs) n = @\evolv{if n >= fst x then snd x else xs n}@
\end{minted}
Assuming we provide a list of tuples where the first element is the threshold and the second is the grade, we can traverse this structure and compare the second argument value with the threshold, if it is within this bound, we return the corresponding grade, otherwise we keep processing. The empty case means that the only option is to fail the student with an `F'.

\section{Anamorphisms}

Here we report the problems that were solved by using anamorphism.

\subsection{For Loop Index}
\emph{Given 3 integer inputs start, end, and step, print the integers in the sequence $n_0$ = start, $n_i$ = $n_i-1$ + step for each $n_i$ < end, each on their own line.}

\begin{minted}[escapeinside=@@]{haskell}
    -- required primitives: (==), (+)
    forLoopIndex :: Int -> Int -> Int -> [Int]
    forLoopIndex start end step = toList (ana coalg @\evolv{start}@) where
        coalg seed = case @\evolv{seed == end}@ of
                       True -> NilF
                       False -> ConsF seed @\evolv{(seed + step)}@
\end{minted}

This is already explained in depth in the text.

\subsection{Digits} 
\emph{Given an integer, print that integer’s digits each on their own line starting with the least significant digit. A negative integer should have the negative sign printed before the most significant digit.}

\begin{minted}[escapeinside=@@]{haskell}
    -- required primitives: constant int, (==)
    -- abs, (<), rem, quot, if-then-else
    digits :: Int -> [Int]
    digits x = toList $ ana coalg @\evolv{x}@ where
        coalg x =
          case @\evolv{x == 0}@ of
            True -> NilF
            False -> ConsF @\evolv{(if abs x < 10 }@
                              @\evolv{then (x \infix{rem} 10) }@
                              @\evolv{else abs (x \infix{rem} 10))}@
                            @\evolv{(x \infix{quot} 10)}@
\end{minted}

Because this problem requires that we handle negative values, the program that produces the next element should verify if we are at the last digit. In case we are not, we should calculate the absolute value to ensure that we do not display the minus sign. Even though this solution is larger than the previous problem, the search process is simplified for the fact that there is a single argument, thus all the operations must envolve the argument and constant values.

\section{Hylomorphisms}

Here we report the problems that were solved by using hylomorphism.

\subsection{Collatz Numbers}
\emph{Given an integer, find the number of terms in the Collatz (hailstone) sequence starting from that integer.}

\begin{minted}[escapeinside=@@]{haskell}
    -- required primitives: constant int, (==)
    -- (+), (*), mod, div
    collatz :: Int -> Int
    collatz = hylo alg coalg where
        alg NilF = @\evolv{1}@
        alg (ConsF x acc) = @\evolv{1 + acc}@
        
        coalg x =
        case @\evolv{x == 1}@ of
            True -> NilF
            False -> ConsF x @\evolv{(if mod x 2 == 0}@
                              @\evolv{then div x 2 }@
                              @\evolv{else div (3*x + 1) 2)}@
\end{minted}

This is already explained in depth in the text.

\subsection{Sum of Squares} 
\emph{Given integer n, return the sum of squaring each integer in the range [1, n].}

\begin{minted}[escapeinside=@@]{haskell}
    -- required primitives: (^), (+), int constants
    sumOfSquares :: Int -> Int
    sumOfSquares n = hylo alg coalg n where
        alg NilF = @\evolv{0}@
        alg (ConsF x xs) = @\evolv{x\^{}2 + xs}@

        coalg x = case @\evolv{x == 0}@ of
                    True -> NilF
                    False -> ConsF @\evolv{x (x - 1)}@
\end{minted}

The initial condition is the neutral element of addition and we just add the square of each value.

\subsection{Even Squares} 
\emph{Given an integer n, print all of the positive even perfect squares less than n on separate lines.}

\begin{minted}[escapeinside=@@]{haskell}
    -- required primitives: constant int, (+),
    -- (^), (>=), mod
    evenSquares :: Int -> [Int]
    evenSquares n = hylo alg coalg n where
        alg NilF = @\evolv{[]}@
        alg (ConsF x xs) = @\evolv{if x >= n then xs else x : xs}@
        coalg seed = case @\evolv{seed <= 1}@ of
                       True -> NilF
                       False -> ConsF @\evolv{(seed \^{} 2)}@ @\evolv{(seed - 1 - (1 - mod seed 2))}@
\end{minted}

In this program, the starting seed must be a constant value, while the single argument is used to form the condition for stopping growing the structure.

\section{Accumulations}

Here we report the problems that were solved by using accumulation.
The following two problems are already explored in the text, and replicated here for completeness.

\subsection{Last Index of Zero} 
\emph{Given a vector of integers, at least one of which is 0, return the index of the last occurrence of 0 in the vector.}

\begin{minted}[escapeinside=@@]{haskell}
    -- required primitives: if-then-else, (+), (==)
    -- (<>), constant int, Maybe, Last
    lastIndexZero :: [Int] -> Int
    lastIndexZero xs = accu st alg (fromList xs) @\evolv{0}@ where
        st NilF s = NilF
        st (ConsF x xs) s = ConsF x (xs, @\evolv{s+1}@)

        alg NilF i = @\evolv{\negative1}@
        alg (ConsF x acc) i = @\evolv{if x == 0 \&\& acc == -1}@
                               @\evolv{then i}@
                               @\evolv{else acc}@
\end{minted}

\subsection{Vector Average}
\emph{Given a vector of floats, return the average of those floats. Results are rounded to 4 decimal places.}

\begin{minted}[escapeinside=@@]{haskell}
    -- required primitives: (+), (/)
    vecAvg :: [Double] -> Double
    vecAvg xs = accu st alg (fromList xs) @\evolv{(0.0, 0.0)}@ where
        st NilF _ = NilF
        st (ConsF x xs) (s1, s2) = ConsF x (xs, (@\evolv{s1 + x, s2 + 1)}@)

        alg NilF (s1, s2) = @\evolv{s1 / s2}@
        alg (ConsF x acc) s = acc
\end{minted}

\subsection{Checksum}
\emph{Given a string, convert each character in the string into its integer ASCII value, sum them, take the sum modulo 64, add the integer value of the space character, and then convert that integer back into its corresponding character (the checksum character). The program must print Check sum is X, where X is replaced by the correct checksum character.}

\begin{minted}[escapeinside=@@]{haskell}
    -- required primitives: mod, fromEnum, toEnum
    -- constant 64, constant ' ', (+)
    checksum :: String -> Char
    checksum = accu st alg (fromList xs) 0 where        
        st NilF s = NilF
        st (ConsF x xs) s = ConsF x (xs, 
                @\evolv{toEnum (mod (fromEnum x + fromEnum s) 64)}@)
        alg NilF s = @\evolv{toEnum (fromEnum s + 32)}@
        alg (ConsF x xs) s = xs
\end{minted}
The base case (\code{NilF}) is the white space character that must be included into the calculation, after that, at every step, we add the current character value to the value of \code{x}, calculates the modulus $6$ and convert it back to a character (as constrained by the type of \code{xs}).

\subsection{String Differences}
\emph{Given 2 strings (without whitespace) as input, find the indices at which the strings have different characters, stopping at the end of the shorter one. For each such index, print a line containing the index as well as the character in each string.}

\begin{minted}[escapeinside=@@]{haskell}
    -- required primitives: if-then-else, (:), (+), (/=)
    -- constant int, null, head, tail
    stringDiffs :: String -> (String -> [(Int, (Char, Char)))]
    stringDiffs xs ys = accu st alg (fromList xs) @\evolv{0}@ ys where
        st NilF s = NilF
        st (ConsF x xs) s = ConsF x (xs, @\evolv{s+1}@)
        alg NilF s zs = @\evolv{[]}@
        alg (ConsF x xs) s zs =
          @\evolv{if null zs}@
               @\evolv{then []}@
               @\evolv{else if x /= head zs}@
                      @\evolv{then (s, (x, head zs)) : xs (tail zs)}@
                      @\evolv{else xs (tail zs)}@
\end{minted}

The accumulator will then store the index of each pair as it traverse the lists. Afterwards, we compare each character and, if they are different, we store it into the final result, other wise we skip it.

\subsection{X-Word Lines}
\emph{Given an integer X and a string that can contain spaces and newlines, print the string with exactly X words per line. The last line may have fewer than X words.}

\newcommand{\spaceChar}{\chLit{ }}
\newcommand{\newlineChar}{\chLit{{\textbackslash}n}}

\begin{minted}[escapeinside=@@]{haskell}
    -- required primitives: tuples, constant int
    -- constants ' ' and '\n', if-then-else, (/=), (==)
    -- (+), bimap, fst, second, const, snoc, (&&)
    xWordLines :: Int -> String -> String
    xWordLines n xs = accu st alg (fromList xs) @\evolv{1}@ where
        st NilF s = NilF
        st (ConsF x xs) s =
          ConsF x (xs, @\evolv{if x /= \spaceChar \&\& x /= \newlineChar}@
                         @\evolv{then s}@
                         @\evolv{else s + 1}@

        alg NilF s = @\evolv{[]}@
        alg (ConsF x xs) s = @\evolv{if}@
                               @\evolv{| x == \spaceChar \&\& mod s n == 0 -> \newlineChar : xs}@
                               @\evolv{| x == \newlineChar \&\& mod s n /= 0 -> \spaceChar : xs}@
                               @\evolv{| otherwise -> x : xs}@
\end{minted}
For this program, we need the accumulator to hold a value of how many words have been seen so far, we do that by checking for a whitespace or carriage return. If the current char is none of those, we simply concatenate the char into the end of the accumulator. If the char is one of those, we check if we already have $n$ words, if we do we reset the counter and insert the carriage return into the string, otherwise we add one to the counter of words and insert the whitespace. In our base library, the functions \code{fst, snd} return the first and second element of a tuple, respectivelly. The functions \code(first, second, bimap), applies a function to the first, second, or two functions to both elements of a tuple.

\subsection{Word Stats} 
\emph{Given a file, print the number of words containing n characters for n from 1 to the length of the longest word. At the end of the output, print a line that gives the number of sentences and line that gives the average sentence length.}

\begin{minted}[escapeinside=@@]{haskell}
    -- required primitives: Map, if-then-else
    -- (/=), constants ' ', '\n', first, fst, snd
    -- insertWith, constant int
    wordDist :: String -> Map Int Int
    wordDist xs = accu st alg (fromList xs) @\evolv{0}@ where
        st NilF s = NilF
        st (ConsF x xs) s =
          ConsF x (xs,
                    @\evolv{if x /= \spaceChar \&\& x /= \newlineChar}@
                      @\evolv{then s + 1 }@
                      @\evolv{else 0)}@
        alg NilF s = @\evolv{singleton s 1}@
        alg (ConsF x xs) s = @\evolv{if x == \spaceChar || x == \newlineChar}@
                                @\evolv{then insertWith (+) s 1 xs }@
                                @\evolv{else xs)}@

    -- required primitives: constant int, constant '\n'
    -- if-then-else
    lineCount :: String -> Int
    lineCount = cata alg . fromList
      where
        alg NilF = @\evolv{1}@
        alg (ConsF x xs) = @\evolv{xs + if x==\newlineChar then 1 else 0}@

    -- required primitives: tuples, constant '\n'
    -- if-then-else, first, second, uncurry, (/)
    avgLineLen :: String -> Double
    avgLineLen xs = accu st alg (fromList xs) @\evolv{(0.0, 0.0)}@
      where
        f x = @\evolv{if x /= \newlineChar }@
                 @\evolv{then s + 1 }@
                 @\evolv{else s}@
        g x = @\evolv{if x /= \newlineChar }@
                 @\evolv{then s }@
                 @\evolv{else s + 1}@
        st NilF s = NilF
        st (ConsF x xs) s = ConsF x (xs, (f (fst s), g (snd s)))
        alg NilF (a, b) = @\evolv{a / (b + 1)}@
        alg (ConsF x xs) s = @\evolv{xs}@
\end{minted}
This problem is possibly the most challenging of the benchmark to be synthesized all at once. 
One should notice that this is actually three problems into one. The first program deals with calculating the word distribution. 
It initializes the accumulator as a tuple of a count of char and an empty map. 
While traversing the structure, it adds $1$ to the count until it reaches the end of a word. 
When it does, it adds the word length with a count of one. 
The algebra should start adding the last count of the accumulator as it will reach the end of a string without breaking a word. 
After that it just returns the map containing the word distribution.

The second program simply count the occurrences of \code{\newlineChar} indicating a new line. 
The last program, increment one counter for every character of a line and the second for every carriage returns, dividing the result in the end.

We also highlight the fact that folds that do not have any post-processing could always be combined by storing the accumulators into an $n$-tuple and applying the functions in their corresponding elements. 
We do not show that as the resulting program would become too large to report here.